\documentclass[sigconf]{acmart}
\AtBeginDocument{%
  }

\setcopyright{acmlicensed}
\copyrightyear{2018}
\acmYear{2018}
\acmDOI{XXXXXXX.XXXXXXX}
\acmConference[Conference acronym 'XX]{Make sure to enter the correct
  conference title from your rights confirmation emai}{June 03--05,
  2018}{Woodstock, NY}
\acmISBN{978-1-4503-XXXX-X/18/06}

\usepackage{newfloat}
\usepackage{listings}
\usepackage{booktabs}
\usepackage{multirow}
\usepackage{subfigure}
\usepackage{tabularx}
\usepackage{adjustbox}
\usepackage{amsmath}
\begin{document}

%%
%% The "title" command has an optional parameter,
%% allowing the author to define a "short title" to be used in page headers.
\title{Exploring the Effectiveness and Interpretability of Texts in LLM-based Time Series Models}

\author{Zhengke Sun}
\affiliation{%
  \institution{Nanyang Technological University}
  \country{Singapore}}
\email{zsun016@e.ntu.edu.sg}

\author{Hangwei Qian}
\affiliation{%
  \institution{CFAR, A*STAR}
  \country{Singapore}}
\email{qian\_hangwei@cfar.a-star.edu.sg}

\author{Ivor Tsang}
\affiliation{%
  \institution{CFAR, A*STAR}
  \country{Singapore}}
\email{ivor\_tsang@cfar.a-star.edu.sg}

%%
%% By default, the full list of authors will be used in the page
%% headers. Often, this list is too long, and will overlap
%% other information printed in the page headers. This command allows
%% the author to define a more concise list
%% of authors' names for this purpose.

% \renewcommand{\shortauthors}{Trovato et al.}

%%
%% The abstract is a short summary of the work to be presented in the
%% article.

\begin{abstract}
Large Language Models (LLMs) have been applied to time series forecasting tasks, leveraging pre-trained language models as the backbone and incorporating textual data to purportedly enhance the comprehensive capabilities of LLMs for time series. However, are these texts really helpful for interpretation? This study seeks to investigate the actual efficacy and interpretability of such textual incorporations. Through a series of empirical experiments on textual prompts and textual prototypes, our findings reveal that the misalignment between two modalities exists, and the textual information does not significantly improve time series forecasting performance in many cases. Furthermore, visualization analysis indicates that the textual representations learned by existing frameworks lack sufficient interpretability when applied to time series data. We further propose a novel metric named Semantic Matching Index (SMI) to better evaluate the matching degree between time series and texts during our post hoc interpretability investigation. Our analysis reveals the misalignment and limited interpretability of texts in current time-series LLMs, and we hope this study can raise awareness of the interpretability of texts for time series. The code is available at https://github.com/zachysun/TS-Lang-Exp.
\end{abstract}


%%
%% Keywords. The author(s) should pick words that accurately describe
%% the work being presented. Separate the keywords with commas.

%% A "teaser" image appears between the author and affiliation
%% information and the body of the document, and typically spans the
%% page.

% \begin{teaserfigure}
%   \includegraphics[width=\textwidth]{sampleteaser}
%   \caption{Seattle Mariners at Spring Training, 2010.}
%   \Description{Enjoying the baseball game from the third-base
%   seats. Ichiro Suzuki preparing to bat.}
%   \label{fig:teaser}
% \end{teaserfigure}

% \received{20 February 2007}
% \received[revised]{12 March 2009}
% \received[accepted]{5 June 2009}

%%
%% This command processes the author and affiliation and title
%% information and builds the first part of the formatted document.
\maketitle
\section{Introduction}

Time Series Analysis is an essential task in numerical fields, including but not limited to traffic \cite{10260564}, healthcare \cite{morid2023time}, economics \cite{liu2023financial}, and climate science \cite{bi2023accurate}. 
Over the decades, a plethora of algorithms have been studied, from statistical to machine learning methods, and more recently, to deep learning~\cite{wen2023transformers, shao2022spatial} and unsupervised learning models \cite{yang2022unsupervised}, to extract more discriminative patterns and features from the dynamic time series streams.
Currently, pretrained models and Large Language Models (LLMs) have excelled in diverse fields owing to their powerful capabilities in reasoning and cross-task generalization~\cite{DBLP:journals/corr/abs-2310-19852}. Consequently, many studies have leveraged LLMs for various time series tasks and achieved state-of-the-art performance \cite{zhang2024large}. 
Notably, some approaches directly input the tokenized time series to LLMs~\cite{vaswani2017attention, zhou2023one}, while others combine numerical time series data with textual modalities, using text prompts as instructions or embedding the data within a textual language semantic space~\cite{jin2023time, liu2024taming, pan2024textbf, sun2023test}. In this paper, we focus on the latter scenario with textual modalities.

While the above works seem promising for aligning numerical time series with textual language modality, more recent empirical studies on time series forecasting methods find that LLMs do no better than shallower models that replace LLMs with a simple model with attention as the encoder~\cite{DBLP:journals/corr/abs-2406-16964}. 
% It is desirable that the interpretability of such LLM-based time series forecasters can be improved. 
In light of this, we seek to investigate the reasons behind such mixed and conflicting observations for LLM-based time series models (TS-LLM models). Concurrently, we also seek to explore the interpretability of existing TS-LLM models. Studies on interpretability seek to make machine learning systems and their decision-making process understandable to human beings~\cite{DBLP:journals/ai/Miller19}. 
Such interpretability is crucial for time series models, especially for high-stake applications such as healthcare and finance, wherein decisions may have profound consequences. With a clearer understanding of the system's inner reasoning, especially the inner workings of opaque and complex LLMs, the decision-making process within the TS-LLM models is more accessible and comprehensible to users. This in turn can enable timely intervention and control, assuring the safety and trustworthiness of the time series systems.

More specifically, we seek to investigate the alignment between different modalities of numerical time series data and textual data. For standard time series tasks such as forecasting and classification, the output is all numerical, which makes it difficult to incorporate proper texts. Without corresponding time series-text pairs, it is hard to align time series with the expected textual semantic space. Most TS-LLM models feed time series embeddings into a white-box LLM, and these embeddings are more abstract than tokenized prompts and require LLMs to engage in inference. During the inference, the textual prompts might not work well due to the lack of textual supervision. Even though the latest works take into account the compositional and structural differences between two modalities, and there are attempts to integrate and map them to the same latent space via an auxiliary contrastive loss \cite{sun2023test} or by cross-attention \cite{jin2023time}, it remains challenging to harmonize the two modalities.
% \zhengke{ 
% (1) Different from vision-language models (VLMs), despite using LLMs, the tasks of most TS-LLM models are still traditional time series tasks, such as forecasting and classification. That is, the outputs are still time series but not languages, which could lead to the underutilization of text;
% (2) Unlike directly inputting time series into a black-box LLM \cite{liu2024lstprompt}, most TS-LLM models feed time series embeddings into the transformer blocks of a white-box LLM. These embeddings are more abstract than tokenized prompts and require the LLMs to engage in inference. During the inference, prompts might be identified as noise because the output does not supervise language.
% (3) Without time series-texts pairs, it's hard to align time series to expected text semantic space.}
% Compared to the number of TS-LLM models, there is 
% insufficient research on the impact of text, for which we designed multiple ablation experiments and conducted more comprehensive visualizations.
To this end, we intend to investigate the importance of textual data in TS-LLM models, and investigate whether the textual data brings comprehensible and truthful reasoning. We conduct post hoc interpretability studies on two state-of-the-art TS-LLM models, i.e., TimeLLM \cite{jin2023time} and CALF \cite{liu2024taming}. To offer a better interpretation, we further analyze the role of textual data therein and show how misalignment issues occur. Our empirical observations suggest the following findings:
\begin{itemize}
    \item textual data in existing TS-LLM models does not always significantly improve downstream forecasting tasks, and it differs for different language models; 
    \item incorporating textual data does not guarantee interpretable TS-LLM models. TS-LLM models lack sufficient understanding of temporal patterns within the time series and it is also difficult to represent patterns as descriptions that are interpretable to human users. 
\end{itemize}

% We finally find that: (1) Text data in current LLM-based time series models can not always improve forecasting results significantly, and the impact caused by the choice of the language model cannot be overlooked;
% (2) These models still lack interpretability. To be more specific, their understanding of time series is abstract. It is challenging for them to simply represent a certain temporal pattern as a few words, and even more difficult to represent it as words that are interpretable to humans.

In summary, our key contributions are as follows:
\begin{itemize}
    \item We design extensive empirical investigation studies on two state-of-the-art TS-LLM models, i.e., TimeLLM and CALF. We evaluate the efficacy of texts for long-term, few-shot and zero-shot forecasting tasks. Results show that models incorporated with texts don't achieve the best performance in many cases. 
    % specific ablation methods for two LLM-based time series models to evaluate the effect of texts for long-term, few-shot and zero-shot forecasting. We find that models with texts don't demonstrate the best performance in many cases. 
    \item We study how the inner semantics of texts affect results by randomly replacing prompt embeddings and word embeddings. We find that more random textual inputs don't necessarily lead to worse results. 
    \item Further visualizations on text prototypes, attention weights and time series tokens show that they are irrelevant to time series properties.
    % \item We visualize the text prototypes, attention weights from alignment module or language models and aligned time series tokens, but they show irrelevance to time series properties. 
    \item We also study the relationship between time series and their semantic representations and propose a novel metric to evaluate the matching degree of them. The finding is that it's difficult for current models to link certain temporal pattern with certain textual token and different language models show varying capabilities in matching time series with texts. 
\end{itemize}

\section{Related Work}
% The related works of our paper are mainly divided into two aspects: (1) LLM-based time series models; (2) interpretability of cross-modality models.

\textbf{LLM-based Time Series Models} There is increasing number of works leveraging the power of LLMs to conduct time series tasks \cite{jiang2024empowering}. 
For example, 
LLM4TS \cite{chang2023llm4ts} is a two-stage framework, forecasting the next patch (LoRA \cite{hu2021lora} fine-tuning) and target time series (full fine-tuning) separately. 
GPT4TS \cite{zhou2023one} is a similar framework by fine-tuning partial layers. 
Moreover, TimeLLM \cite{jin2023time} incorporates texts into the architecture without fine-tuning language models and proposes two influential designs: prompt-as-prefix and an attention layer reprogramming time series and text prototypes. 
Similarly, UniTime \cite{liu2024unitime} uses domain instructions as prompts to handle varying characteristics of time series.
TEST \cite{sun2023test} matches texts and time series through contrastive learning. 
CALF \cite{liu2024taming} also utilizes textual information and designs two losses to reduce the modality distribution gap. 
TimeCMA \cite{liu2024timecma} encodes time series and text prompts separately and then aligns the two modalities.
Many approaches incorporate LLMs into time series, and some of them achieve state-of-the-art results by incorporating textual information. However, it remains unclear how important those texts are and how it contribute to the interpretability of such models. This paper aims to elucidate the role of textual data therein.

% From above, researchers incorporated LLMs into time series 
% forecasting, and some of them, by combining textual information, achieved state-of-the-art results.
% However, the work exploring the importance of texts is not enough. 
% So we conducted comprehensive ablation experiments to further elucidate the role of textual data in time series forecasting.

\textbf{Interpretability of Cross-Modality Models}
Compared to the interpretability works on TS-LLM models, there have been some works related to visualization and causal tracing for vision-language models. \cite{Aflalo_2022_CVPR, Palit_2023_ICCV, wang2024language}. 
For example, VL-InterpreT \cite{Aflalo_2022_CVPR} is one
of visualization tools for interpreting attentions and hidden representations in VLMs. 
These attentions can reflect the relation between tokens and image patches, for instance, token “person\_0” is highly related to a person in an image, while token “plants” is highly related to the grass in the same image. 
Currently, there are some works on the interpretability of LLMs for time series \cite{yu2023harnessing, liu2024large}, but they mainly focus on adopting black-box LLMs to understand time series, rather than digging into the TS-LLM model and conducting interpretability research on the internal weights of the network.
For existing TS-LLM models, research on interpretability is shallow and lacks comprehensive inspection. For example, TimeLLM \cite{jin2023time} provides visualization of attentions of the reprogramming layer and a subset of text prototypes. In CALF \cite{liu2024taming}, some time series instances (variables) are visualized. Both models indicate relevant texts about time series properties. We expand upon their works and visualize more modules to explore interpretability more cohesively and conclusively.

% For existing TS-LLM models, research on interpretability is shallow and lacks comprehensive inspection. For example, in TimeLLM \cite{jin2023time}, they visualized the attentions of reprogramming layer and some of text prototypes. In CALF \cite{liu2024taming}, they visualized the time series instances (variables). Both of them show relevance to words about time series properties. We expanded upon their works and visualized more modules to explore the interpretability.

\section{Experimental Setup}
We study two state-of-the-art TS-LLM models: TimeLLM \cite{jin2023time} and CALF \cite{liu2024taming}. TimeLLM is significant as it represents the first TS-LLM model that combines both text prompts and text prototypes. This combination is designed to enhance the model’s ability to predict time series by leveraging textual information. 
CALF, although similar to TimeLLM in its use of text prototypes, differs fundamentally in its architecture. It employs a unique two-branch structure, which is special and different from most existing TS-LLM models. 
We primarily conduct experiments on TimeLLM, and provide supplementary experiments on CALF to show that our evaluations can be applicable and generalizable to more models. The detailed frameworks and language model selections are as follows:

\textbf{TimeLLM }\cite{jin2023time}: TimeLLM transforms time series into tokens through channel-independence and patching \cite{Yuqietal-2023-PatchTST}. These tokens are put together with text prototypes into a cross-attention module, i.e., the reprogramming layer, to align with natural language. Next, the aligned tokens, concatenated with prompts, are processed with a frozen pre-trained language model. Finally, the outputs are mapped to the target shape by a linear layer. To ensure the robustness and generalizability of experimental results, we select multiple language models for TimeLLM: 1) GPT-2: GPT-2 \cite{radford2019language} is a commonly used transformer-based language model trained with the autoregressive manner. 2) BERT Family: BERT \cite{Devlin2019BERTPO} is a bidirectional transformer model trained by masked language modeling and next sentence prediction; DistilBERT \cite{sanh2019distilbert}, a distilled version of BERT, is smaller and faster. 3) OPT Family: OPT \cite{zhang2022opt} is a suite of robust and open-source decoder-only pre-trained language models. The parameters range from 125M to 175B. We adopt two smaller versions (125M and 350M). 
4) Llama 2: Llama 2 \cite{touvron2023llama} is a cutting-edge open-source large language model by Meta, renowned for its superior performance, scalability, and accessibility in the open-source LLM ecosystem.

% \begin{figure*}[ht!]
%     \centering
%     \subfigure[Base]{
%         \includegraphics[width=0.22\textwidth]{figures/paper_TimeLLM_Base.png}
%         \label{base}
%     }\hfill
%     \subfigure[Base\_Prompt]{
%         \includegraphics[width=0.22\textwidth]{figures/paper_TimeLLM_Prompt.png}
%         \label{base_prompt}
%     }\hfill
%     \subfigure[Base\_Prototype]{
%         \includegraphics[width=0.22\textwidth]{figures/paper_TimeLLM_Prototype.png}
%         \label{base_prototype}
%     }\hfill
%     \subfigure[TimeLLM]{
%         \includegraphics[width=0.25\textwidth]{figures/paper_TimeLLM_Visualization_Parts.png}
%         \label{timellm}
%     }
%     \caption{TimeLLM and Its Ablations.V1-V5 in (d): Visualization parts of TimeLLM}
% \end{figure*}

% \begin{figure*}[!htbp]
%     \centering
%     \includegraphics[width=0.6\textwidth]{figures/paper_TimeLLM_Visualization_Parts.png}
%     \caption{Architecture of TimeLLM. V1-V5: Visualization parts of TimeLLM; Base model: remove both part 1 and part 2; Base\_Prompt model: only remove part 2; Base\_Prototype model: only remove part 1.}
%     \label{timellm}
% \end{figure*}

\textbf{CALF} \cite{liu2024taming}: Unlike TimeLLM, CALF uses each variable as a token without patching. It includes two branches: the textual source branch and the temporal target branch. The two branches are combined with a cross-modal match module and are processed by a frozen pretrained language model and a LoRA fine-tuning pretrained language model separately. For CALF, we follow the original paper to utilize the GPT2 as a language model. 
\begin{figure*}[ht!]
    \centering
    \subfigure[Architecture of TimeLLM. V1-V5: Visualization parts of TimeLLM.]{
        \includegraphics[width=0.34\textwidth]{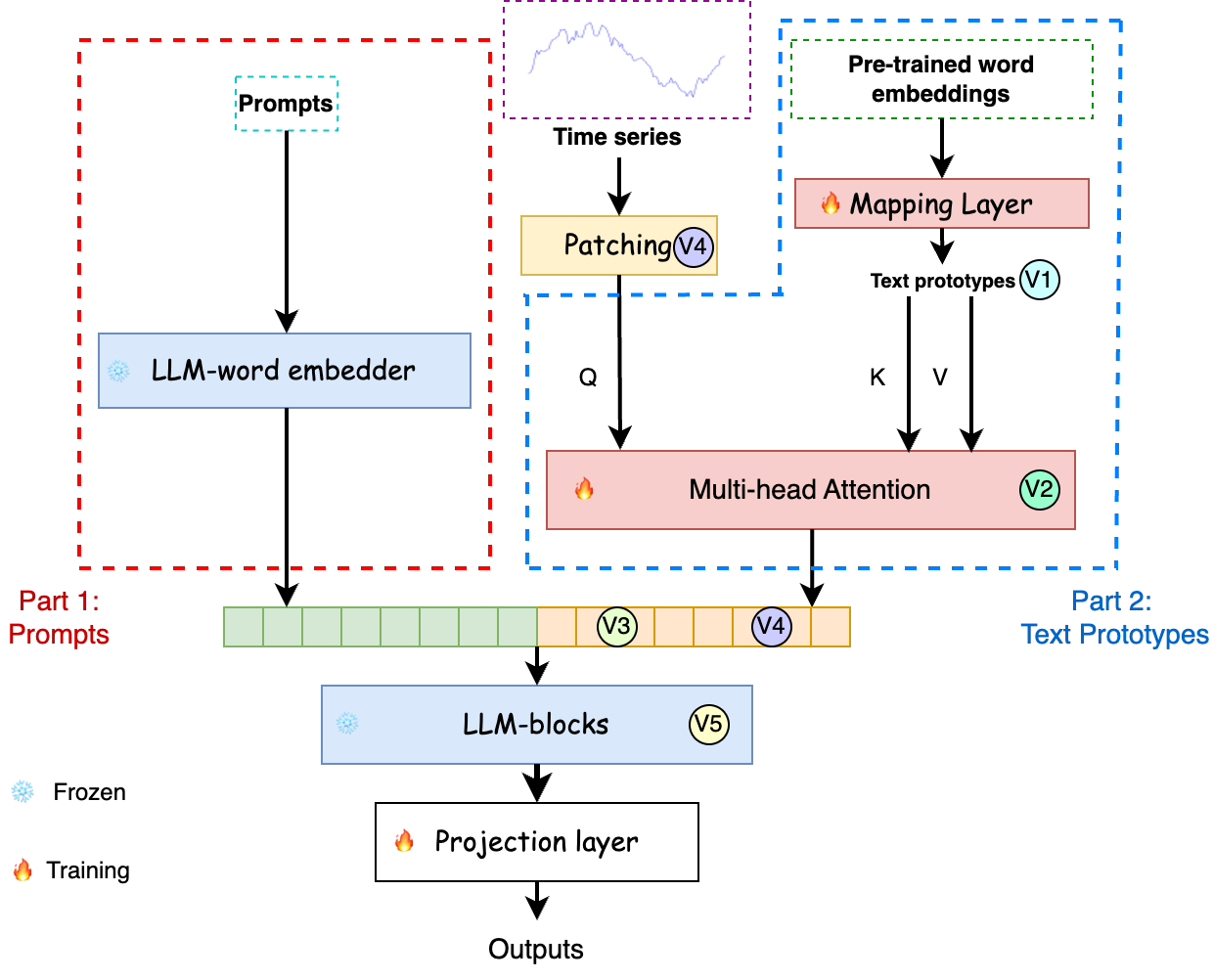}
        \label{timellm}
    }\hfill
    \subfigure[Architecture of CALF. V1-V3: Visualization parts of CALF.]{
        \includegraphics[width=0.58\textwidth]{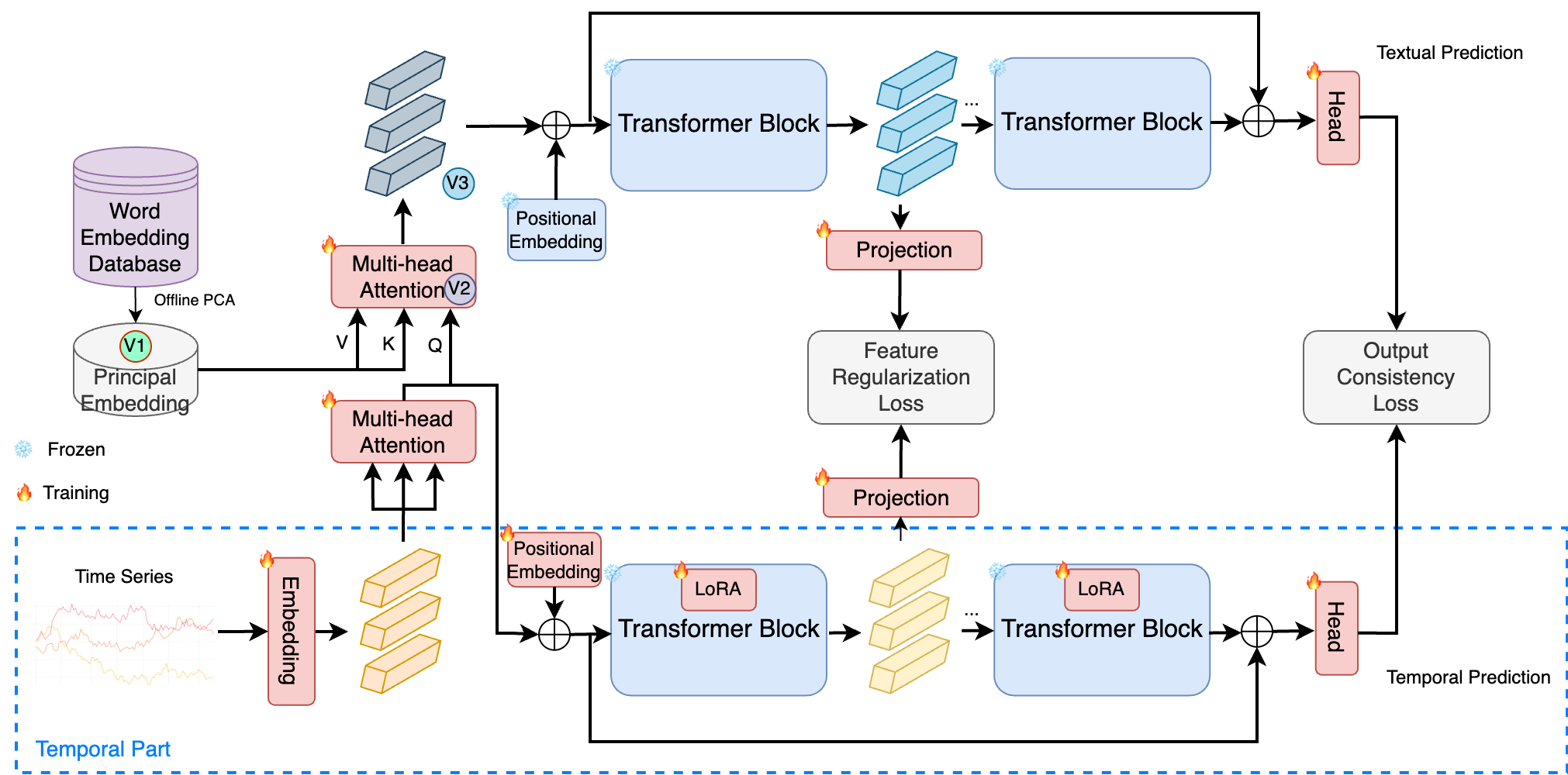}
        \label{calf}
    }
    \caption{Overview of TimeLLM and CALF}
\end{figure*}

% \begin{figure*}[!htbp]
%     \centering
%     \includegraphics[width=0.7\textwidth]{figures/paper_CALF_Visualization_Parts.png}
%     \caption{Architecture of CALF. V1-V3: Visualization parts of CALF; CALF\_Temporal model is the temporal part of it.}
%     \label{calf}
% \end{figure*}

\section{Experiments}
Our experiments can be summarized into three parts: 
\textbf{Part 1:} Investigating whether text prompts and text prototypes are helpful for forecasting, including long-term, few-shot and zero-shot forecasting. We intend to comprehensively evaluate the forecasting capabilities of TS-LLM models and hope that by adding texts, the models can demonstrate better few-shot and zero-shot capabilities \textbf{(Q1-Q3)}; \textbf{Part 2:} Exploring the impact of text prototypes on forecasting, including different text prototypes extraction methods, different numbers and hidden dimensions of text prototypes \textbf{(Q4-Q6)}. We aim to determine the feasibility of improving the forecasting results by altering the text prototypes involved in alignment modules; \textbf{Part 3:} Investigating whether the model has truly learned the inner semantics in the text prototypes and text prompts \textbf{(Q7)}. If the text is helpful, we want to find out whether this help is truly due to its inner semantics, rather than an increase in the number of parameters. 

All experiments are implemented on 4 NVIDIA L40 GPUs. GPT2 is the default language model unless otherwise stated.

 % The experimental environment and parameter settings may differ slightly from the original paper.
 
\subsection{The effectiveness of text prompts and text prototypes for forecasting (Part 1)}

For TimeLLM (Fig.\ref{timellm}), we propose three variants: (1) Base\_Prompt: only removing the text prototypes module; (2) Base\_Prototype: only removing the text prompts module; (3) Base: removing both of them; 
For CALF (Fig.\ref{calf}), as it doesn't contain prompts, there is only one variant CALF\_Temporal (Fig.\ref{calf_temporal} in Appendix A.4\footnote{We will omit `Appendix' thereafter to be more concise.}), which removes textual source branch and cross-modal match module.

% For all questions in this part, we conduct experiments on TimeLLM and its ablations.
% For question 1, we also experiment with TimeLLM based on other language models (BERT, DistilBERT, OPT) and test CALF with its ablation.

\textit{\textbf{Are text prompts and text prototypes helpful for long-term forecasting? (Q1)}}
Table \ref{tab:ltf-timellm-part-gpt2} shows the results of TimeLLM and its ablations (full results in Table \ref{tab:ltf-timellm-full} in B.1). It can be observed that Base model exceeds models with texts in various datasets. Table \ref{tab:ltf-calf-part} displays the performance of CALF and CALF\_Temporal (full results refer to Table \ref{tab:ltf-calf-full} in B.1), and the latter achieves better results in most datasets.
We can also find that the selection of language model affects the forecasting performance. If GPT-2 is adopted, models without texts are the best. 
% For BERT or DistilBERT (see Table \ref{tab:ltf-timellm-bert} and Table \ref{tab:ltf-timellm-distilbert} in B.1), leveraging texts is a better choice. 
For BERT (see Table \ref{tab:ltf-timellm-bert}, \ref{tab:ltf-timellm-distilbert} in B.1) and Llama 2 (see Table \ref{tab:ltf-timellm-part-llama2}), leveraging texts is a better choice. 
However, it doesn't mean both text prompts and text prototypes are required to achieve the best performance, i.e., using either one of them may be better. For instance, only leveraging text prompts performs better on ETTh1 and ETTh2 datasets for BERT and DistilBERT. Furthermore, if we employ OPT-125M or OPT-350M (see Table \ref{tab:ltf-timellm-opt125} and Table \ref{tab:ltf-timellm-opt350} in B.1) as language models, the situation would be different too.
For these two language models, only combining text prompts performs the worst, especially OPT-350M.
Experimental results indicate that current alignment methods are not universally suitable for language models.
% Besides, the results of using two types of texts simultaneously is not always better than using just one of them. 

\begin{table*}[!htbp]
\centering
\begin{tabular}{@{}ccccccccc}
\toprule
Models & \multicolumn{2}{c}{Base} & \multicolumn{2}{c}{Base\_Prompt} & \multicolumn{2}{c}{Base\_Prototype} & \multicolumn{2}{c}{TimeLLM}   \\ \midrule
Dataset\textbackslash Metrics & MSE & MAE & MSE & MAE & MSE & MAE & MSE & MAE \\ \midrule
ETTh1& \textbf{0.419}& \textbf{0.436}& \underline{0.433}& \underline{0.444}& 0.441& 0.451& 0.448& 0.456\\
ETTh2& 0.368& 0.407& \textbf{0.363}& \underline{0.404}& 0.373& 0.409& \underline{0.366}& \textbf{0.403}\\
ETTm1& 0.359& \textbf{0.386}& 0.363& \underline{0.387}& \textbf{0.352}& 0.388& \underline{0.355}& 0.388\\
ETTm2& \textbf{0.262}& \textbf{0.324}& \underline{0.263}& \underline{0.326}& 0.272& 0.330& 0.268& \underline{0.326}\\
Weather& \underline{0.228}& \underline{0.266}& \textbf{0.227}& \textbf{0.265}& 0.231& 0.268& 0.231& 0.269 \\
Electricity & \textbf{0.162}& \textbf{0.258}& \underline{0.163}& \underline{0.260}& 0.169& 0.273& 0.167& 0.269\\
Traffic & \underline{0.400}& \textbf{0.274}& \underline{0.400}& \underline{0.275}& \textbf{0.395}& 0.281& \textbf{0.395}& 0.279\\ 
\hline
1st count & 3& 4& 2& 1& 2& 0& 1& 1\\
2nd count & 2& 1& 4& 6& 0& 0& 2& 1\\
\bottomrule
\end{tabular}
\caption{Long-term forecasting results of TimeLLM and its ablations. Bold: best results; underline: second best results. “1st count”: number of best results; “2nd count”: number of second best results.}
\label{tab:ltf-timellm-part-gpt2}
\end{table*}

\begin{table*}[!htbp]
\centering
\begin{tabular}{@{}ccccccccc}
\toprule
Models & \multicolumn{2}{c}{Base} & \multicolumn{2}{c}{Base\_Prompt} & \multicolumn{2}{c}{Base\_Prototype} & \multicolumn{2}{c}{TimeLLM}   \\ \midrule
Dataset\textbackslash Metrics & MSE & MAE & MSE & MAE & MSE & MAE & MSE & MAE \\ \midrule
ETTh1& 0.442& 0.452& 0.436& \textbf{0.446}& \textbf{0.430}& 0.450& \underline{0.434}& \underline{0.448}\\
ETTh2& 0.380& 0.414& \underline{0.378}& \underline{0.411}& 0.380& 0.412& \textbf{0.370}& \textbf{0.405}\\
ETTm1& \underline{0.362}& \underline{0.393}& \textbf{0.359}& \textbf{0.392}& 0.376& 0.399& 0.368& \underline{0.393}\\
ETTm2& 0.282& 0.338& 0.281& 0.340& \textbf{0.265}& \textbf{0.322}& \underline{0.268}& \underline{0.324}\\
\hline
1st count & 0& 0& 1& 2& 2& 1& 1& 1\\
2nd count & 1& 1& 1& 1& 0& 0& 2& 3\\
\bottomrule
\end{tabular}
\caption{Long-term forecasting results of TimeLLM and its ablations. (Language model: Llama2)}
\label{tab:ltf-timellm-part-llama2}
\end{table*}

\begin{table}[!ht]
\centering
\begin{tabular}{@{}ccccc}
\toprule
Models & \multicolumn{2}{c}{CALF\_Temporal} & \multicolumn{2}{c}{CALF}     \\ \midrule
Dataset\textbackslash Metrics& MSE & MAE & MSE & MAE \\ \midrule
ETTh1& 0.442& \textbf{0.427}& \textbf{0.435}& 0.432\\
ETTh2& \textbf{0.367}& \textbf{0.391}& 0.374& 0.397\\
ETTm1& \textbf{0.382}& \textbf{0.379}& 0.398& 0.393\\
ETTm2& \textbf{0.278}& \textbf{0.319}& 0.284& 0.324\\
Weather & \textbf{0.248}& \textbf{0.273}& 0.251& 0.274\\
Electricity & 0.180& 0.270& \textbf{0.174}& \textbf{0.265}\\
Traffic & 0.463& 0.302& \textbf{0.441}& \textbf{0.282}\\ 
\hline
1st count & 4& 5& 3& 2\\
\bottomrule
\end{tabular}
\caption{Long-term forecasting results of CALF and its ablations.  }
\label{tab:ltf-calf-part}
\end{table}

\textit{\textbf{Are text prompts and text prototypes helpful for few-shot forecasting? (Q2)}}
% LLMs are shown to require few-shot examples before achieving strong performance \cite{brown2020language}.
We conduct this experiment to explore whether adding texts can enhance or diminish the few-shot capability of the TS-LLM model.
Results in Table \ref{tab:fsf-timellm-part} indicate that texts may not benefit TS-LLMs (full results refer to Table \ref{tab:fsf-timellm-full} in B.2). Base model achieves the best performance overall, followed by Base\_Prompt model. 
Furthermore, the few-shot ability highly depends on datasets.
For ETT datasets, TimeLLM and Base\_Prototype model are obviously worse than other two models. But for Electricity dataset, the results of four models are comparable.

\begin{table*}[!htbp]
\centering
\begin{tabular}{@{}ccccccccc}
\toprule
Models & \multicolumn{2}{c}{Base} & \multicolumn{2}{c}{Base\_Prompt} & \multicolumn{2}{c}{Base\_Prototype} & \multicolumn{2}{c}{TimeLLM}   \\ \midrule
Dataset\textbackslash Metrics& MSE & MAE & MSE & MAE & MSE & MAE & MSE & MAE \\ \midrule
ETTh1& \textbf{0.632}& \textbf{0.538}& \underline{0.639}& \underline{0.542}& 0.766& 0.596& 0.746& 0.586\\
ETTh2& \textbf{0.397}& \textbf{0.430}& \underline{0.399}& \underline{0.432}& 0.430& 0.451& 0.448& 0.493\\
ETTm1& \textbf{0.496}& \textbf{0.460}& 0.538& 0.475& 0.507& 0.471& \textbf{0.486}& \underline{0.462}\\
ETTm2& \underline{0.291}& \textbf{0.337}& 0.303& 0.346& \textbf{0.285}& \underline{0.338}& 0.299& 0.345\\
Electricity & \textbf{0.180}& \underline{0.278}& \textbf{0.180}& \textbf{0.277}& 0.182& 0.279& \underline{0.181}& \textbf{0.277}\\
\hline
$1^{st}$ count & 4& 4& 1& 1& 1& 0& 1& 1\\
$2^{nd}$ count & 1& 1& 2& 2& 0& 1& 1& 1\\
\bottomrule
\end{tabular}
\caption{Few-shot forecasting results of TimeLLM and its ablations. The percent of dataset is set to 10\%. }
\label{tab:fsf-timellm-part}
\end{table*}

\textit{\textbf{Are text prompts and text prototypes helpful for zero-shot forecasting? (Q3)}}
Recent TS-LLM models \cite{jin2023time, liu2024taming, liu2024unitime} all consider zero-shot ability as an advantage of pretrained language models. We hope to explore whether the addition of texts can enable the pretrained language model to recognize specific inputs, thereby improving forecasting performance.
As Table \ref{tab:zsf-timellm-part}, TimeLLM (full results refer to Table \ref{tab:zsf-timellm-full} in B.3) using 100 text prototypes performs the best, while Base model without pretraining and fune-tuning performs the worst. However, Base model is better than models combining texts in last two cases (h1$\rightarrow$m1, h1$\rightarrow$m2).
Comparing Base model without pretraining and fune-tuning with other models, pretrained language models indeed play a role in enhancing zero-shot forecasting capabilities. However, on this basis, adding texts does not necessarily improve results. Therefore, adding more text prototypes may not have a positive effect. 
% The zero-shot ability of pretrained language models are enough.

\begin{table*}[!htbp]
\centering
\begin{adjustbox}{max width=\textwidth}
\begin{tabular}{@{}ccccccccccccccc}
\toprule
Models & \multicolumn{2}{c}{Base(woPre+woFT)} &  \multicolumn{2}{c}{Base}& \multicolumn{2}{c}{Base\_Prompt} & \multicolumn{2}{c}{Base\_Prototype(100)} & \multicolumn{2}{c}{Base\_Prototype(1000)}& \multicolumn{2}{c}{TimeLLM(100)} & \multicolumn{2}{c}{TimeLLM(1000)}\\ \midrule
Dataset\textbackslash Metrics& MSE & MAE   & MSE&MAE& MSE & MAE & MSE & MAE  & MSE&MAE& MSE & MAE  & MSE&MAE\\ \midrule
h1$\rightarrow$h2& 0.354&   0.401& \underline{0.344}&0.392& 0.347& 0.396& 0.345&  \underline{0.389}& 0.352&0.394& \textbf{0.338}&  \textbf{0.386}& \underline{0.344}&0.390\\
m1$\rightarrow$m2& 0.270&   0.326& 0.267&0.323& 0.267& 0.325& 0.269&  0.322& \underline{0.263}&\underline{0.320}& \underline{0.263}&  \underline{0.320}& \textbf{0.262}&\textbf{0.318}\\
h1$\rightarrow$m1& 0.782&   0.591& \textbf{0.730}&\textbf{0.561}& \underline{0.736}& \underline{0.566}& 0.810&  0.582& 0.786&0.583& 0.775&  0.579& 0.767&0.580\\
h1$\rightarrow$m2& 0.306&   0.364& \textbf{0.300}&\underline{0.358}& 0.305& 0.362& 0.322&  0.367& 0.313&0.365& \textbf{0.300}&  \textbf{0.356}& \underline{0.303}&0.360\\
\hline
$1^{st}$ count & 0& 0& 2& 1& 0& 0& 0& 0& 0& 0& 2& 2& 1& 1\\
$2^{nd}$ count & 0& 0& 1& 1& 1& 1& 0& 1& 1& 1& 1& 1& 2& 0\\
\hline
\end{tabular}
\end{adjustbox}
    \caption{Zero-shot forecasting results of TimeLLM and its ablations. ‘100/1000' means the number of text prototypes is 100/1000. ‘h1$\rightarrow$h2': training on ETTh1, testing on ETTh2. }
\label{tab:zsf-timellm-part}
\end{table*}

\subsection{The effectiveness of text prototypes for forecasting (Part 2)}
The text prototype is one of the main textual inputs. Compared with the more format-fixed text prompts \cite{liu2024timecma}, text prototypes obtained through different extraction methods vary greatly.
Up to now, all the text prototypes extraction methods can be grouped to two categories: one is learning-based, including the linear mapping layer (TimeLLM), and the other is rule-based extraction, including PCA \cite{liu2024taming}, K-means, random selection, and provided text \cite{sun2023test}. In addition, we supplement a method namely “provided text with their similar tokens", which  compared to directly providing text, can offer a more flexible number of text prototypes.

\textit{\textbf{Do different text prototypes extraction methods affect forecasting performance? (Q4)}}
Different text prototypes extraction methods yield significantly different prototypes, and these different prototypes can align the time series to various semantic spaces.
We find that using PCA to obtain text prototypes yields the best results (refer to Table \ref{tab:pm-timellm-full} in B.4), followed by the linear mapping layer. Directly using text prototypes (such as method “provided text”) related to time series do not get better results. Furthermore, the PCA method's performance still does not significantly outperform the Base model.
Using specific text prototypes can force the association of time series with interpretable words, but this does not necessarily lead to better forecasting results. Text prototypes obtained through PCA are orthogonal to each other, hence may cover a richer semantic content. Maintaining better forecasting performance while enhancing the interpretability of text prototypes is still a problem that requires further research.

% \begin{table*}[!htbp]
% \centering
% \begin{adjustbox}{max width=\textwidth}
% \begin{tabular}{@{}ccccccccccccc}
% \toprule
% Methods & \multicolumn{2}{c}{Linear (Default)} & \multicolumn{2}{c}{PCA}  & \multicolumn{2}{c}{Kmeans} & \multicolumn{2}{c}{Random} & \multicolumn{2}{c}{Text} & \multicolumn{2}{c}{Similarity}\\ \midrule
% Dataset\textbackslash Metrics& MSE & MAE & MSE & MAE & MSE & MAE  & MSE&MAE & MSE& MAE& MSE&MAE\\ \midrule
% ETTh1& \underline{0.452}& 0.459& \textbf{0.431}& 0.449& 0.464&  0.470& 0.482& 0.481& \underline{0.452}& 0.460& 0.473&0.475\\
% ETTh2& \textbf{0.360}& \textbf{0.399}& \underline{0.362}& \underline{0.401}& 0.371&  0.407& 0.369& 0.405& 0.372& 0.403& 0.370&0.406\\
% ETTm1& \underline{0.355}& \underline{0.387}& \textbf{0.351}& \textbf{0.385}& 0.363&  0.394& 0.360& 0.391& 0.363& 0.395& 0.359&0.392\\
% ETTm2& 0.269& 0.327& \underline{0.261}& 0.323& \underline{0.261}&  0.322& \underline{0.261}& \underline{0.321}& 0.266& 0.325& \textbf{0.260}& \textbf{0.320}\\
% \hline
% $1^{st}$ count & 1& 1& 2& 1& 0& 0& 0& 0& 0& 0& 1& 1\\
% $2^{nd}$ count & 2& 1& 2& 1& 1& 0& 1& 1& 1& 0& 0& 0\\
% \hline
% \end{tabular}
% \end{adjustbox}
% \caption{Long-term forecasting results with different text prototypes extraction methods. The number of text prototypes is set to 100. }
% \label{tab:pm-timellm-part}
% \end{table*}

\textit{\textbf{Do different numbers of text prototypes affect forecasting performance? (Q5)}}
The scarcity of text prototypes may lead to a lack of semantic information, while an excess of text prototypes might introduce noise, causing the time series to be aligned to incorrect semantic space. We aim to find an appropriate number of text prototypes to enhance forecasting performance.
The best results are achieved using only 50 text prototypes (refer to Table \ref{tab:pn-timellm-full} in B.5), but in the meantime, difference in results is not significant.
Therefore, the number of text prototypes is not the key factor affecting results. One possible reason is that only a few text prototypes have higher weights in relation to time series.

% \begin{table*}[!htbp]
% \centering
% \begin{tabular}{cccccccccc}
% \hline
% \multicolumn{2}{c}{Number} & \multicolumn{2}{c}{50} & \multicolumn{2}{c}{100} & \multicolumn{2}{c}{500} & \multicolumn{2}{c}{1000}\\
%  Dataset&  Window& MSE& MAE& MSE&MAE & MSE&MAE & MSE&MAE\\
% \hline
% \multirow{2}{*}{ETTh1}& 96 & \textbf{0.376}&  \textbf{0.404}& 0.377& 0.405& 0.377& 0.405& 0.378&0.407\\
%  & 192 & 0.417&  0.433& 0.421& 0.437& 0.419& 0.434& \textbf{0.414}&\textbf{0.431}\\
% \hline
% \multirow{2}{*}{ETTm1}& 96 & 0.295&  0.351& 0.291& \textbf{0.347}& \textbf{0.290}& 0.348& 0.293&0.350\\
%  & 192 & \textbf{0.331}&  \textbf{0.374}& 0.344& 0.379& 0.337& 0.378& 0.334&\textbf{0.374}\\
% \hline
% \multicolumn{2}{c}{$1^{st}$ count} & 2&   2& 0& 1& 1& 0& 1&2\\
% \hline
% \end{tabular}
% \caption{Long-term forecasting results with different numbers of text prototypes. }
% \label{tab:pn-timellm-part}
% \end{table*}

\textit{\textbf{Do different hidden dimensions of text prototypes affect forecasting performance? (Q6)}}
We intend to explore the effectiveness of dimensionality reduction for text prototypes, as their dimensions are generally much larger than those of time series embeddings, and this disparity may affect the alignment of the two modalities.
We can find the best results are achieved without dimensionality reduction of text prototypes (as 768 for GPT2), followed by a reduction to 256 (refer to Table \ref{tab:pd-timellm-full} in B.6). Similar to Q5, the differences in results are not significant too.
Although sometimes dimensionality reduction of text prototypes can lead to better results, it is currently not considered a necessity.

% \begin{table*}[!htbp]
% \centering
% \begin{tabular}{cccccccccc}
% \hline
% \multicolumn{2}{c}{Dimension} & \multicolumn{2}{c}{16} & \multicolumn{2}{c}{64} & \multicolumn{2}{c}{256} & \multicolumn{2}{c}{768}\\
%  Dataset&  Window& MSE& MAE& MSE&MAE & MSE&MAE & MSE&MAE\\
% \hline
% \multirow{2}{*}{ETTh1}& 96 & 0.382&  0.411& 0.388& 0.415& 0.379& 0.407& \textbf{0.377}&\textbf{0.405}\\
%  & 192 & 0.423&  0.436& 0.428& 0.441& \textbf{0.413}& \textbf{0.429}& 0.421&0.437\\
% \hline
% \multirow{2}{*}{ETTm1}& 96 & 0.295&  0.352& 0.295& 0.352& 0.295& 0.350& \textbf{0.291}&\textbf{0.347}\\
%  & 192 & \textbf{0.333}&  \textbf{0.375}& 0.334& \textbf{0.375}& 0.335& \textbf{0.375}& 0.344&0.379\\
% \hline
% \multicolumn{2}{c}{$1^{st}$ count} & 1&   1& 0& 1& 1& 2& 2&2\\
% \hline
% \end{tabular}
% \caption{Long-term forecasting results with different hidden dimensions of text prototypes. }
% \label{tab:pd-timellm-part}
% \end{table*}

\subsection{The influence of inner semantics for forecasting (Part 3)}

\textit{\textbf{Do the inner semantics of text prompts and text prototypes affect forecasting performance? (Q7)}}
Although models that integrate texts can achieve good results in some cases, it is questionable whether these models have truly learned the inner semantics. If some values of the prompt or word embeddings are randomly replaced, the model receives disrupted texts, and ideally, its performance should be worse.
In Table \ref{tab:ltf-timellm-replaced-part}, if replacing prompts, for GPT 2, the results of 10\% replacement ratio are better than others. 
However, the results with 100\% replacement ratio do not show a significant decline (full results refer to Table \ref{tab:ltf-timellm-random-replace-prompts-full} in B.7); for Llama 2, no replacement achieves the best performance, but the results with 100\% also have two $2^{nd}$ best values (full results refer to Table \ref{tab:ltf-timellm-random-replace-prompts-full-llama2} in B.7).
If replacing words , for both GPT 2 and Llama 2, the best results are obtained without random replacing and higher ratios do not always lead to a decline (full results refer to Table \ref{tab:ltf-timellm-random-replace-words-full} and Table \ref{tab:ltf-timellm-random-replace-words-full-llama2} in B.7).
Despite the best results being achieved with prompt or word embeddings that are not randomly replaced or replaced by a small ratio, the relationship between the replacement ratio and forecasting performance is not inversely proportional, making it still difficult to prove that the model is truly sensitive to languages.

\begin{table*}[!ht]
\centering
\begin{tabular}{@{}llcccccccccc@{}}
\hline
\multicolumn{2}{l}{Method} & \multicolumn{2}{c}{0\%} & \multicolumn{2}{c}{10\%} & \multicolumn{2}{c}{40\%} & \multicolumn{2}{c}{70\%} & \multicolumn{2}{c}{100\%} \\ 
\cline{3-12}
\multicolumn{2}{l}{Dataset\textbackslash Metrics} & MSE & MAE & MSE & MAE & MSE & MAE & MSE & MAE & MSE & MAE \\ 
\hline
\multicolumn{12}{l}{\textbf{Random Replaced Prompts}} \\
ETTh1   &         & 0.448 & \underline{0.456} & \textbf{0.441} & \textbf{0.453} & \underline{0.447} & 0.457 & 0.453 & 0.462 & 0.466 & 0.467 \\
ETTh2   &         & \textbf{0.366} & 0.403 & \underline{0.367} & \underline{0.402} & \textbf{0.366} & 0.404 & \textbf{0.366} & \textbf{0.400} & \underline{0.367} & \underline{0.402} \\
\hline
$1^{st}$ count & & 1 & 1 & 1 & 1 & 1 & 0 & 1 & 1 & 0 & 0 \\
$2^{nd}$ count & & 0 & 0 & 1 & 1 & 1 & 0 & 0 & 0 & 1 & 1 \\
\midrule
\multicolumn{12}{l}{\textbf{Random Replaced Words}} \\
ETTh1   &         & \underline{0.448} & \underline{0.456} & \textbf{0.439} & \textbf{0.452} & 0.452 & 0.459 & 0.469 & 0.471 & 0.450 & 0.464 \\
ETTh2   &         & \textbf{0.366} & \underline{0.403} & 0.376 & 0.409 & \underline{0.369} & \textbf{0.402} & 0.371 & 0.405 & 0.370 & \underline{0.403} \\
\hline
$1^{st}$ count & & 1 & 0 & 1 & 1 & 0 & 1 & 0 & 0 & 0 & 0 \\
$2^{nd}$ count & & 1 & 2 & 0 & 0 & 1 & 0 & 0 & 0 & 0 & 1 \\
\midrule
\multicolumn{12}{l}{\textbf{Random Replaced Prompts (Llama 2)}} \\
ETTh1   &         & \textbf{0.440} & \textbf{0.450} & 0.449 & 0.457 & 0.455 & 0.458 & 0.471 & 0.469 & \underline{0.441} & \underline{0.452} \\
ETTh2   &         & \textbf{0.373} & \textbf{0.409} & \underline{0.375} & \underline{0.410} & 0.378 & 0.414 & 0.387 & 0.416 & 0.392 & 0.422 \\
\hline
$1^{st}$ count & & 2 & 2 & 0 & 0 & 0 & 0 & 0 & 0 & 0 & 0 \\
$2^{nd}$ count & & 0 & 0 & 1 & 1 & 0 & 0 & 0 & 0 & 1 & 1 \\
\midrule
\multicolumn{12}{l}{\textbf{Random Replaced Words (Llama 2)}} \\
ETTh1   &         & \textbf{0.431} & \textbf{0.447} & \underline{0.441} & \underline{0.451} & 0.461 & 0.463 & \underline{0.441} & 0.452 & 0.463 & 0.464 \\
ETTh2   &         & \textbf{0.377} & \underline{0.411} & 0.414 & 0.429 & 0.439 & 0.446 & \underline{0.378} & \textbf{0.410} & 0.393 & 0.421 \\
\hline
$1^{st}$ count & & 2 & 1 & 0 & 0 & 0 & 0 & 0 & 1 & 0 & 0 \\
$2^{nd}$ count & & 0 & 0 & 1 & 1 & 0 & 0 & 2 & 0 & 0 & 0 \\
\hline
\end{tabular}
\caption{Long-term forecasting results with random replaced prompts and random replaced words.}
\label{tab:ltf-timellm-replaced-part}
\end{table*}

\subsection{Visualization}
To explore interpretability, we visualize five modules that could be related to texts in the order from input to output.
For TimeLLM, we select two language models (GPT-2, BERT). 
For CALF, we select only GPT-2. The default method for text prototypes extraction is the linear mapping layer. The default dataset is ETTh1. For better visualization, the number of text prototypes is uniformly set to 100. 

\subsubsection{Text prototypes (V1)}

% \begin{table}[!ht]
%     \centering
%     \begin{adjustbox}{max width=0.5\textwidth}
%     \begin{tabular}{>{\centering\arraybackslash}m{0.4\linewidth}>{\centering\arraybackslash}m{0.5\linewidth}}
%     \hline
%     Baseline-LM-EM & Tokens \\
%     \hline
%     TimeLLM-GPT2-Linear& ‘ĠexternalToEVA’, ‘Ġthe’, ‘Ġand’, ‘Ġexperts’, ‘-’\\
%     \hline
%     TimeLLM-BERT-Linear& ‘[unused613]’, ‘\#\#elial’, ‘[SEP]’, ‘1738’, ‘\#\#\$’\\
%     \hline
%     CALF-GPT2-PCA& ‘Ġthe’, ‘Ġand’, ‘Ġhipp’, ‘ĠJack’, ‘Ġweaving’\\
%     \hline
%     \end{tabular}
%     \end{adjustbox}
%     \caption{Tokens corresponding to 5 of text prototypes. LM: Language model; EM: Text prototypes extraction method.}
%     \label{tab:v1}
% \end{table}
Text prototypes often serve as one of the inputs for a cross-modality module. Their own semantics, as well as the interrelationships, may influence the results. We calculate the similarity between text prototype embeddings and tokens in language models' vocabulary, and select the token with the highest similarity as the semantic meaning for that prototype (see Table \ref{tab:v1-sup} in C.1). 
If the text prototypes are not specified, its semantic meaning is likely to be an abstract token which is unrelated to the time series. Moreover, text prototypes obtained through the linear mapping layer exhibit high similarity, meaning that the model learned a multitude of text prototypes with similar semantics.

\begin{figure}[!ht]
    \centering
    \includegraphics[width=0.45\textwidth]{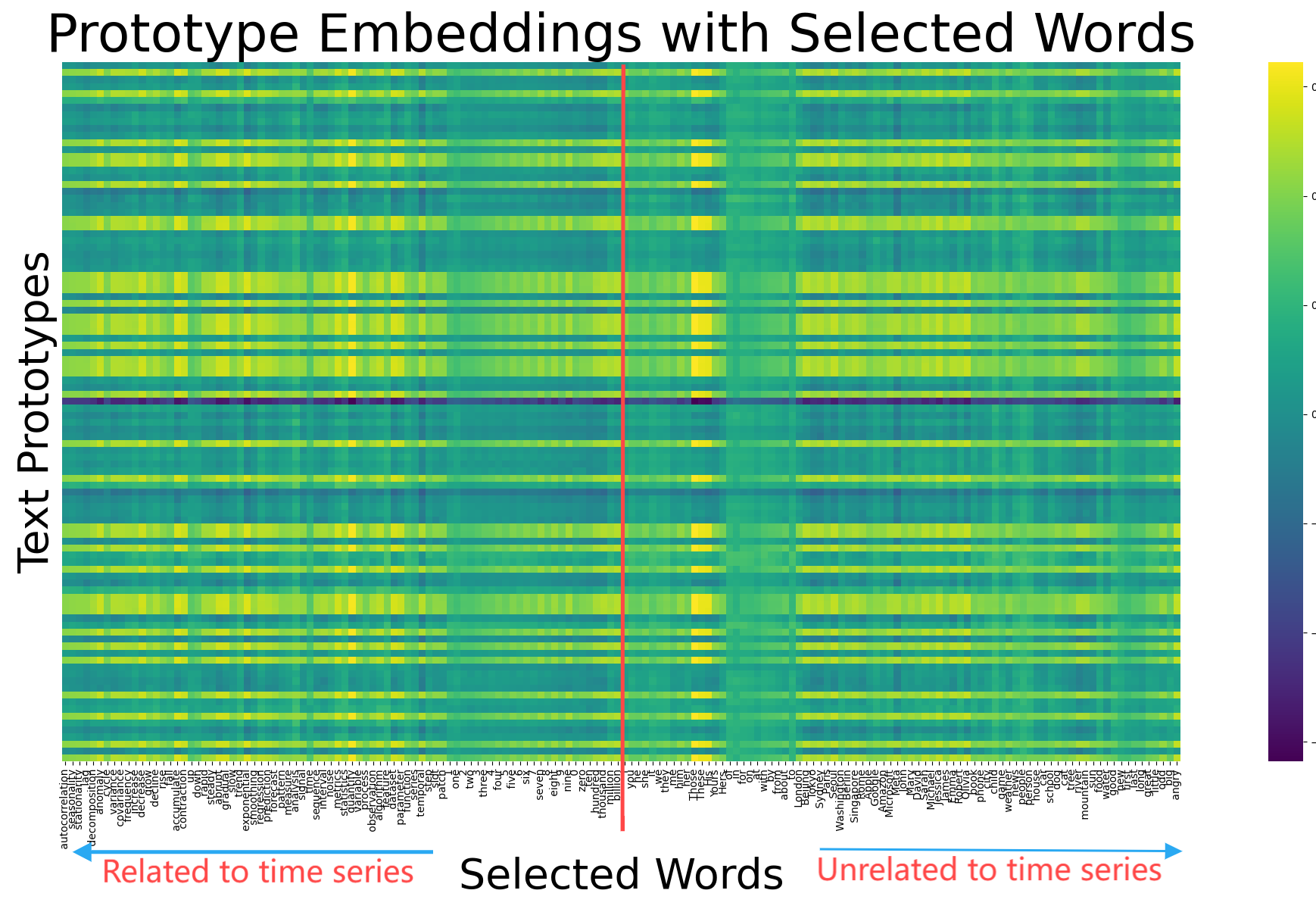}
    \caption{Similarity between text prototypes and selected words in TimeLLM.}
    \label{fig:v1-proto-emb}
\end{figure}

Considering that the learned text prototype embeddings (by linear mapping layer) may not be entirely similar to the tokens in the vocabulary, to better understand these text prototypes, we select 160 words and calculate the similarity between text prototypes and them (as Fig.\ref{fig:v1-proto-emb}, result based on BERT as Fig. \ref{fig:v1-proto-emb-bert} in C.1). The first 80 words are related to time series (such as “seasonality”, “decomposition”) while the last 80 words (such as “book”, “food”) are not (full words refer to Table \ref{tab:selected-words-r} and \ref{tab:selected-words-nr} in A.2). Fig.\ref{fig:v1-proto-emb} does not reveal some clear patterns, that is, the similarity between text prototypes and time series-related words is not significantly higher or lower than that of words unrelated to time series.

\subsubsection{Attentions of cross-modality alignment module (V2)}

\begin{figure}[!ht]
    \centering
    \includegraphics[width=0.5\textwidth]{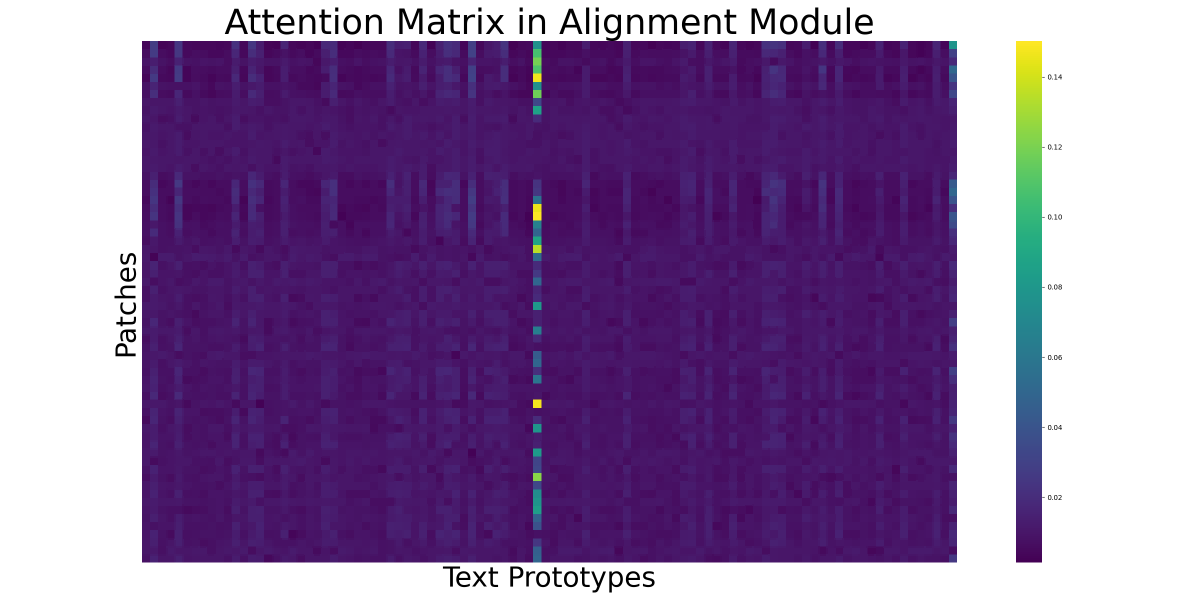}
    \caption{Attention of cross-modality alignment module in TimeLLM. }
    \label{fig:v2-att-alignment-module-timellm}
\end{figure}

The cross-modality alignment module is of vital importance in TS-LLM models, which is generally a multi-head attention mechanism which accepts data from both the time series and texts modalities \cite{jin2023time, liu2024timecma, liu2024taming}. 
The attention matrix of this module can reflect the relevance between patches (TimeLLM) or variables (CALF) and text prototypes. From Fig.\ref{fig:v2-att-alignment-module-timellm}, it can be observed that only a few text prototypes are highly related to patches and some of them appeared repeatedly. For example, the top 5 text prototypes for one patch are ‘Ġthe’, ‘ĠexternalToEVA’, ‘ÿ’, ‘ĠexternalToEVA’, ‘ÿ’. Furthermore, the token ‘Ġthe’ has the highest weights among most patches, but it doesn't make much sense. 
(The result of TimeLLM based on BERT and result of CALF can be seen from Fig.\ref{v2-timellm-bert-att-align} and \ref{v2-calf-gpt2-att-align} in C.2 respectively.)

\subsubsection{Time series embeddings after alignment (V3)}

\begin{figure}[!ht]
    \centering
    \includegraphics[width=0.45\textwidth]{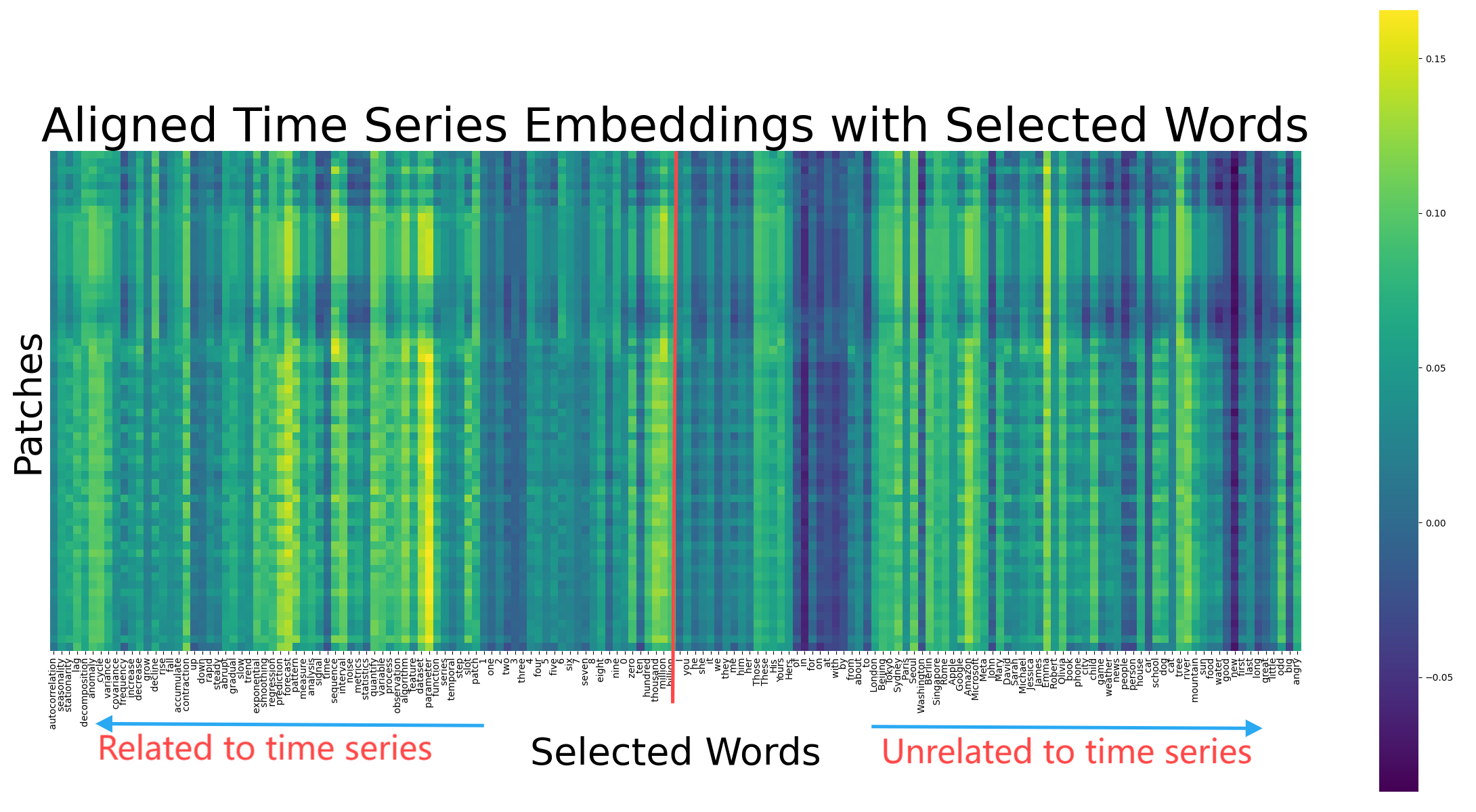}
    \caption{Similarity between aligned time series embeddings and selected words.}
    \label{fig:v3-aligned-ts-timellm}
\end{figure}

The primary purpose of the alignment module is to merge textual information to time series. So it's necessary to analyze the semantic meaning of aligned time series embeddings. If these embeddings are more consistent with some words in language models' vocabulary, they would be easier to be understood by language models. We calculate the similarity between aligned time series embeddings and tokens in the vocabulary, and select the first similar token to represent this time series embedding. So some of them can be represented as ‘Mania’, ‘Ġbeetle’, ‘ĠModule’ and ‘StreamerBot’, which are unrelated to time series. Similar to visualize text prototypes, we use the same 160 words to visualize the relation between aligned time series embeddings and specific words (as  Fig.\ref{fig:v3-aligned-ts-timellm}). From this figure, it cannot be concluded that the similarity of aligned embeddings to words related to time series is generally higher than the similarity to words unrelated to time series. For example, many of aligned embeddings are more similar to the words “parameter” and “forecast”, but not as similar to the words “new” and “big”, which meets expected interpretability. However, their similarity to the words “three” and “ten” is much lower than that to the words “Emma” and “Amazon”. (The result of TimeLLM based on BERT and result of CALF can be seen from Fig.\ref{v3-timellm-bert-align-words} and \ref{v3-calf-gpt2-align-words} in C.3 respectively.)

\subsubsection{Time series patches and corresponding textual representations (V4)}

\begin{figure}[ht!]
    \centering
    \includegraphics[width=0.5\textwidth]{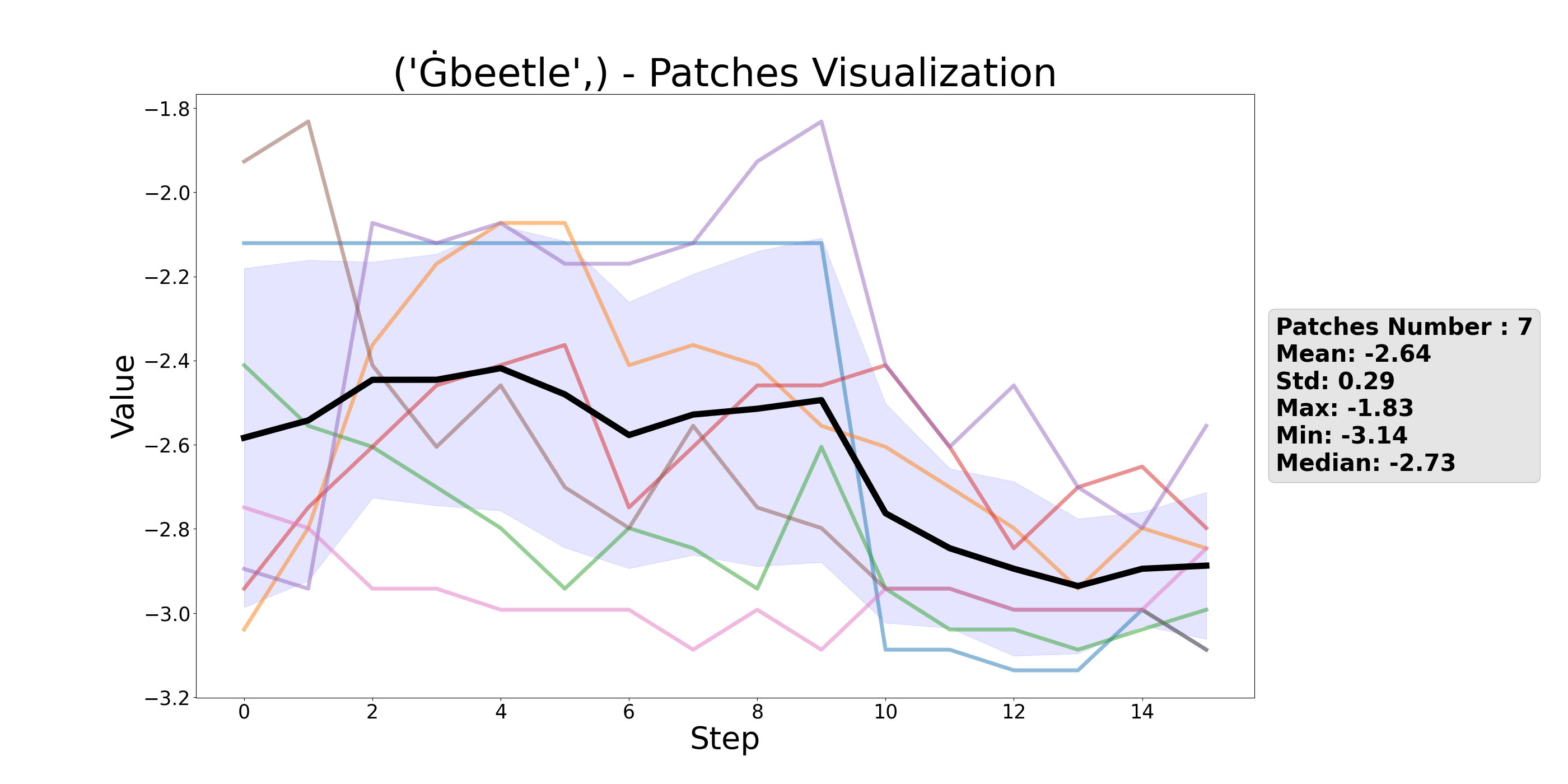}
    \caption{Patches belong to token set ‘Gbeetle’. The bold black line indicates the mean value, and the blue area represents the range within one standard deviation. More tokens sets refer to  Fig.\ref{fig:v4-gpt2} and \ref{fig:v4-bert} in Appendic C.4.}
    \label{fig:v4-token-set-1}
\end{figure}

Inspired by V3, we can try to represent aligned time series embeddings with the top $k$ tokens similar to them (the order among tokens is considered). 
If $k$ is relatively large, i.e., equal to or close to the size of the vocabulary, then the set of tokens representing each time series embedding will be distinct. Conversely, if $k$ is relatively small (less than 5), there will be instances where some time series embeddings are represented by a same set of tokens (as showed in  Fig.\ref{fig:v4-token-set-1}). Because each time series embedding is from initial time series patches, we expect time series patches corresponding to a same set of tokens possess similar statistical characteristics. 

In order to evaluate the matching degree between patches and their textual representations, we propose a novel metric named Semantic Matching Index (SMI).
Each set of tokens can be seen as a “class”. The smaller the intra-class feature distance and the larger the inter-class feature distance, the higher the index. The features selected are inspired by time series prompt from ltsm \cite{chuang2024understanding}, such as “autocorrelation”, “negative turning points” and “mean absolute differences”. 

% We define the mean $\mu_k(T_i)$ of the $k$-th feature in the $i$-th tokens set $T_i$:
% \begin{equation}
% \mu_k(T_i) = \frac{1}{|T_i|} \sum_{p \in T_i} \phi_k(p)
% \label{eq:mean}
% \end{equation}
% where $|T_i|$ denotes the number of patches belongs to this tokens set $T_i$, and $\phi_k(p)$ represents the value of the $k$-th feature for patch $p$.

% The standard deviation $\sigma_k(T_i)$ of the $k$-th feature in the $i$-th tokens set $T_i$ is defined as:
% \begin{equation}
% \sigma_k(T_i) = \sqrt{\frac{1}{|T_i|} \sum_{p \in T_i} (\phi_k(p) - \mu_k(T_i))^2}
% \label{eq:std}
% \end{equation}

The intra-class feature distance $D_{\text{intra}}$ is calculated as the sum of standard deviations for all features across all tokens sets:
\begin{equation}
D_{\text{intra}} = \sum_{i=1}^n \sum_{k=1}^m \sigma_k(T_i)
\label{eq:dwithin}
\end{equation}
where $n$ is the number of token sets and $m$ is the number of features,
$\sigma_k(T_i)$ is standard deviation the of the $k$-th feature in the $i$-th token set $T_i$.

The inter-class feature distance $D_{\text{inter}}$ is calculated as the sum of absolute differences between the means of all features across all pairs of token sets:
\begin{equation}
D_{\text{inter}} = \sum_{i=1}^{n-1} \sum_{j=i+1}^n \sum_{k=1}^m \left| \mu_k(T_i) - \mu_k(T_j) \right|
\label{eq:dbetween}
\end{equation}
where $\mu_k(T_i)$ is the mean of the $k$-th feature in the $i$-th token set $T_i$. The Semantic Matching Index (SMI) is defined as:
\begin{equation}
\text{SMI} = 
\begin{cases}
1 - e^{-b\frac{aD_{\text{inter}}}{D_{\text{intra}}}} & \text{if } D_{\text{intra}} \neq 0 \\
1 & \text{if } D_{\text{intra}} = 0
\end{cases}
\label{eq:smi}
\end{equation}
where a, b are hyperparameters and are set to 0.5, 0.1 respectively.

The range of SMI is a continuous number from 0 to 1. The value of 0 indicates that the patches within different token sets are exactly the same, suggesting that any token set does not match any specific pattern of time series; The value of 1 indicates that there is only one unique patch in each token set, suggesting that each token set can match a specific pattern of time series. Overall, the higher the SMI, the better the match between time series and token sets. 

\begin{figure}[ht!]
    \centering
    \includegraphics[width=0.45\textwidth]{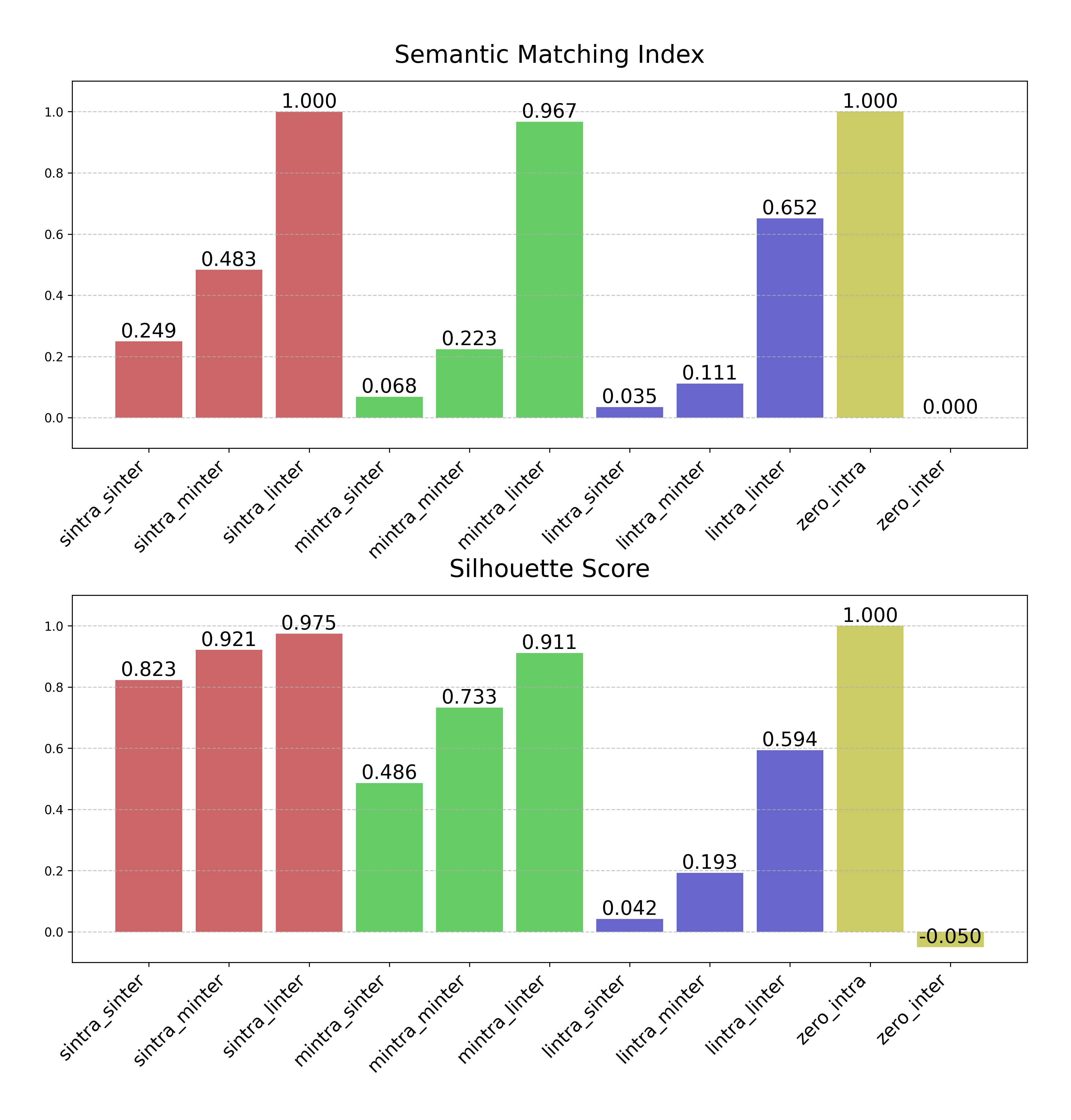}
    \caption{SMI and Silhouette Score of 11 types of time series data with different intra-class and inter-class distances.}
    \label{fig:smi_val}
\end{figure}

In order to verify the effectiveness of SMI, we divide the feature distance within and between token sets into three levels: small, median, and large, with a total of 9 combinations (for example, "sintra\_linter" means small intra-class distance, large inter-class distance) and synthesize corresponding time series data. Besides, we also add two extreme types of data: one with zero intra-class feature distance and another with zero inter-class feature distance. Fig.\ref{fig:smi_val} shows the SMI and one clustering metric Silhouette Score \cite{9260048} of all types of data. We can see that both metrics are inversely proportional to intra-class distance and directly proportional to inter-class distance, and the values of two extreme cases also meet the design goals. However, compared with Silhouette Score, SMI captures statistical features rather than general high-dimensional features, so it is more suitable for analyzing time series.

\begin{figure}[ht!]
    \centering
    \includegraphics[width=0.45\textwidth]{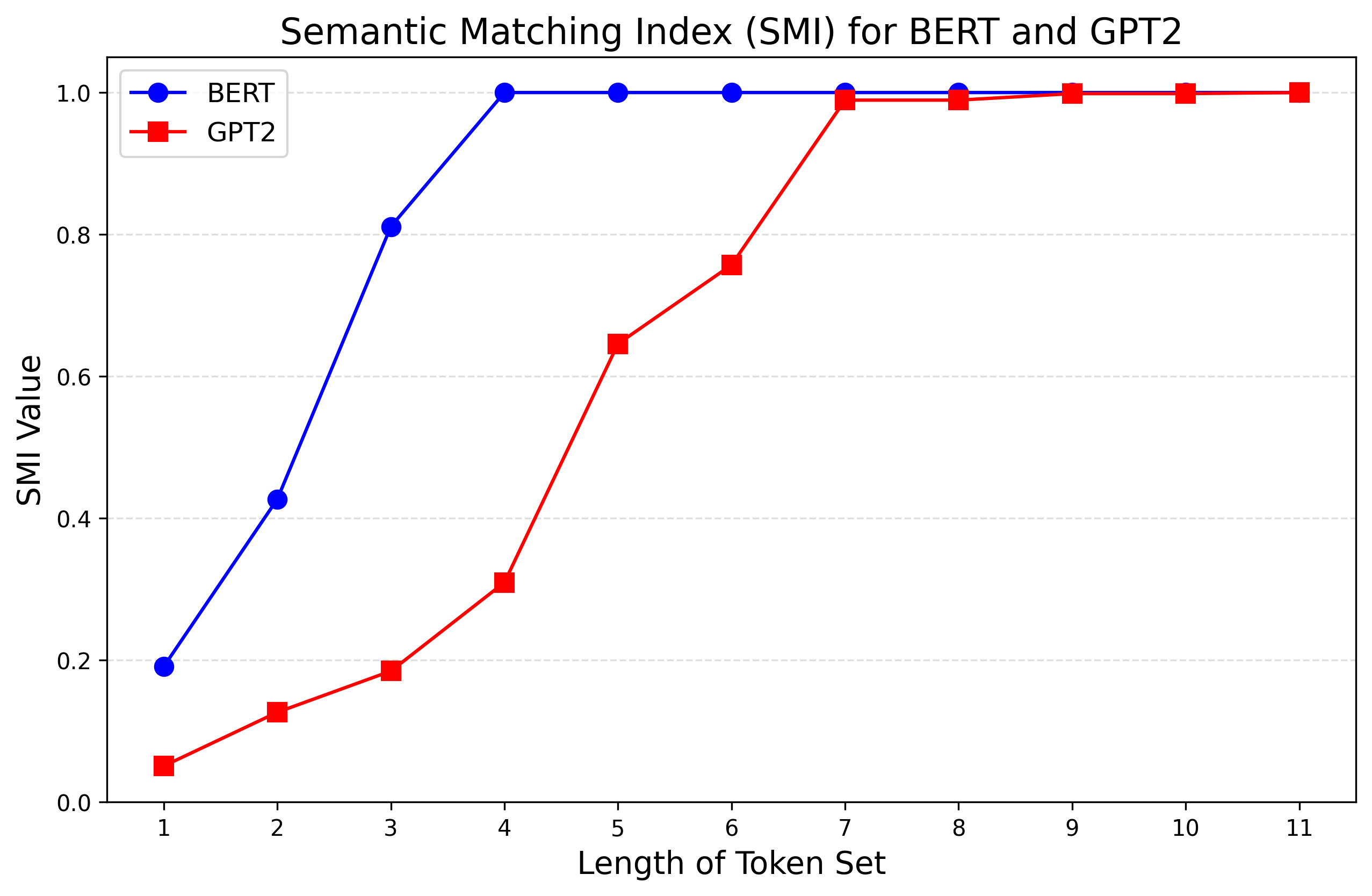}
    \caption{SMI values across different length and language models.}
    \label{fig:v4-smi}
\end{figure}

Fig.\ref{fig:v4-smi} presents the relationship between the length of the tokens set and the SMI value. It can be observed that when the length is 4, the SMI value using BERT as the language model has reached 1, indicating that using 4 tokens can completely distinguish different patches, whereas when using GPT-2 as the language model, 11 tokens are needed, and the SMI value of GPT-2 is always lower than that of BERT when the lengths are the same. It may suggest that different language models have varying abilities of understanding time series.

\subsubsection{Attentions of LLMs (V5)}

Since prompts are often concatenated with time series and then input into the language model, the relationships between patches and between patches and language tokens can be observed from the attentions of the language model.

\begin{figure}[ht!]
    \centering
    \includegraphics[width=0.5\textwidth]{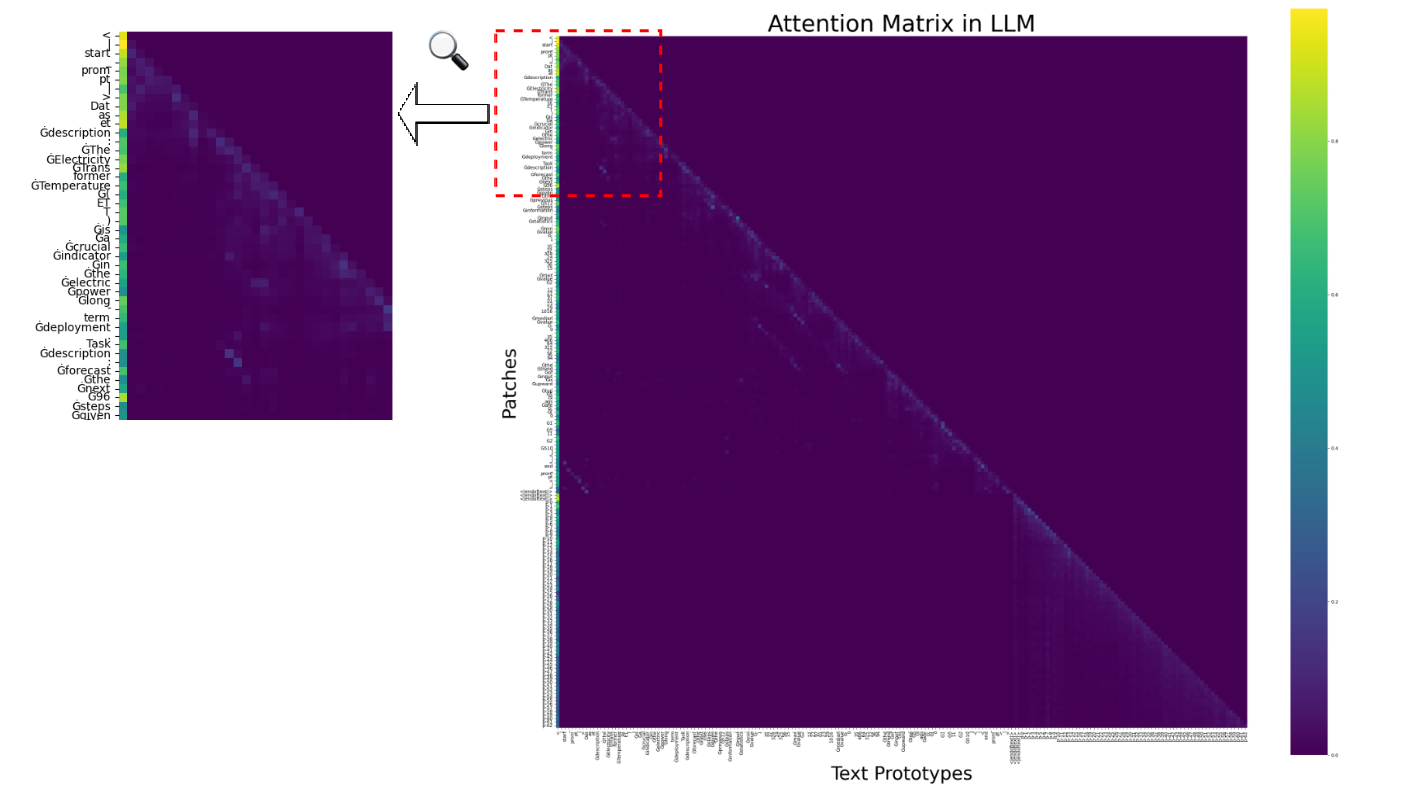}
    \caption{Attention of GPT2 in TimeLLM. }
    \label{fig:v5-att-llm-timellm}
\end{figure}

Taking the attentions in the last layer as an example (Fig.\ref{fig:v5-att-llm-timellm}), only the weights in the first column are relatively high, indicating that the pre-trained language model considered the first token “$<$” in the prompt to be the most important. But for other tokens the relationship is less clear. (The attention of BERT is shown in Fig.\ref{fig:v5-att-llm-timellm-bert} in C.5.) 
Analyzing the top 5 tokens with the highest weights to each token, it is also found that the tokens most relevant to prompts almost all belong to prompts, while the tokens most relevant to time series almost all belong to time series. These two modalities are almost only connected by the first token “$<$” (examples in Table \ref{tab:v5-tokens} in C.5). 

\section{Conclusion and future work}
In this work, we find that texts (e.g. prompts, text prototypes) in LLM-based time series models are not significantly helpful for forecasting and highly influenced by selected pretrained language models. On the basis of standard time series tasks, it still remains a challenging issue to better align time series to the semantic space of text, or to better integrate these two modalities. Potential future works can focus on 1) adding language-related tasks besides the original tasks and 2) designing cross-modalities models which can understand time series according to instructions.

%%
%% The acknowledgments section is defined using the "acks" environment
%% (and NOT an unnumbered section). This ensures the proper
%% identification of the section in the article metadata, and the
%% consistent spelling of the heading.
% \begin{acks}

% \end{acks}

%%
%% The next two lines define the bibliography style to be used, and
%% the bibliography file.
\bibliographystyle{ACM-Reference-Format}
\bibliography{sample-base}

%%% -*-BibTeX-*-
%%% Do NOT edit. File created by BibTeX with style
%%% ACM-Reference-Format-Journals [18-Jan-2012].

\begin{thebibliography}{37}

%%% ====================================================================
%%% NOTE TO THE USER: you can override these defaults by providing
%%% customized versions of any of these macros before the \bibliography
%%% command.  Each of them MUST provide its own final punctuation,
%%% except for \shownote{}, \showDOI{}, and \showURL{}.  The latter two
%%% do not use final punctuation, in order to avoid confusing it with
%%% the Web address.
%%%
%%% To suppress output of a particular field, define its macro to expand
%%% to an empty string, or better, \unskip, like this:
%%%
%%% \newcommand{\showDOI}[1]{\unskip}   % LaTeX syntax
%%%
%%% \def \showDOI #1{\unskip}           % plain TeX syntax
%%%
%%% ====================================================================

\ifx \showCODEN    \undefined \def \showCODEN     #1{\unskip}     \fi
\ifx \showDOI      \undefined \def \showDOI       #1{#1}\fi
\ifx \showISBNx    \undefined \def \showISBNx     #1{\unskip}     \fi
\ifx \showISBNxiii \undefined \def \showISBNxiii  #1{\unskip}     \fi
\ifx \showISSN     \undefined \def \showISSN      #1{\unskip}     \fi
\ifx \showLCCN     \undefined \def \showLCCN      #1{\unskip}     \fi
\ifx \shownote     \undefined \def \shownote      #1{#1}          \fi
\ifx \showarticletitle \undefined \def \showarticletitle #1{#1}   \fi
\ifx \showURL      \undefined \def \showURL       {\relax}        \fi
% The following commands are used for tagged output and should be
% invisible to TeX
\providecommand\bibfield[2]{#2}
\providecommand\bibinfo[2]{#2}
\providecommand\natexlab[1]{#1}
\providecommand\showeprint[2][]{arXiv:#2}

\bibitem[Aflalo et~al\mbox{.}(2022)]%
        {Aflalo_2022_CVPR}
\bibfield{author}{\bibinfo{person}{Estelle Aflalo}, \bibinfo{person}{Meng Du}, \bibinfo{person}{Shao-Yen Tseng}, \bibinfo{person}{Yongfei Liu}, \bibinfo{person}{Chenfei Wu}, \bibinfo{person}{Nan Duan}, {and} \bibinfo{person}{Vasudev Lal}.} \bibinfo{year}{2022}\natexlab{}.
\newblock \showarticletitle{VL-InterpreT: An Interactive Visualization Tool for Interpreting Vision-Language Transformers}. In \bibinfo{booktitle}{\emph{Proceedings of the IEEE/CVF Conference on Computer Vision and Pattern Recognition (CVPR)}}. \bibinfo{pages}{21406--21415}.
\newblock


\bibitem[Bi et~al\mbox{.}(2023)]%
        {bi2023accurate}
\bibfield{author}{\bibinfo{person}{Kaifeng Bi}, \bibinfo{person}{Lingxi Xie}, \bibinfo{person}{Hengheng Zhang}, \bibinfo{person}{Xin Chen}, \bibinfo{person}{Xiaotao Gu}, {and} \bibinfo{person}{Qi Tian}.} \bibinfo{year}{2023}\natexlab{}.
\newblock \showarticletitle{Accurate medium-range global weather forecasting with 3D neural networks}.
\newblock \bibinfo{journal}{\emph{Nature}} \bibinfo{volume}{619}, \bibinfo{number}{7970} (\bibinfo{year}{2023}), \bibinfo{pages}{533--538}.
\newblock


\bibitem[Chang et~al\mbox{.}(2023)]%
        {chang2023llm4ts}
\bibfield{author}{\bibinfo{person}{Ching Chang}, \bibinfo{person}{Wei-Yao Wang}, \bibinfo{person}{Wen-Chih Peng}, {and} \bibinfo{person}{Tien-Fu Chen}.} \bibinfo{year}{2023}\natexlab{}.
\newblock \showarticletitle{Llm4ts: Two-stage fine-tuning for time-series forecasting with pre-trained llms}.
\newblock \bibinfo{journal}{\emph{arXiv preprint arXiv:2308.08469}} (\bibinfo{year}{2023}).
\newblock


\bibitem[Chuang et~al\mbox{.}(2024)]%
        {chuang2024understanding}
\bibfield{author}{\bibinfo{person}{Yu-Neng Chuang}, \bibinfo{person}{Songchen Li}, \bibinfo{person}{Jiayi Yuan}, \bibinfo{person}{Guanchu Wang}, \bibinfo{person}{Kwei-Herng Lai}, \bibinfo{person}{Leisheng Yu}, \bibinfo{person}{Sirui Ding}, \bibinfo{person}{Chia-Yuan Chang}, \bibinfo{person}{Qiaoyu Tan}, \bibinfo{person}{Daochen Zha}, {et~al\mbox{.}}} \bibinfo{year}{2024}\natexlab{}.
\newblock \showarticletitle{Understanding Different Design Choices in Training Large Time Series Models}.
\newblock \bibinfo{journal}{\emph{arXiv preprint arXiv:2406.14045}} (\bibinfo{year}{2024}).
\newblock


\bibitem[Devlin et~al\mbox{.}(2019)]%
        {Devlin2019BERTPO}
\bibfield{author}{\bibinfo{person}{Jacob Devlin}, \bibinfo{person}{Ming-Wei Chang}, \bibinfo{person}{Kenton Lee}, {and} \bibinfo{person}{Kristina Toutanova}.} \bibinfo{year}{2019}\natexlab{}.
\newblock \showarticletitle{BERT: Pre-training of Deep Bidirectional Transformers for Language Understanding}. In \bibinfo{booktitle}{\emph{North American Chapter of the Association for Computational Linguistics}}.
\newblock
\urldef\tempurl%
\url{https://api.semanticscholar.org/CorpusID:52967399}
\showURL{%
\tempurl}


\bibitem[Hu et~al\mbox{.}(2021)]%
        {hu2021lora}
\bibfield{author}{\bibinfo{person}{Edward~J Hu}, \bibinfo{person}{Yelong Shen}, \bibinfo{person}{Phillip Wallis}, \bibinfo{person}{Zeyuan Allen-Zhu}, \bibinfo{person}{Yuanzhi Li}, \bibinfo{person}{Shean Wang}, \bibinfo{person}{Lu Wang}, {and} \bibinfo{person}{Weizhu Chen}.} \bibinfo{year}{2021}\natexlab{}.
\newblock \showarticletitle{Lora: Low-rank adaptation of large language models}.
\newblock \bibinfo{journal}{\emph{arXiv preprint arXiv:2106.09685}} (\bibinfo{year}{2021}).
\newblock


\bibitem[Ji et~al\mbox{.}(2023)]%
        {DBLP:journals/corr/abs-2310-19852}
\bibfield{author}{\bibinfo{person}{Jiaming Ji}, \bibinfo{person}{Tianyi Qiu}, \bibinfo{person}{Boyuan Chen}, \bibinfo{person}{Borong Zhang}, \bibinfo{person}{Hantao Lou}, \bibinfo{person}{Kaile Wang}, \bibinfo{person}{Yawen Duan}, \bibinfo{person}{Zhonghao He}, \bibinfo{person}{Jiayi Zhou}, \bibinfo{person}{Zhaowei Zhang}, \bibinfo{person}{Fanzhi Zeng}, \bibinfo{person}{Kwan~Yee Ng}, \bibinfo{person}{Juntao Dai}, \bibinfo{person}{Xuehai Pan}, \bibinfo{person}{Aidan O'Gara}, \bibinfo{person}{Yingshan Lei}, \bibinfo{person}{Hua Xu}, \bibinfo{person}{Brian Tse}, \bibinfo{person}{Jie Fu}, \bibinfo{person}{Stephen McAleer}, \bibinfo{person}{Yaodong Yang}, \bibinfo{person}{Yizhou Wang}, \bibinfo{person}{Song{-}Chun Zhu}, \bibinfo{person}{Yike Guo}, {and} \bibinfo{person}{Wen Gao}.} \bibinfo{year}{2023}\natexlab{}.
\newblock \showarticletitle{{AI} Alignment: {A} Comprehensive Survey}.
\newblock \bibinfo{journal}{\emph{CoRR}}  \bibinfo{volume}{abs/2310.19852} (\bibinfo{year}{2023}).
\newblock
\urldef\tempurl%
\url{https://doi.org/10.48550/ARXIV.2310.19852}
\showDOI{\tempurl}
\showeprint[arXiv]{2310.19852}


\bibitem[Jiang et~al\mbox{.}(2024)]%
        {jiang2024empowering}
\bibfield{author}{\bibinfo{person}{Yushan Jiang}, \bibinfo{person}{Zijie Pan}, \bibinfo{person}{Xikun Zhang}, \bibinfo{person}{Sahil Garg}, \bibinfo{person}{Anderson Schneider}, \bibinfo{person}{Yuriy Nevmyvaka}, {and} \bibinfo{person}{Dongjin Song}.} \bibinfo{year}{2024}\natexlab{}.
\newblock \bibinfo{title}{Empowering Time Series Analysis with Large Language Models: A Survey}.
\newblock
\newblock
\showeprint[arxiv]{2402.03182}~[cs.LG]


\bibitem[Jin et~al\mbox{.}(2024)]%
        {jin2023time}
\bibfield{author}{\bibinfo{person}{Ming Jin}, \bibinfo{person}{Shiyu Wang}, \bibinfo{person}{Lintao Ma}, \bibinfo{person}{Zhixuan Chu}, \bibinfo{person}{James~Y Zhang}, \bibinfo{person}{Xiaoming Shi}, \bibinfo{person}{Pin-Yu Chen}, \bibinfo{person}{Yuxuan Liang}, \bibinfo{person}{Yuan-Fang Li}, \bibinfo{person}{Shirui Pan}, {and} \bibinfo{person}{Qingsong Wen}.} \bibinfo{year}{2024}\natexlab{}.
\newblock \showarticletitle{{Time-LLM}: Time series forecasting by reprogramming large language models}. In \bibinfo{booktitle}{\emph{International Conference on Learning Representations (ICLR)}}.
\newblock


\bibitem[Lin et~al\mbox{.}(2023)]%
        {10260564}
\bibfield{author}{\bibinfo{person}{Junpeng Lin}, \bibinfo{person}{Ziyue Li}, \bibinfo{person}{Zhishuai Li}, \bibinfo{person}{Lei Bai}, \bibinfo{person}{Rui Zhao}, {and} \bibinfo{person}{Chen Zhang}.} \bibinfo{year}{2023}\natexlab{}.
\newblock \showarticletitle{Dynamic Causal Graph Convolutional Network for Traffic Prediction}. In \bibinfo{booktitle}{\emph{2023 IEEE 19th International Conference on Automation Science and Engineering (CASE)}}. \bibinfo{pages}{1--8}.
\newblock
\urldef\tempurl%
\url{https://doi.org/10.1109/CASE56687.2023.10260564}
\showDOI{\tempurl}


\bibitem[Liu et~al\mbox{.}(2024c)]%
        {liu2024timecma}
\bibfield{author}{\bibinfo{person}{Chenxi Liu}, \bibinfo{person}{Qianxiong Xu}, \bibinfo{person}{Hao Miao}, \bibinfo{person}{Sun Yang}, \bibinfo{person}{Lingzheng Zhang}, \bibinfo{person}{Cheng Long}, \bibinfo{person}{Ziyue Li}, {and} \bibinfo{person}{Rui Zhao}.} \bibinfo{year}{2024}\natexlab{c}.
\newblock \showarticletitle{TimeCMA: Towards LLM-Empowered Time Series Forecasting via Cross-Modality Alignment}.
\newblock \bibinfo{journal}{\emph{arXiv preprint arXiv:2406.01638}} (\bibinfo{year}{2024}).
\newblock


\bibitem[Liu et~al\mbox{.}(2024d)]%
        {liu2024large}
\bibfield{author}{\bibinfo{person}{Jun Liu}, \bibinfo{person}{Chaoyun Zhang}, \bibinfo{person}{Jiaxu Qian}, \bibinfo{person}{Minghua Ma}, \bibinfo{person}{Si Qin}, \bibinfo{person}{Chetan Bansal}, \bibinfo{person}{Qingwei Lin}, \bibinfo{person}{Saravan Rajmohan}, {and} \bibinfo{person}{Dongmei Zhang}.} \bibinfo{year}{2024}\natexlab{d}.
\newblock \showarticletitle{Large Language Models can Deliver Accurate and Interpretable Time Series Anomaly Detection}.
\newblock \bibinfo{journal}{\emph{arXiv preprint arXiv:2405.15370}} (\bibinfo{year}{2024}).
\newblock


\bibitem[Liu et~al\mbox{.}(2024a)]%
        {liu2024taming}
\bibfield{author}{\bibinfo{person}{Peiyuan Liu}, \bibinfo{person}{Hang Guo}, \bibinfo{person}{Tao Dai}, \bibinfo{person}{Naiqi Li}, \bibinfo{person}{Jigang Bao}, \bibinfo{person}{Xudong Ren}, \bibinfo{person}{Yong Jiang}, {and} \bibinfo{person}{Shu-Tao Xia}.} \bibinfo{year}{2024}\natexlab{a}.
\newblock \showarticletitle{Taming Pre-trained LLMs for Generalised Time Series Forecasting via Cross-modal Knowledge Distillation}.
\newblock \bibinfo{journal}{\emph{arXiv preprint arXiv:2403.07300}} (\bibinfo{year}{2024}).
\newblock


\bibitem[Liu et~al\mbox{.}(2023)]%
        {liu2023financial}
\bibfield{author}{\bibinfo{person}{Shun Liu}, \bibinfo{person}{Kexin Wu}, \bibinfo{person}{Chufeng Jiang}, \bibinfo{person}{Bin Huang}, {and} \bibinfo{person}{Danqing Ma}.} \bibinfo{year}{2023}\natexlab{}.
\newblock \showarticletitle{Financial time-series forecasting: Towards synergizing performance and interpretability within a hybrid machine learning approach}.
\newblock \bibinfo{journal}{\emph{arXiv preprint arXiv:2401.00534}} (\bibinfo{year}{2023}).
\newblock


\bibitem[Liu et~al\mbox{.}(2024b)]%
        {liu2024unitime}
\bibfield{author}{\bibinfo{person}{Xu Liu}, \bibinfo{person}{Junfeng Hu}, \bibinfo{person}{Yuan Li}, \bibinfo{person}{Shizhe Diao}, \bibinfo{person}{Yuxuan Liang}, \bibinfo{person}{Bryan Hooi}, {and} \bibinfo{person}{Roger Zimmermann}.} \bibinfo{year}{2024}\natexlab{b}.
\newblock \showarticletitle{Unitime: A language-empowered unified model for cross-domain time series forecasting}. In \bibinfo{booktitle}{\emph{Proceedings of the ACM on Web Conference 2024}}. \bibinfo{pages}{4095--4106}.
\newblock


\bibitem[Miller(2019)]%
        {DBLP:journals/ai/Miller19}
\bibfield{author}{\bibinfo{person}{Tim Miller}.} \bibinfo{year}{2019}\natexlab{}.
\newblock \showarticletitle{Explanation in artificial intelligence: Insights from the social sciences}.
\newblock \bibinfo{journal}{\emph{Artif. Intell.}}  \bibinfo{volume}{267} (\bibinfo{year}{2019}), \bibinfo{pages}{1--38}.
\newblock
\urldef\tempurl%
\url{https://doi.org/10.1016/J.ARTINT.2018.07.007}
\showDOI{\tempurl}


\bibitem[Morid et~al\mbox{.}(2023)]%
        {morid2023time}
\bibfield{author}{\bibinfo{person}{Mohammad~Amin Morid}, \bibinfo{person}{Olivia R~Liu Sheng}, {and} \bibinfo{person}{Joseph Dunbar}.} \bibinfo{year}{2023}\natexlab{}.
\newblock \showarticletitle{Time series prediction using deep learning methods in healthcare}.
\newblock \bibinfo{journal}{\emph{ACM Transactions on Management Information Systems}} \bibinfo{volume}{14}, \bibinfo{number}{1} (\bibinfo{year}{2023}), \bibinfo{pages}{1--29}.
\newblock


\bibitem[Nie et~al\mbox{.}(2023)]%
        {Yuqietal-2023-PatchTST}
\bibfield{author}{\bibinfo{person}{Yuqi Nie}, \bibinfo{person}{Nam H.~Nguyen}, \bibinfo{person}{Phanwadee Sinthong}, {and} \bibinfo{person}{Jayant Kalagnanam}.} \bibinfo{year}{2023}\natexlab{}.
\newblock \showarticletitle{A Time Series is Worth 64 Words: Long-term Forecasting with Transformers}. In \bibinfo{booktitle}{\emph{International Conference on Learning Representations}}.
\newblock


\bibitem[Palit et~al\mbox{.}(2023)]%
        {Palit_2023_ICCV}
\bibfield{author}{\bibinfo{person}{Vedant Palit}, \bibinfo{person}{Rohan Pandey}, \bibinfo{person}{Aryaman Arora}, {and} \bibinfo{person}{Paul~Pu Liang}.} \bibinfo{year}{2023}\natexlab{}.
\newblock \showarticletitle{Towards Vision-Language Mechanistic Interpretability: A Causal Tracing Tool for BLIP}. In \bibinfo{booktitle}{\emph{Proceedings of the IEEE/CVF International Conference on Computer Vision (ICCV) Workshops}}. \bibinfo{pages}{2856--2861}.
\newblock


\bibitem[Pan et~al\mbox{.}(2024)]%
        {pan2024textbf}
\bibfield{author}{\bibinfo{person}{Zijie Pan}, \bibinfo{person}{Yushan Jiang}, \bibinfo{person}{Sahil Garg}, \bibinfo{person}{Anderson Schneider}, \bibinfo{person}{Yuriy Nevmyvaka}, {and} \bibinfo{person}{Dongjin Song}.} \bibinfo{year}{2024}\natexlab{}.
\newblock \showarticletitle{S2IP-LLM: Semantic Space Informed Prompt Learning with LLM for Time Series Forecasting}.
\newblock \bibinfo{journal}{\emph{arXiv preprint arXiv:2403.05798}} (\bibinfo{year}{2024}).
\newblock


\bibitem[Radford et~al\mbox{.}(2019)]%
        {radford2019language}
\bibfield{author}{\bibinfo{person}{Alec Radford}, \bibinfo{person}{Jeffrey Wu}, \bibinfo{person}{Rewon Child}, \bibinfo{person}{David Luan}, \bibinfo{person}{Dario Amodei}, \bibinfo{person}{Ilya Sutskever}, {et~al\mbox{.}}} \bibinfo{year}{2019}\natexlab{}.
\newblock \showarticletitle{Language models are unsupervised multitask learners}.
\newblock \bibinfo{journal}{\emph{OpenAI blog}} \bibinfo{volume}{1}, \bibinfo{number}{8} (\bibinfo{year}{2019}), \bibinfo{pages}{9}.
\newblock


\bibitem[Sanh et~al\mbox{.}(2019)]%
        {sanh2019distilbert}
\bibfield{author}{\bibinfo{person}{Victor Sanh}, \bibinfo{person}{Lysandre Debut}, \bibinfo{person}{Julien Chaumond}, {and} \bibinfo{person}{Thomas Wolf}.} \bibinfo{year}{2019}\natexlab{}.
\newblock \showarticletitle{DistilBERT, a distilled version of BERT: smaller, faster, cheaper and lighter}.
\newblock \bibinfo{journal}{\emph{arXiv preprint arXiv:1910.01108}} (\bibinfo{year}{2019}).
\newblock


\bibitem[Shahapure and Nicholas(2020)]%
        {9260048}
\bibfield{author}{\bibinfo{person}{Ketan~Rajshekhar Shahapure} {and} \bibinfo{person}{Charles Nicholas}.} \bibinfo{year}{2020}\natexlab{}.
\newblock \showarticletitle{Cluster Quality Analysis Using Silhouette Score}. In \bibinfo{booktitle}{\emph{2020 IEEE 7th International Conference on Data Science and Advanced Analytics (DSAA)}}. \bibinfo{pages}{747--748}.
\newblock
\urldef\tempurl%
\url{https://doi.org/10.1109/DSAA49011.2020.00096}
\showDOI{\tempurl}


\bibitem[Shao et~al\mbox{.}(2022)]%
        {shao2022spatial}
\bibfield{author}{\bibinfo{person}{Zezhi Shao}, \bibinfo{person}{Zhao Zhang}, \bibinfo{person}{Fei Wang}, \bibinfo{person}{Wei Wei}, {and} \bibinfo{person}{Yongjun Xu}.} \bibinfo{year}{2022}\natexlab{}.
\newblock \showarticletitle{Spatial-temporal identity: A simple yet effective baseline for multivariate time series forecasting}. In \bibinfo{booktitle}{\emph{Proceedings of the 31st ACM International Conference on Information \& Knowledge Management}}. \bibinfo{pages}{4454--4458}.
\newblock


\bibitem[Sun et~al\mbox{.}(2023)]%
        {sun2023test}
\bibfield{author}{\bibinfo{person}{Chenxi Sun}, \bibinfo{person}{Yaliang Li}, \bibinfo{person}{Hongyan Li}, {and} \bibinfo{person}{Shenda Hong}.} \bibinfo{year}{2023}\natexlab{}.
\newblock \showarticletitle{TEST: Text prototype aligned embedding to activate LLM's ability for time series}.
\newblock \bibinfo{journal}{\emph{arXiv preprint arXiv:2308.08241}} (\bibinfo{year}{2023}).
\newblock


\bibitem[Tan et~al\mbox{.}(2024)]%
        {DBLP:journals/corr/abs-2406-16964}
\bibfield{author}{\bibinfo{person}{Mingtian Tan}, \bibinfo{person}{Mike~A. Merrill}, \bibinfo{person}{Vinayak Gupta}, \bibinfo{person}{Tim Althoff}, {and} \bibinfo{person}{Thomas Hartvigsen}.} \bibinfo{year}{2024}\natexlab{}.
\newblock \showarticletitle{Are Language Models Actually Useful for Time Series Forecasting?}
\newblock \bibinfo{journal}{\emph{CoRR}}  \bibinfo{volume}{abs/2406.16964} (\bibinfo{year}{2024}).
\newblock
\urldef\tempurl%
\url{https://doi.org/10.48550/ARXIV.2406.16964}
\showDOI{\tempurl}
\showeprint[arXiv]{2406.16964}


\bibitem[Touvron et~al\mbox{.}(2023)]%
        {touvron2023llama}
\bibfield{author}{\bibinfo{person}{Hugo Touvron}, \bibinfo{person}{Louis Martin}, \bibinfo{person}{Kevin Stone}, \bibinfo{person}{Peter Albert}, \bibinfo{person}{Amjad Almahairi}, \bibinfo{person}{Yasmine Babaei}, \bibinfo{person}{Nikolay Bashlykov}, \bibinfo{person}{Soumya Batra}, \bibinfo{person}{Prajjwal Bhargava}, \bibinfo{person}{Shruti Bhosale}, {et~al\mbox{.}}} \bibinfo{year}{2023}\natexlab{}.
\newblock \showarticletitle{Llama 2: Open foundation and fine-tuned chat models}.
\newblock \bibinfo{journal}{\emph{arXiv preprint arXiv:2307.09288}} (\bibinfo{year}{2023}).
\newblock


\bibitem[Vaswani et~al\mbox{.}(2017)]%
        {vaswani2017attention}
\bibfield{author}{\bibinfo{person}{Ashish Vaswani}, \bibinfo{person}{Noam Shazeer}, \bibinfo{person}{Niki Parmar}, \bibinfo{person}{Jakob Uszkoreit}, \bibinfo{person}{Llion Jones}, \bibinfo{person}{Aidan~N Gomez}, \bibinfo{person}{{\L}ukasz Kaiser}, {and} \bibinfo{person}{Illia Polosukhin}.} \bibinfo{year}{2017}\natexlab{}.
\newblock \showarticletitle{Attention is all you need}.
\newblock \bibinfo{journal}{\emph{Advances in neural information processing systems}}  \bibinfo{volume}{30} (\bibinfo{year}{2017}).
\newblock


\bibitem[Wang et~al\mbox{.}(2024)]%
        {wang2024language}
\bibfield{author}{\bibinfo{person}{Ning Wang}, \bibinfo{person}{Guangming Zhu}, \bibinfo{person}{HS Li}, \bibinfo{person}{Liang Zhang}, \bibinfo{person}{Syed Afaq~Ali Shah}, {and} \bibinfo{person}{Mohammed Bennamoun}.} \bibinfo{year}{2024}\natexlab{}.
\newblock \showarticletitle{Language Model Guided Interpretable Video Action Reasoning}. In \bibinfo{booktitle}{\emph{Proceedings of the IEEE/CVF Conference on Computer Vision and Pattern Recognition}}. \bibinfo{pages}{18878--18887}.
\newblock


\bibitem[Wen et~al\mbox{.}(2023)]%
        {wen2023transformers}
\bibfield{author}{\bibinfo{person}{Qingsong Wen}, \bibinfo{person}{Tian Zhou}, \bibinfo{person}{Chaoli Zhang}, \bibinfo{person}{Weiqi Chen}, \bibinfo{person}{Ziqing Ma}, \bibinfo{person}{Junchi Yan}, {and} \bibinfo{person}{Liang Sun}.} \bibinfo{year}{2023}\natexlab{}.
\newblock \showarticletitle{Transformers in time series: A survey}. In \bibinfo{booktitle}{\emph{International Joint Conference on Artificial Intelligence(IJCAI)}}.
\newblock


\bibitem[Wu et~al\mbox{.}(2023)]%
        {wu2023timesnet}
\bibfield{author}{\bibinfo{person}{Haixu Wu}, \bibinfo{person}{Tengge Hu}, \bibinfo{person}{Yong Liu}, \bibinfo{person}{Hang Zhou}, \bibinfo{person}{Jianmin Wang}, {and} \bibinfo{person}{Mingsheng Long}.} \bibinfo{year}{2023}\natexlab{}.
\newblock \showarticletitle{TimesNet: Temporal 2D-Variation Modeling for General Time Series Analysis}. In \bibinfo{booktitle}{\emph{International Conference on Learning Representations}}.
\newblock


\bibitem[Yang and Hong(2022)]%
        {yang2022unsupervised}
\bibfield{author}{\bibinfo{person}{Ling Yang} {and} \bibinfo{person}{Shenda Hong}.} \bibinfo{year}{2022}\natexlab{}.
\newblock \showarticletitle{Unsupervised time-series representation learning with iterative bilinear temporal-spectral fusion}. In \bibinfo{booktitle}{\emph{International conference on machine learning}}. PMLR, \bibinfo{pages}{25038--25054}.
\newblock


\bibitem[Yu et~al\mbox{.}(2023)]%
        {yu2023harnessing}
\bibfield{author}{\bibinfo{person}{Xinli Yu}, \bibinfo{person}{Zheng Chen}, {and} \bibinfo{person}{Yanbin Lu}.} \bibinfo{year}{2023}\natexlab{}.
\newblock \showarticletitle{Harnessing LLMs for temporal data-a study on explainable financial time series forecasting}. In \bibinfo{booktitle}{\emph{Proceedings of the 2023 Conference on Empirical Methods in Natural Language Processing: Industry Track}}. \bibinfo{pages}{739--753}.
\newblock


\bibitem[Zhang et~al\mbox{.}(2022)]%
        {zhang2022opt}
\bibfield{author}{\bibinfo{person}{Susan Zhang}, \bibinfo{person}{Stephen Roller}, \bibinfo{person}{Naman Goyal}, \bibinfo{person}{Mikel Artetxe}, \bibinfo{person}{Moya Chen}, \bibinfo{person}{Shuohui Chen}, \bibinfo{person}{Christopher Dewan}, \bibinfo{person}{Mona Diab}, \bibinfo{person}{Xian Li}, \bibinfo{person}{Xi~Victoria Lin}, {et~al\mbox{.}}} \bibinfo{year}{2022}\natexlab{}.
\newblock \showarticletitle{Opt: Open pre-trained transformer language models}.
\newblock \bibinfo{journal}{\emph{arXiv preprint arXiv:2205.01068}} (\bibinfo{year}{2022}).
\newblock


\bibitem[Zhang et~al\mbox{.}(2024)]%
        {zhang2024large}
\bibfield{author}{\bibinfo{person}{Xiyuan Zhang}, \bibinfo{person}{Ranak~Roy Chowdhury}, \bibinfo{person}{Rajesh~K Gupta}, {and} \bibinfo{person}{Jingbo Shang}.} \bibinfo{year}{2024}\natexlab{}.
\newblock \showarticletitle{Large Language Models for Time Series: A Survey}.
\newblock \bibinfo{journal}{\emph{arXiv preprint arXiv:2402.01801}} (\bibinfo{year}{2024}).
\newblock


\bibitem[Zhou et~al\mbox{.}(2021)]%
        {zhou2021informer}
\bibfield{author}{\bibinfo{person}{Haoyi Zhou}, \bibinfo{person}{Shanghang Zhang}, \bibinfo{person}{Jieqi Peng}, \bibinfo{person}{Shuai Zhang}, \bibinfo{person}{Jianxin Li}, \bibinfo{person}{Hui Xiong}, {and} \bibinfo{person}{Wancai Zhang}.} \bibinfo{year}{2021}\natexlab{}.
\newblock \showarticletitle{Informer: Beyond efficient transformer for long sequence time-series forecasting}. In \bibinfo{booktitle}{\emph{Proceedings of the AAAI conference on artificial intelligence}}.
\newblock


\bibitem[Zhou et~al\mbox{.}(2023)]%
        {zhou2023one}
\bibfield{author}{\bibinfo{person}{Tian Zhou}, \bibinfo{person}{Peisong Niu}, \bibinfo{person}{Liang Sun}, \bibinfo{person}{Rong Jin}, {et~al\mbox{.}}} \bibinfo{year}{2023}\natexlab{}.
\newblock \showarticletitle{One fits all: Power general time series analysis by pretrained lm}.
\newblock \bibinfo{journal}{\emph{Advances in neural information processing systems}}  \bibinfo{volume}{36} (\bibinfo{year}{2023}), \bibinfo{pages}{43322--43355}.
\newblock


\end{thebibliography}

%%
%% If your work has an appendix, this is the place to put it.
\appendix
\onecolumn

\section{Detailed Experimental Settings}

\subsection{Overview Configuration}
Experimental configurations are summarized in Table \ref{tab:params}. 

\begin{table}[!htbp]
\centering
\begin{adjustbox}{max width=\textwidth}
\begin{tabular}{c|c|c|c|c|c|c|c|c|c}
\toprule
\multirow{2}{*}{Model-Task-Dataset / Configuration} & \multicolumn{5}{c|}{Model Hyperparameter}& \multicolumn{4}{c}{Training Process} \\
\cline{2-10}
 & Num. of Text Prototypes & Language Models & Input Length & Patch Dim. & Heads & LR & Batch Size & Epochs & Patience \\
\midrule
 TimeLLM*-LTF-ETT& 1000& GPT2,BERT*,OPT*,Llama2*& 512& 16& 8& 1e-3& 62\&16& 100&10\\ 
 TimeLLM*-LTF-Electricity& 1000& GPT2& 512& 16& 8& 1e-2& 24& 100&5\\ 
 TimeLLM*-LTF-Weather& 1000& GPT2& 512& 16& 8& 1e-2& 24& 100&5\\ 
 TimeLLM*-LTF-Traffic& 1000& GPT2& 512& 16& 8& 1e-2& 24& 10&5\\ 
 TimeLLM*-FSF-ETT& 1000& GPT2& 512& 16& 8& 1e-3& 62& 100&10\\ 
 TimeLLM*-ZSF-ETT& 1000& GPT2& 512& 16& 8& 1e-3& 62& 10&5\\
 TimeLLM-Visualization-ETTh1& 100& GPT2,BERT2& 512& 16& 8& 1e-3& 62& 100&10\\ 
 CALF*-LTF-ETT& 500& GPT2& 96& -& 4& 5e-4& 62& 100&10\\ 
 CALF*-LTF-Electricity& 500& GPT2& 96& -& 4& 5e-4& 32& 20&5\\  
 CALF*-LTF-Weather& 500& GPT2& 96& -& 4& 5e-4& 64& 100&5\\ 
 CALF*-LTF-Traffic& 500& GPT2& 96& -& 4& 5e-4& 8& 10&5\\  
 CALF-Visualization-ETTh1& 100& GPT2& 96& -& 4& 5e-4& 62& 100&10\\
\bottomrule

\end{tabular}
\end{adjustbox}
\caption{Overview configurations of experiments for two models. TimeLLM*: TimeLLM and its ablations; CALF*: CALF and its ablations; BERT*: BERT and Distil BERT; OPT*: OPT-125M and OPT-350M; Llama2*: only first 8 layers are selected; LTF: long-term forecasting; FSF: few-shot forecasting; ZSF: zero-shot forecasting; LR: learning rate.}
\label{tab:params}
\end{table}

\subsection{Selected Words for Visualization}
The words related to time series are divided into four categories, which are “Characteristics”, “Changes”, “Degree of changes”, “Number” and “Others” (as Table \ref{tab:selected-words-r}). The words unrelated to time series can be categorized into seven categories, including “Pronouns”, “Prepositions”, “Cities”, “Companies”, “Common Names”, “Common Nouns”, and “Common Adjectives” (as Table \ref{tab:selected-words-nr}).

\begin{table}[!htbp]
    \centering
    \begin{adjustbox}{max width=\textwidth}
    \begin{tabular}{>{\centering\arraybackslash}m{0.3\linewidth}>{\centering\arraybackslash}m{0.7\linewidth}}
    \hline
    Category & Words\\
    \hline
    Characteristics (x10) & “autocorrelation” “seasonality” “stationarity” “lag” “decomposition” “anomaly” “cycle” “variance” “covariance” “frequency” \\
    \hline
    Changes (x10) & “increase” “decrease” “grow” “decline” “rise” “fall” “accumulate” “contraction” “up” “down” \\
    \hline
    Degree of changes (x5) & “rapid” “steady” “abrupt” “gradual” “slow” \\
    \hline
    Number (x25) & “1” “one” “2” “two” “3” “three” “4” “four” “5” “five” “6” “six” “7” “seven” “8” “eight” “9” “nine” “0” “zero” “ten” “hundred” “thousand” “million” “billion" \\
    \hline
    Others (x30) & “trend” “exponential” “smoothing” “regression” “prediction” “forecast” “pattern” “measure” “analysis” “signal” “time” “sequence” “interval” “noise” “metrics” “statistics” “quantify” “variable” “process” “observation” “algorithm” “feature” “dataset” “parameter” “function” “series” “temporal” “step” “from" “to"\\
    \hline
    \end{tabular}
    \end{adjustbox}
    \caption{Selected words related to time series}
    \label{tab:selected-words-r}
\end{table}

\begin{table}[!htbp]
    \centering
    \begin{adjustbox}{max width=\textwidth}
    \begin{tabular}{>{\centering\arraybackslash}m{0.3\linewidth}>{\centering\arraybackslash}m{0.7\linewidth}}
    \hline
    Category & Words\\
    \hline
    Pronouns(x15)& “I" “you” “he” “she” “it” “we” “they” “me” “him” “her” “Those” “These” “His" “Yours" “Hers"\\
    \hline
    Prepositions(x10)& “of” “in” “for” “on” “at” “with” “by” “from” “about” “to"\\
    \hline
    Cities(x10)& “London” “Beijing” “Tokyo” “Sydney” “Paris” “Seoul” “Washington” “Berlin” “Singapore” “Rome"\\
    \hline
    Companies(x5)& “Apple” “Google” “Amazon” “Microsoft” “Meta"\\
    \hline
    Common Names(x10)& “John” “Mary” “David” “Sarah” “Michael” “Jessica” “James” “Emma” “Robert” “Olivia"\\
    \hline
 Common Nouns(x20)& “book” “phone” “city” “child” “game” “weather” “news” “people” “person” “house” “car” “school” “dog” “cat” “tree" “river” “mountain” “sun” “food” “water"\\
 \hline
 Common Adjectives(x10)& “good” “new” “first” “last” “long” “great” “little” “odd” “big” “angry”\\
 \hline
    \end{tabular}
    \end{adjustbox}
    \caption{Selected words unrelated to time series}
    \label{tab:selected-words-nr}
\end{table}

\subsection{Dataset Details}
Dataset statistics can be seen from Table \ref{tab:dataset-details}. 
We select a total of seven commonly used datasets to evaluate the results of long-term forecasting, which are ETT (ETTh1, ETTh2, ETTm1, ETTm2) \cite{zhou2021informer}, Electricity\footnote{https://archive.ics.uci.edu/dataset/321/electricityloaddiagrams20112014}, Traffic\footnote{https://pems.dot.ca.gov/} and Weather\footnote{https://www.bgc-jena.mpg.de/wetter/}.

The ETT datasets contains the data of electricity transformers and can be divided to two types: ETTh1, ETTh2: sampled every 1 hour; ETTm1, ETTm2: sampled every 15 minutes; 
The Electricity dataset contains electricity consumption sampled every 1 hour of 321 customers from 2012 to 2014; 
The Traffic dataset, sourced from the California Department of Transportation, provides hourly measurements of road occupancy rates across San Francisco Bay area freeways, collected by numerous sensors; 
The Weather dataset comprises a full year’s worth of data from 21 meteorological stations in Germany, with observations taken every 10 minutes.

\begin{table}[!htbp]
    \centering
    \begin{adjustbox}{max width=\textwidth}
    \begin{tabular}{c|c|c|c|c|c}
    \toprule
    Datasets & Dim& Series Length & Dataset Size & Frequency & Information\\
    \midrule
    ETTh1, ETTh2& 7 & \{96, 192, 336, 720\} & (8545, 2881, 2881) & 1 hour & Temperature \\
    ETTm1, ETTm2& 7 & \{96, 192, 336, 720\} & (34465, 11521, 11521) & 15 min & Temperature \\
    Electricity & 321 & \{96, 192, 336, 720\} & (18317, 2633, 5261) & 1 hour & Electricity \\
    Traffic & 862 & \{96, 192, 336, 720\} & (12185, 1757, 3509) & 1 hour & Transportation \\
    Weather & 21 & \{96, 192, 336, 720\} & (36792, 5271, 10540) & 10 min & Weather \\
    \bottomrule
    \end{tabular}
    \end{adjustbox}
    \caption{Dataset statistics are from \cite{wu2023timesnet}. The dataset size is organized in (Train, Validation, Test)}
    \label{tab:dataset-details}
\end{table}

\subsection{Architectures of ablations}
\begin{figure*}[!htbp]
    \centering
    \subfigure[Base]{
        \includegraphics[width=0.3\textwidth]{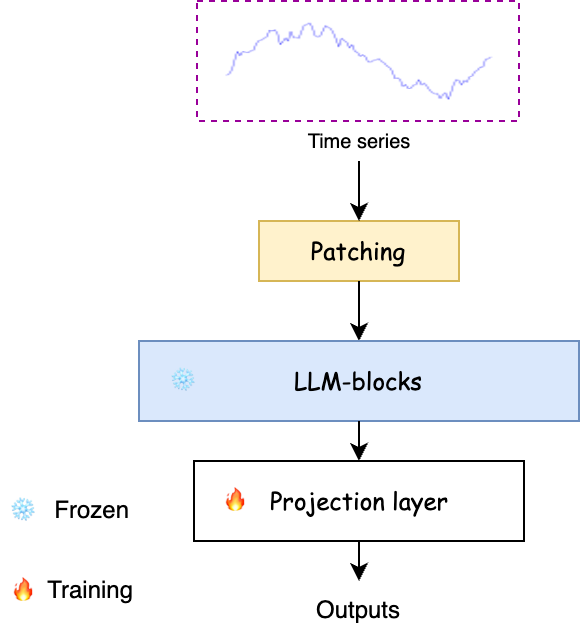}
        \label{base}
    }\hfill
    \subfigure[Base\_Prompt]{
        \includegraphics[width=0.3\textwidth]{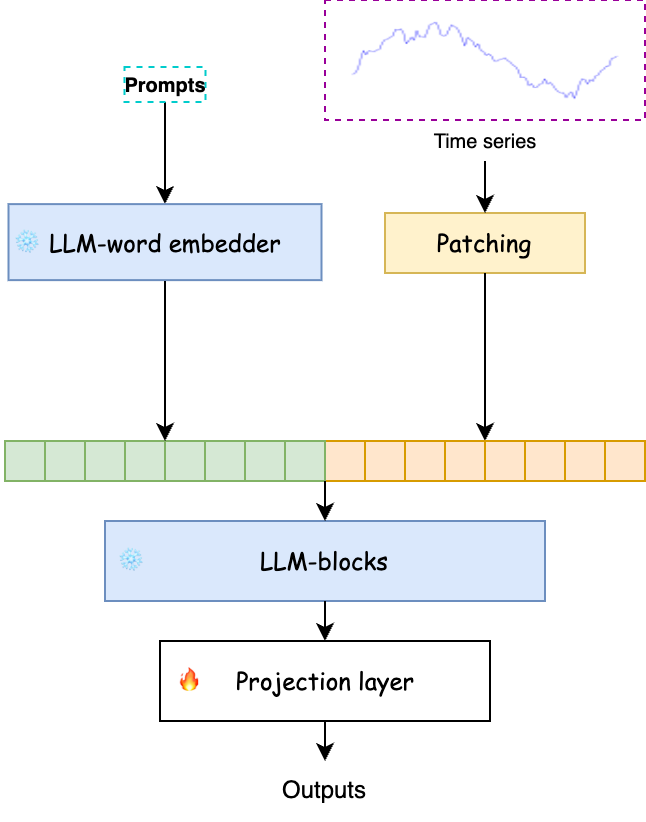}
        \label{base_prompt}
    }\hfill
    \subfigure[Base\_Prototype]{
        \includegraphics[width=0.3\textwidth]{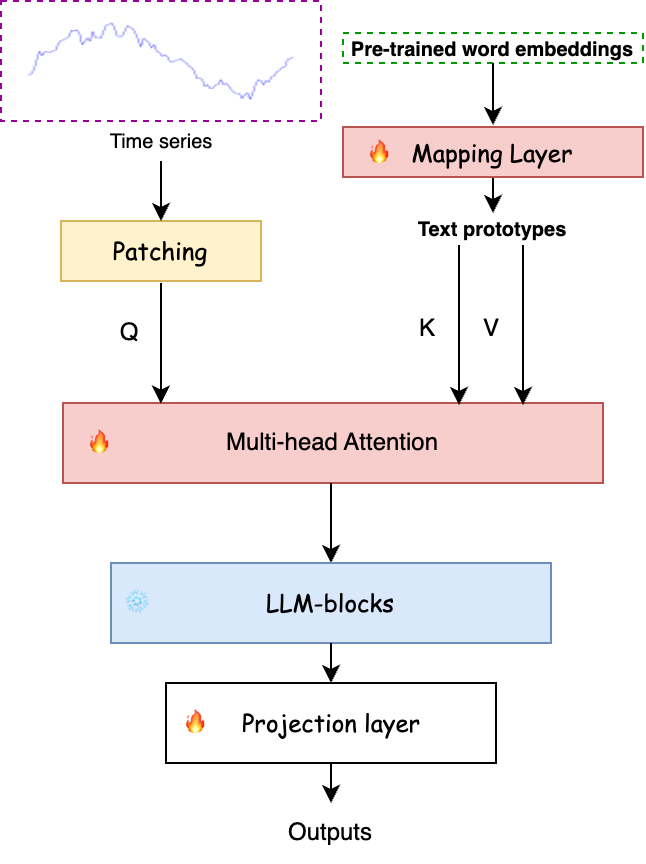}
        \label{base_prototype}
    }
    \caption{Ablations of TimeLLM.}
    \label{ablation-timellm}
\end{figure*}

\begin{figure*}[!htbp]
    \centering
    \includegraphics[width=0.8\textwidth]{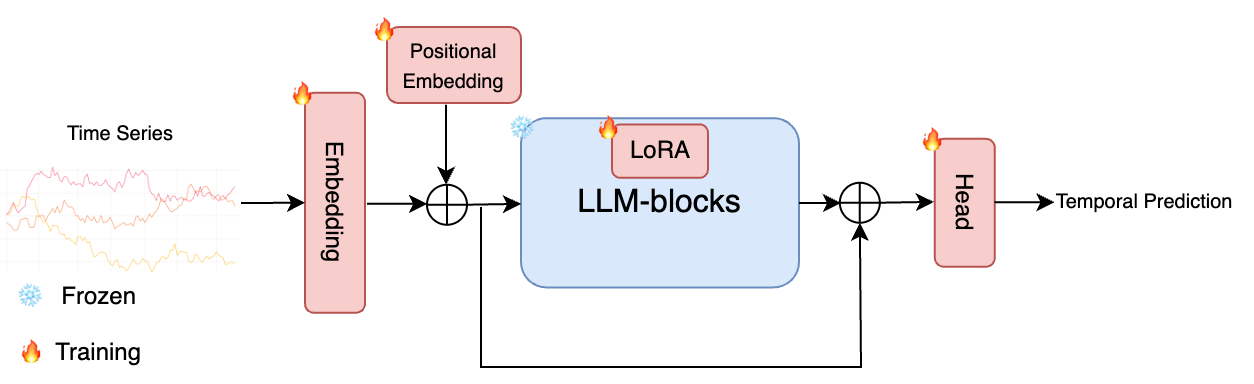}
    \caption{Ablation of CALF. The LLM-blocks module is a simplified form of transformer blocks in temporal part.}
    \label{calf_temporal}
\end{figure*}

The ablation models of TimeLLM and CALF can be seen in Fig.\ref{ablation-timellm} and \ref{calf_temporal} respectively.

\subsection{Evaluation Metrics}
For our all experiments, we utilize the mean square error (MSE) and mean absolute error (MAE) as metrics. The two formulas are:
$\text{MSE} = \frac{1}{n} \sum_{i=1}^{n} (y_i - \hat{y}_i)^2$, 
$\text{MAE} = \frac{1}{n} \sum_{i=1}^{n} |y_i - \hat{y}_i|$
where $n$ is the total number of points.  
$y_i$ represents the ground truth of the $i$-th point.
$\hat{y}_i$ represents the predicted value of the $i$-th point.

\section{Full Experimental Results}
% ------------------------------------
% ------------ question 1 ------------
% ------------------------------------
\subsection{Long-term Forecasting}
The full long-term forecasting results of TimeLLM and its ablations are showed in Table \ref{tab:ltf-timellm-full}. Texts in this case does not improve the performance.
Table \ref{tab:ltf-timellm-bert}, \ref{tab:ltf-timellm-distilbert}, \ref{tab:ltf-timellm-opt125}, \ref{tab:ltf-timellm-opt350} are results changing language model from GPT2 to BERT, DistilBERT, OPT-125M and OPT-350M, respectively.
The full long-term forecasting results of CALF and its ablation are showed in Table
\ref{tab:ltf-calf-full}. 

\begin{table*}[!htbp]
\centering
\begin{tabular}{cccccccccc}
\hline
\multicolumn{2}{c}{Models}& \multicolumn{2}{c}{Base} & \multicolumn{2}{c}{Base\_Prompt} & \multicolumn{2}{c}{Base\_Prototype} & \multicolumn{2}{c}{TimeLLM} \\
 Dataset&  Window& MSE& MAE& MSE& MAE& MSE& MAE& MSE& MAE\\
\hline
\multirow{4}{*}{ETTh1} & 96 & \underline{0.379}& \textbf{0.404}& 0.393& 0.411& 0.387& 0.413& \textbf{0.377}& \underline{0.405}\\
 & 192 & \textbf{0.411}& \textbf{0.425}& 0.422& \underline{0.431}& 0.419& 0.434& \underline{0.414}& \underline{0.431}\\
 & 336 & \textbf{0.426}& \textbf{0.437}& \underline{0.434}& \underline{0.442}& 0.435& 0.443& 0.447& 0.457\\
 & 720 & \textbf{0.459}& \textbf{0.477}& \underline{0.481}& \underline{0.491}& 0.524& 0.515& 0.554& 0.529\\
\hline
\multirow{4}{*}{ETTh2} & 96 & \underline{0.295}& \underline{0.353}& \textbf{0.294}& 0.354& 0.301& \underline{0.353}& \textbf{0.294}& \textbf{0.349}\\
 & 192 & \textbf{0.361}& \textbf{0.397}& \underline{0.362}& \textbf{0.397}& 0.375& 0.403& 0.367& \underline{0.400}\\
 & 336 & 0.392& 0.425& \underline{0.381}& \underline{0.416}& 0.389& 0.426& \textbf{0.380}& \textbf{0.412}\\
 & 720 & 0.425& 0.454& \textbf{0.414}& \textbf{0.448}& 0.427& 0.455& \underline{0.421}& \underline{0.451}\\
\hline
\multirow{4}{*}{ETTm1} & 96 & 0.303& 0.352& 0.307& 0.355& \textbf{0.292}& \textbf{0.349}& \underline{0.293}& \underline{0.350}\\
 & 192 & 0.341& \underline{0.375}& 0.340& 0.377& \textbf{0.330}& 0.376& \underline{0.334}& \textbf{0.374}\\
 & 336 & 0.377& 0.397& 0.376& \underline{0.395}& \textbf{0.362}& \textbf{0.391}& \underline{0.368}& 0.397\\
 & 720 & \textbf{0.413}& \textbf{0.418}& 0.429& \underline{0.422}& \underline{0.423}& 0.435& 0.426& 0.430\\
\hline
\multirow{4}{*}{ETTm2} & 96 & \textbf{0.170}& \underline{0.264}& \underline{0.172}& 0.266& 0.175& \underline{0.264}& \underline{0.172}& \textbf{0.260}\\
 & 192 & \underline{0.230}& \underline{0.302}& \textbf{0.224}& \textbf{0.299}& 0.242& 0.313& 0.231& 0.303\\
 & 336 & \textbf{0.283}& \textbf{0.341}& \underline{0.286}& 0.345& 0.290& \underline{0.344}& 0.289& 0.345\\
 & 720 & \textbf{0.364}& \textbf{0.389}& \underline{0.368}& \underline{0.393}& 0.380& 0.397& 0.379& 0.397\\
\hline
\multirow{4}{*}{Weather}& 96& \textbf{0.147}& \textbf{0.198}& \underline{0.149}& \underline{0.200}& 0.155& 0.207& 0.156& 0.208\\
 & 192& \underline{0.197}& \underline{0.244}& \textbf{0.192}& \textbf{0.242}& 0.200& 0.248& 0.200& 0.249\\
 & 336& \underline{0.248}& 0.285& \textbf{0.246}& \underline{0.284}& \underline{0.248}& \textbf{0.283}& 0.249& 0.285\\
 & 720& \underline{0.319}& 0.336& \underline{0.319}& \underline{0.333}& 0.319& 0.335& \textbf{0.317}& \textbf{0.332}\\
\hline
\multirow{4}{*}{Electricity}& 96 & \underline{0.133}& \textbf{0.230}& 0.133& 0.232& 0.139& 0.248& \textbf{0.132}& \underline{0.232}\\
 & 192 & \textbf{0.149}& \textbf{0.245}& \underline{0.150}& \underline{0.248}& 0.156& 0.259& 0.154& 0.257\\
 & 336 & \textbf{0.165}& \textbf{0.262}& \textbf{0.165}& \underline{0.263}& 0.174& 0.279& \underline{0.173}& 0.277\\
 & 720 & \textbf{0.202}& \textbf{0.294}& \textbf{0.202}& \underline{0.295}& \underline{0.205}& 0.305& 0.208& 0.308\\
\hline
\multirow{4}{*}{Traffic}& 96 & 0.374& \underline{0.261}& 0.375& 0.263& \underline{0.365}& 0.268& \textbf{0.363}& \textbf{0.260}\\
 & 192 & \underline{0.389}& \textbf{0.268}& \underline{0.389}& \underline{0.269}& \textbf{0.382}& 0.275& \textbf{0.382}& 0.272\\
 & 336 & \underline{0.399}& \underline{0.274}& \underline{0.399}& \textbf{0.273}& \textbf{0.397}& 0.281& \underline{0.399}& 0.282\\
 & 720 & \underline{0.436}& \textbf{0.292}& \underline{0.436}& \underline{0.294}& \textbf{0.434}& 0.301& \underline{0.436}& 0.303\\
\hline
\multicolumn{2}{c}{1st count} & 12&  15& 7&  5& 6&  3& 7&  6\\
\multicolumn{2}{c}{2nd count} & 10&  7& 14&  16& 5&  3& 9&  6\\
\hline
\end{tabular}
\caption{Full long-term forecasting results of TimeLLM and its ablations.}
\label{tab:ltf-timellm-full}
\end{table*}

\begin{table*}[!htbp]
\centering
\begin{tabular}{cccccccccc}
\hline
\multicolumn{2}{c}{Models}& \multicolumn{2}{c}{Base} & \multicolumn{2}{c}{Base\_Prompt} & \multicolumn{2}{c}{Base\_Prototype} & \multicolumn{2}{c}{TimeLLM} \\
 Dataset&  Window& MSE& MAE& MSE& MAE& MSE& MAE& MSE& MAE\\
\hline
\multirow{4}{*}{ETTh1} & 96 & 0.407& 0.424& 0.399& \underline{0.418}& \underline{0.391}& 0.420& \textbf{0.390}& \textbf{0.415}\\
 & 192 & 0.440& \underline{0.444}& \textbf{0.434}& \textbf{0.438}& \underline{0.437}& 0.451& 0.446& 0.451\\
 & 336 & 0.453& 0.457& 0.447& \underline{0.452}& \underline{0.438}& 0.457& \textbf{0.424}& \textbf{0.439}\\
 & 720 & 0.469& 0.481& \underline{0.464}& \underline{0.477}& \textbf{0.455}& \textbf{0.473}& 0.477& 0.487\\
\hline
\multirow{4}{*}{ETTh2} & 96 & 0.307& 0.365& 0.313& 0.365& \textbf{0.283}& \textbf{0.344}& \underline{0.298}& \underline{0.352}\\
 & 192 & \underline{0.374}& \underline{0.405}& 0.381& 0.406& 0.381& \underline{0.405}& \textbf{0.368}& \textbf{0.398}\\
 & 336 & 0.400& 0.428& \underline{0.391}& \textbf{0.421}& \textbf{0.389}& \underline{0.423}& 0.395& \textbf{0.421}\\
 & 720 & 0.439& 0.459& \underline{0.427}& \underline{0.453}& 0.468& 0.477& \textbf{0.418}& \textbf{0.447}\\
\hline
\multirow{4}{*}{ETTm1} & 96 & \underline{0.300}& 0.358& 0.301& \underline{0.357}& 0.338& 0.376& \textbf{0.299}& \textbf{0.351}\\
 & 192 & 0.344& 0.381& \underline{0.339}& \underline{0.376}& \textbf{0.338}& \underline{0.376}& \underline{0.339}& \textbf{0.374}\\
 & 336 & \underline{0.373}& \underline{0.403}& \textbf{0.366}& \textbf{0.398}& 0.395& 0.415& 0.407& 0.416\\
 & 720 & \underline{0.429}& \textbf{0.429}& \underline{0.429}& 0.435& 0.432& \textbf{0.429}& \textbf{0.427}& \underline{0.431}\\
\hline
\multirow{4}{*}{ETTm2} & 96 & 0.182& \underline{0.273}& 0.183& 0.275& \textbf{0.170}& \textbf{0.260}& \underline{0.172}& \textbf{0.260}\\
 & 192 & 0.248& 0.318& 0.251& 0.322& \textbf{0.226}& \textbf{0.294}& \underline{0.231}& \underline{0.300}\\
 & 336 & \underline{0.300}& 0.350& 0.301& 0.356& 0.303& \underline{0.349}& \textbf{0.298}& \textbf{0.346}\\
 & 720 & 0.398& 0.410& 0.387& 0.406& \textbf{0.360}& \textbf{0.384}& \underline{0.371}& \underline{0.391}\\
\hline
\multicolumn{2}{c}{1st count} & 0&  1& 2&  3& 7&  6& 7&  9\\
\multicolumn{2}{c}{2nd count} & 5&  4& 5&  6& 3&  4& 5&  4\\
\hline
\end{tabular}
\caption{Full long-term forecasting results of TimeLLM and its ablations. (Language model: Llama 2)}
\label{tab:ltf-timellm-llama2}
\end{table*}

\begin{table*}[!htbp]
\centering
\begin{tabular}{cccccccccc}
\hline
\multicolumn{2}{c}{Models}& \multicolumn{2}{c}{Base} & \multicolumn{2}{c}{Base\_Prompt} & \multicolumn{2}{c}{Base\_Prototype} & \multicolumn{2}{c}{TimeLLM} \\
 Dataset&  Window& MSE& MAE& MSE& MAE& MSE& MAE& MSE& MAE\\
\hline
\multirow{4}{*}{ETTh1} & 96 & 0.394& 0.422& \underline{0.386}& 0.414& 0.387& \underline{0.413}& \textbf{0.377}& \textbf{0.410}\\
 & 192 & 0.433& 0.447& 0.433& 0.477& \textbf{0.411}& \textbf{0.429}& \underline{0.418}& \underline{0.435}\\
 & 336 & 0.448& 0.457& \underline{0.444}& \underline{0.456}& 0.520& 0.501& \textbf{0.439}& \textbf{0.451}\\
 & 720 & 0.488& 0.493& \textbf{0.460}& \textbf{0.475}& \underline{0.464}& \underline{0.481}& 0.562& 0.522\\
\hline
\multirow{4}{*}{ETTh2} & 96 & 0.301& \underline{0.358}& \textbf{0.298}& \textbf{0.355}& 0.315& 0.364& \underline{0.300}& 0.362\\
 & 192 & 0.373& 0.402& \textbf{0.358}& \textbf{0.389}& 0.383& 0.401& \underline{0.369}& \underline{0.397}\\
 & 336 & \textbf{0.375}& \underline{0.413}& \textbf{0.375}& \textbf{0.409}& 0.431& 0.431& \underline{0.395}& 0.419\\
 & 720 & \underline{0.417}& \underline{0.449}& \textbf{0.412}& \textbf{0.441}& \underline{0.417}& 0.450& 0.428& 0.450\\
\hline
\multirow{4}{*}{ETTm1} & 96 & 0.310& 0.359& 0.304& 0.358& \underline{0.300}& \textbf{0.350}& \textbf{0.299}& \underline{0.351}\\
 & 192 & 0.354& 0.384& 0.347& \underline{0.381}& \underline{0.345}& \underline{0.381}& \textbf{0.336}& \textbf{0.375}\\
 & 336 & 0.380& \underline{0.400}& 0.380& \textbf{0.399}& \textbf{0.373}& \underline{0.400}& \underline{0.374}& \textbf{0.399}\\
 & 720 & 0.426& 0.430& 0.427& \textbf{0.426}& \underline{0.425}& \underline{0.427}& \textbf{0.414}& \textbf{0.426}\\
\hline
\multirow{4}{*}{ETTm2} & 96 & 0.184& 0.275& 0.185& 0.276& \underline{0.179}& \underline{0.267}& \textbf{0.173}& \textbf{0.262}\\
 & 192 & 0.251& 0.321& \underline{0.236}& 0.311& 0.239& \underline{0.308}& \textbf{0.232}& \textbf{0.304}\\
 & 336 & 0.291& 0.345& 0.301& 0.353& \textbf{0.277}& \textbf{0.334}& \underline{0.278}& \underline{0.337}\\
 & 720 & 0.378& 0.398& 0.381& 0.398& \underline{0.361}& \textbf{0.390}& \textbf{0.360}& \underline{0.391}\\
\hline
\multicolumn{2}{c}{1st count} & 1&  0& 5&  7& 3&  4& 8&  7\\
\multicolumn{2}{c}{2nd count} & 1&  4& 3&  2& 7&  7& 6&  5\\
\hline
\end{tabular}
\caption{Full long-term forecasting results of TimeLLM and its ablations. (Language model: BERT)}
\label{tab:ltf-timellm-bert}
\end{table*}

\begin{table*}[!htbp]
\centering
\begin{tabular}{cccccccccc}
\hline
\multicolumn{2}{c}{Models}& \multicolumn{2}{c}{Base} & \multicolumn{2}{c}{Base\_Prompt} & \multicolumn{2}{c}{Base\_Prototype} & \multicolumn{2}{c}{TimeLLM} \\
 Dataset&  Window& MSE& MAE& MSE& MAE& MSE& MAE& MSE& MAE\\
\hline
\multirow{4}{*}{ETTh1} & 96 & \textbf{0.385}& \textbf{0.416}& 0.392& \textbf{0.416} 
 & 0.419& 0.442& \underline{0.387}& \underline{0.417}\\
 & 192 & \underline{0.413}& \underline{0.434}& \textbf{0.411}& \textbf{0.429}& 0.485& 0.474& 0.512& 0.480\\
 & 336 & \underline{0.494}& 0.496& \textbf{0.418}& \textbf{0.435}& 0.514& 0.503& 0.513& \underline{0.490}\\
 & 720 & \textbf{0.483}& \textbf{0.492}& 0.647& 0.564& \underline{0.525}& \underline{0.499}& \underline{0.525}& 0.510\\
\hline
\multirow{4}{*}{ETTh2} & 96 & \underline{0.295}& 0.356& \textbf{0.290}& \textbf{0.348}& 0.319& 0.373& 0.298& \underline{0.354}\\
 & 192 & \underline{0.357}& \underline{0.392}& \textbf{0.356}& \textbf{0.389}& 0.407& 0.416& 0.368& 0.395\\
 & 336 & \underline{0.372}& \underline{0.408}& \textbf{0.366}& \textbf{0.401}& 0.424& 0.448& 0.394& 0.417\\
 & 720 & \underline{0.411}& \underline{0.444}& \textbf{0.406}& \textbf{0.439}& 0.451& 0.472& 0.451& 0.468\\
\hline
\multirow{4}{*}{ETTm1} & 96 & 0.319& 0.367& 0.319& 0.366& \underline{0.308}& \underline{0.357}& \textbf{0.302}& \textbf{0.354}\\
 & 192 & 0.365& 0.389& 0.361& 0.386& \underline{0.343}& \textbf{0.378}& \textbf{0.337}& \underline{0.379}\\
 & 336 & 0.387& \underline{0.405}& \underline{0.385}& \textbf{0.401}& \underline{0.385}& 0.406& \textbf{0.375}& \textbf{0.401}\\
 & 720 & \underline{0.429}& \underline{0.429}& 0.432& \textbf{0.426}& 0.433& 0.439& \textbf{0.428}& 0.433\\
\hline
\multirow{4}{*}{ETTm2} & 96 & 0.189& 0.277& 0.187& 0.279& \underline{0.182}& \underline{0.268}& \textbf{0.170}& \textbf{0.262}\\
 & 192 & 0.252& 0.319& \underline{0.240}& 0.313& 0.241& \underline{0.311}& \textbf{0.234}& \textbf{0.306}\\
 & 336 & \underline{0.307}& 0.354& \underline{0.307}& \textbf{0.305}& 0.323& 0.368& \textbf{0.291}& \underline{0.340}\\
 & 720 & 0.379& 0.398& 0.372& \underline{0.392}& \textbf{0.364}& \textbf{0.389}& \underline{0.371}& 0.394\\
\hline
\multicolumn{2}{c}{1st count} & 2&  2& 5&  8& 1&  2& 7&  4\\
\multicolumn{2}{c}{2nd count} & 8&  5& 4&  3& 5&  4& 3&  5\\
\hline
\end{tabular}
\caption{Full long-term forecasting results of TimeLLM and its ablations. (Language model: DistilBERT)}
\label{tab:ltf-timellm-distilbert}
\end{table*}

\begin{table*}[!htbp]
\centering
\begin{tabular}{cccccccccc}
\hline
\multicolumn{2}{c}{Models}& \multicolumn{2}{c}{Base} & \multicolumn{2}{c}{Base\_Prompt} & \multicolumn{2}{c}{Base\_Prototype} & \multicolumn{2}{c}{TimeLLM} \\
 Dataset&  Window& MSE& MAE& MSE& MAE& MSE& MAE& MSE& MAE\\
\hline
\multirow{4}{*}{ETTh1} & 96 & \underline{0.381}& \underline{0.408}& \textbf{0.379}& \textbf{0.407}& \underline{0.381}& \textbf{0.407}& 0.386& 0.411\\
 & 192 & 0.423& 0.438& 0.420& 0.435& \underline{0.417}& \underline{0.432}& \textbf{0.415}& \textbf{0.429}\\
 & 336 & 0.453& 0.463& 0.451& 0.461& \textbf{0.443}& \textbf{0.453}& \underline{0.449}& \underline{0.459}\\
 & 720 & 0.492& 0.503& 0.491& 0.500& \underline{0.465}& \underline{0.481}& \textbf{0.452}& \textbf{0.476}\\
\hline
\multirow{4}{*}{ETTh2} & 96 & 0.300& 0.360& 0.304& 0.361& \underline{0.299}& \underline{0.351}& \textbf{0.284}& \textbf{0.348}\\
 & 192 & \textbf{0.358}& \textbf{0.397}& \underline{0.368}& \underline{0.399}& 0.372& 0.406& 0.420& 0.433\\
 & 336 & \underline{0.381}& 0.415& 0.382& 0.415& \textbf{0.375}& \textbf{0.406}& 0.385& \underline{0.412}\\
 & 720 & 0.425& 0.452& \textbf{0.413}& \textbf{0.445}& 0.430& 0.455& \underline{0.420}& \underline{0.448}\\
\hline
\multirow{4}{*}{ETTm1} & 96 & 0.309& 0.361& 0.309& 0.360& \textbf{0.294}& \textbf{0.348}& \underline{0.297}& \underline{0.352}\\
 & 192 & 0.344& \underline{0.380}& \underline{0.341}& \underline{0.380}& \textbf{0.333}& \textbf{0.372}& \underline{0.341}& 0.381\\
 & 336 & 0.374& 0.397& \underline{0.372}& \underline{0.396}& \textbf{0.369}& \textbf{0.395}& 0.383& 0.401\\
 & 720 & \textbf{0.418}& \textbf{0.424}& 0.431& 0.441& \underline{0.421}& \underline{0.427}& 0.440& 0.437\\
\hline
\multirow{4}{*}{ETTm2} & 96 & \underline{0.173}& 0.268& 0.174& 0.270& 0.178& \underline{0.263}& \textbf{0.172}& \textbf{0.261}\\
 & 192 & 0.236& 0.307& \underline{0.233}& 0.311& 0.236& \underline{0.306}& \textbf{0.232}& \textbf{0.300}\\
 & 336 & \textbf{0.283}& 0.344& 0.287& 0.347& 0.286& \textbf{0.339}& \underline{0.285}& \underline{0.341}\\
 & 720 & \underline{0.367}& 0.393& 0.375& 0.397& 0.371& \underline{0.388}& \textbf{0.363}& \textbf{0.386}\\
\hline
\multicolumn{2}{c}{1st count} & 3&  2& 2&  2& 5&  7& 6&  6\\
\multicolumn{2}{c}{2nd count} & 4&  1& 4&  3& 5&  7& 5&  5\\
\hline
\end{tabular}
\caption{Full long-term forecasting results of TimeLLM and its ablations. (Language model: OPT-125M)}
\label{tab:ltf-timellm-opt125}
\end{table*}

\begin{table*}[!htbp]
\centering
\begin{tabular}{cccccccccc}
\hline
\multicolumn{2}{c}{Models}& \multicolumn{2}{c}{Base} & \multicolumn{2}{c}{Base\_Prompt} & \multicolumn{2}{c}{Base\_Prototype} & \multicolumn{2}{c}{TimeLLM} \\
 Dataset&  Window& MSE& MAE& MSE& MAE& MSE& MAE& MSE& MAE\\
\hline
\multirow{4}{*}{ETTh1} & 96 & 0.425& 0.439& 0.441& 0.450& \underline{0.381}& \underline{0.413}& \textbf{0.378}& \textbf{0.408}\\
 & 192 & 0.439& \underline{0.447}& 0.472& 0.468& \textbf{0.423}& \textbf{0.438}& \underline{0.434}& \underline{0.447}\\
 & 336 & 0.451& \underline{0.460}& 0.514& 0.495& \textbf{0.426}& \textbf{0.444}& \underline{0.448}& \underline{0.460}\\
 & 720 & \underline{0.493}& \underline{0.500}& 0.560& 0.530& 0.548& 0.526& \textbf{0.491}& \textbf{0.498}\\
\hline
\multirow{4}{*}{ETTh2} & 96 & 0.304& 0.366& 0.317& 0.374& \underline{0.292}& \underline{0.351}& \textbf{0.289}& \textbf{0.347}\\
 & 192 & 0.368& 0.405& 0.386& 0.415& \textbf{0.350}& \textbf{0.387}& \underline{0.362}& \underline{0.394}\\
 & 336 & \underline{0.383}& \textbf{0.418}& 0.408& 0.436& \textbf{0.382}& \underline{0.419}& 0.384& 0.420\\
 & 720 & \underline{0.441}& \underline{0.461}& 0.456& 0.469& 0.474& 0.486& \textbf{0.415}& \textbf{0.450}\\
\hline
\multicolumn{2}{c}{1st count} & 0&  1& 0&  0& 4&  3& 4&  4\\
\multicolumn{2}{c}{2nd count} & 3&  4& 0&  0& 2&  3& 3&  3\\
\hline
\end{tabular}
\caption{Full long-term forecasting results of TimeLLM and its ablations. (Language model: OPT-350M)}
\label{tab:ltf-timellm-opt350}
\end{table*}

\begin{table*}[!htbp]
\centering
\begin{tabular}{cccccc}
\hline
\multicolumn{2}{c}{Models}& \multicolumn{2}{c}{CALF\_Temporal} & \multicolumn{2}{c}{CALF}   \\
 Dataset&  Window& MAE & MSE& MAE & MSE\\
\hline
\multirow{4}{*}{ETTh1} 
 & 96 & 0.381& \textbf{0.388}& \textbf{0.374}& 0.394\\
 & 192 & 0.435& \textbf{0.419}& \textbf{0.426}& 0.423\\
 & 336 & 0.483& \textbf{0.443}& \textbf{0.473}& 0.448\\
 & 720 & 0.469& \textbf{0.458}& \textbf{0.467}& 0.463\\
\hline
\multirow{4}{*}{ETTh2} 
 & 96 & \textbf{0.291}& \textbf{0.337}& 0.301& 0.343\\
 & 192 & \textbf{0.349}& \textbf{0.375}& 0.366& 0.386\\
 & 336 & \textbf{0.409}& \textbf{0.417}& 0.412& 0.424\\
 & 720 & \textbf{0.418}& \textbf{0.434}& \textbf{0.418}& 0.435\\
\hline
\multirow{4}{*}{ETTm1} 
 & 96 & \textbf{0.319}& \textbf{0.344}& 0.322& 0.350\\
 & 192 & \textbf{0.361}& \textbf{0.363}& 0.374& 0.377\\
 & 336 & \textbf{0.390}& \textbf{0.385}& 0.412& 0.402\\
 & 720 & \textbf{0.459}& \textbf{0.425}& 0.483& 0.442\\
\hline
\multirow{4}{*}{ETTm2} 
 & 96 & \textbf{0.173}& \textbf{0.251}& 0.178& 0.258\\
 & 192 & \textbf{0.242}& \textbf{0.296}& 0.247& 0.300\\
 & 336 & \textbf{0.298}& \textbf{0.333}& 0.306& 0.340\\
 & 720 & \textbf{0.400}& \textbf{0.394}& 0.405& 0.397\\
\hline
\multirow{4}{*}{Weather}
 & 96& \textbf{0.165}& 0.206& 0.167& \textbf{0.205}\\
 & 192& \textbf{0.210}& \textbf{0.249}& 0.214& 0.253\\
 & 336& \textbf{0.267}& \textbf{0.291}& 0.271& 0.292\\
 & 720& \textbf{0.348}& \textbf{0.344}& 0.350& 0.346\\
\hline
\multirow{4}{*}{Electricity}
 & 96 & 0.151& 0.245& \textbf{0.146}& \textbf{0.239}\\
 & 192 & 0.166& 0.259& \textbf{0.162}& \textbf{0.253}\\
 & 336 & 0.184& 0.275& \textbf{0.178}& \textbf{0.269}\\
 & 720 & 0.220& 0.302& \textbf{0.210}& \textbf{0.299}\\
\hline
\multirow{4}{*}{Traffic}
 & 96 & 0.435& \textbf{0.292}& \textbf{0.411}& 0.269\\
 & 192 & 0.451& 0.295& \textbf{0.430}& \textbf{0.276}\\
 & 336 & 0.467& 0.302& \textbf{0.444}& \textbf{0.281}\\
 & 720 & 0.499& 0.320& \textbf{0.480}& \textbf{0.302}\\
\hline
\multicolumn{2}{c}{1st count} & 16&  20& 13&  8\\
\hline
\end{tabular}
\caption{Full long-term forecasting results of CALF and its ablation.}
\label{tab:ltf-calf-full}
\end{table*}

% ------------------------------------
% ------------ question 2 ------------
% ------------------------------------
\subsection{Few-shot Forecasting}
The full few-shot forecasting results of TimeLLM and its ablations are showed in Table \ref{tab:fsf-timellm-full}.

\begin{table*}[!htbp]
\centering
\begin{tabular}{cccccccccc}
\hline
\multicolumn{2}{c}{Models}& \multicolumn{2}{c}{Base} & \multicolumn{2}{c}{Base\_Prompt} & \multicolumn{2}{c}{Base\_Prototype} & \multicolumn{2}{c}{TimeLLM} \\
 Dataset&  Window& MSE& MAE& MSE& MAE& MSE& MAE& MSE& MAE\\
\hline
\multirow{4}{*}{ETTh1} & 96 & \underline{0.658}& \underline{0.517}& \textbf{0.653}& \textbf{0.516}& 0.799& 0.564& 0.752& 0.556\\
 & 192 & \textbf{0.524}& \textbf{0.488}& \underline{0.546}& \underline{0.499}& 0.621& 0.556& 0.613& 0.540\\
 & 336 & \textbf{0.611}& \textbf{0.539}& \underline{0.613}& \underline{0.541}& 0.730& 0.595& 0.766& 0.612\\
 & 720 & \textbf{0.735}& \textbf{0.608}& \underline{0.744}& \underline{0.611}& 0.912& 0.670& 0.854& 0.635\\
\hline
\multirow{4}{*}{ETTh2} & 96 & \textbf{0.344}& \textbf{0.393}& \underline{0.346}& \underline{0.396}& 0.379& 0.415& 0.373& 0.413\\
 & 192 & \textbf{0.398}& \textbf{0.424}& \underline{0.399}& \underline{0.426}& 0.430& 0.448& 0.414& 0.435\\
 & 336 & \textbf{0.409}& \textbf{0.439}& \underline{0.411}& \underline{0.440}& 0.413& 0.442& 0.501& 0.487\\
 & 720 & \textbf{0.436}& \textbf{0.464}& \underline{0.439}& \underline{0.466}& 0.497& 0.497& 0.502& 0.498\\
\hline
\multirow{4}{*}{ETTm1} & 96 & \textbf{0.399}& \textbf{0.412}& 0.417& 0.421& 0.446& 0.435& \underline{0.404}& \underline{0.415}\\
 & 192 & \textbf{0.412}& \textbf{0.418}& \underline{0.440}& \underline{0.429}& 0.466& 0.445& 0.468& 0.450\\
 & 336 & 0.547& \underline{0.481}& 0.625& 0.515& \underline{0.523}& 0.482& \textbf{0.489}& \textbf{0.463}\\
 & 720 & 0.626& 0.529& 0.670& 0.535& \underline{0.594}& \underline{0.523}& \textbf{0.583}& \textbf{0.518}\\
\hline
\multirow{4}{*}{ETTm2} & 96 & 0.209& 0.286& 0.213& 0.290& \textbf{0.188}& \textbf{0.272}& \underline{0.197}& \underline{0.279}\\
 & 192 & \underline{0.255}& \underline{0.316}& 0.263& 0.321& \textbf{0.252}& \textbf{0.314}& 0.275& 0.337\\
 & 336 & \textbf{0.301}& \textbf{0.342}& \underline{0.310}& 0.353& 0.311& 0.354& \underline{0.310}& \underline{0.352}\\
 & 720 & \underline{0.399}& \textbf{0.403}& 0.426& 0.421& \textbf{0.388}& \underline{0.411}& 0.412& 0.413\\
\hline
\multirow{4}{*}{Electricity}& 96 & \textbf{0.141}& \textbf{0.243}& \underline{0.142}& \textbf{0.243}& 0.143& 0.245& 0.143& \underline{0.244}\\
 & 192 & \textbf{0.158}& 0.259& \textbf{0.158}& \underline{0.258}& 0.160& 0.259& \underline{0.159}& \textbf{0.257}\\
 & 336 & \textbf{0.179}& \underline{0.280}& \textbf{0.179}& \textbf{0.279}& 0.183& 0.281& \underline{0.182}& 0.281\\
 & 720 & \textbf{0.240}& 0.331& \textbf{0.240}& \underline{0.329}& \underline{0.241}& 0.332& \textbf{0.240}& \textbf{0.326}\\
\hline
\multicolumn{2}{c}{1st count} & 14&  12& 4&  3& 3&  2& 3&  4\\
\multicolumn{2}{c}{2nd count} & 3&  4& 10&  10& 3&  2& 5&  4\\
\hline
\end{tabular}
\caption{Full few-shot forecasting results of TimeLLM and its ablations.}
\label{tab:fsf-timellm-full}
\end{table*}

% ------------------------------------
% ------------ question 3 ------------
% ------------------------------------
\subsection{Zero-shot Forecasting}
The full zero-shot forecasting results of TimeLLM and its ablations are showed in Table \ref{tab:zsf-timellm-full}.

\begin{table*}[!htbp]
\centering
\begin{adjustbox}{max width=\textwidth}
\begin{tabular}{cccccccccccccccc}
\hline
\multicolumn{2}{c}{Models} & \multicolumn{2}{c}{Base(woPre+woFT)}& \multicolumn{2}{c}{Base} & \multicolumn{2}{c}{Base\_Prompt} & \multicolumn{2}{c}{Base\_Prototype(100)} & \multicolumn{2}{c}{Base\_Prototype(1000)}& \multicolumn{2}{c}{TimeLLM(100)} & \multicolumn{2}{c}{TimeLLM(1000)}\\
 Dataset&  Window  &MSE&MAE& MSE& MAE& MSE& MAE& MSE& MAE & MSE&MAE& MSE& MAE & MSE&MAE\\
\hline
\multirow{4}{*}{h1$\rightarrow$h2}& 96   &0.274&0.342& 0.274& 0.342& 0.276& 0.346& \underline{0.271}& \textbf{0.335} & 0.280&0.340& \textbf{0.270}&  \textbf{0.335}& \underline{0.271}& \underline{0.337}\\
 & 192   &0.334&0.380& 0.333& 0.379& 0.336& 0.383& 0.344&  0.379& 0.333& \underline{0.375}& \textbf{0.328}&  \textbf{0.372}& \underline{0.331}& \underline{0.375}\\
 & 336   &0.391&0.431& 0.366& 0.407& 0.372& 0.414& 0.370&  0.405& \underline{0.359}& \underline{0.404}& \textbf{0.357}&  \textbf{0.403}& 0.363&0.405\\
 & 720   &0.417&0.451& 0.401& 0.438& 0.405& 0.442& \textbf{0.395}&  \underline{0.436}& 0.436&0.455& \underline{0.397}&  \textbf{0.434}& 0.409&0.444\\
\hline
\multirow{4}{*}{m1$\rightarrow$m2}& 96   &0.175&0.262& 0.175& 0.263& 0.176& 0.266& 0.174&  0.258& 0.171&0.258& \underline{0.169}&  \underline{0.257}& \textbf{0.167}& \textbf{0.254}\\
 & 192   &0.231&0.299& 0.230& 0.300& 0.231& 0.302& 0.235&  0.298& \underline{0.225}& \underline{0.295}& \textbf{0.223}&  \textbf{0.293}& \textbf{0.223}&\textbf{0.293}\\
 & 336   &0.302&0.354& 0.289& 0.338& 0.290& 0.339& \textbf{0.282}&  \textbf{0.332}& 0.284&0.337& 0.284&  \underline{0.333}& \underline{0.283}&\underline{0.333}\\
 & 720   &\textbf{0.370}&\textbf{0.389}& \underline{0.372}& \underline{0.390}& \textbf{0.370}& 0.392& 0.383&  0.400& \textbf{0.370}&\textbf{0.389}& 0.376&  0.397& 0.374&0.391\\
\hline
\multirow{4}{*}{h1$\rightarrow$m1}& 96   &0.715&0.544& \underline{0.695}& \textbf{0.538}& \textbf{0.694}& \underline{0.541}& 0.749&  0.549& 0.771&0.559& 0.754&  0.551& 0.728&0.547\\
 & 192   &0.738&0.562& \underline{0.714}& \textbf{0.551}& \textbf{0.709}& \underline{0.552}& 0.776&  0.573& 0.788&0.579& 0.774&  0.573& 0.747&0.567\\
 & 336   &0.813&0.612& 0.754& \underline{0.568}& 0.766& 0.576& \underline{0.749}&  0.573& 0.768&0.592& \textbf{0.740}&  \textbf{0.565}& 0.794&0.603\\
 & 720   &0.863&0.644& \textbf{0.757}& \textbf{0.587}& \underline{0.776}& \underline{0.594}& 0.964&  0.634& 0.816&0.606& 0.833&  0.625& 0.799&0.604\\
\hline
\multirow{4}{*}{h1$\rightarrow$m2}& 96   &0.210&0.304& 0.207& \underline{0.302}& 0.211& 0.307& 0.215&  0.303& 0.211&0.303& \underline{0.203}&  \textbf{0.296}& \textbf{0.202}&\textbf{0.296}\\
 & 192   &0.264&0.337& \textbf{0.259}& \textbf{0.334}& \underline{0.261}& \underline{0.336}& 0.284&  0.345& 0.270&0.339& 0.262&  \textbf{0.334}& 0.271&0.340\\
 & 336   &0.333&0.386& 0.324& 0.375& 0.321& 0.375& \textbf{0.315}&  \textbf{0.366}& 0.320&0.369& \underline{0.317}&  \underline{0.367}& 0.327&0.377\\
 & 720   &0.417&0.429& \textbf{0.408}& \textbf{0.419}& 0.425& 0.428& 0.472&  0.455& 0.449&0.447& 0.419&  \underline{0.425}& \underline{0.412}& \underline{0.425}\\
\hline
\multicolumn{2}{c}{1st count} & 1& 1& 3& 5& 3& 0& 3& 3& 1& 1& 5& 8& 3& 3\\
\multicolumn{2}{c}{2nd count} & 0& 0& 3& 3& 2& 3& 2& 1& 2& 3& 4& 4& 4& 4\\
\hline
\end{tabular}
\end{adjustbox}
\caption{Full zero-shot forecasting results of TimeLLM and its ablations.}
\label{tab:zsf-timellm-full}
\end{table*}

% ------------------------------------
% ------------ question 4 ------------
% ------------------------------------
\subsection{Text Prototypes Extraction Methods}
The full results of six text prototypes extraction methods can be seen from Table \ref{tab:pm-timellm-full}. PCA and Linear achieve the best and second-best results, respectively. 

\begin{table*}[!htbp]
\centering
\begin{adjustbox}{max width=\textwidth}
\begin{tabular}{cccccccccccccc}
\hline
\multicolumn{2}{c}{Methods}& \multicolumn{2}{c}{Linear (Default)} & \multicolumn{2}{c}{PCA}  & \multicolumn{2}{c}{Kmeans} & \multicolumn{2}{c}{Random} & \multicolumn{2}{c}{Text} & \multicolumn{2}{c}{Similarity}\\
 Dataset&  Window& MSE& MAE& MSE& MAE& MSE& MAE & MSE& MAE& MSE& MAE& MSE&MAE\\
\hline
\multirow{4}{*}{ETTh1} & 96 & \textbf{0.377}& \textbf{0.405}& \underline{0.384}& \underline{0.412}& 0.401&  0.427& 0.403& 0.428& 0.390& 0.415& 0.422&0.441\\
 & 192 & \textbf{0.421}& \underline{0.437}& \textbf{0.421}& \textbf{0.435}& 0.465&  0.467& 0.490& 0.481& \underline{0.435}& 0.447& 0.491&0.482\\
 & 336 & 0.469& 0.472& \textbf{0.442}& \textbf{0.455}& 0.485&  0.482& 0.512& 0.498& 0.460& 0.466& \underline{0.449}& \underline{0.459}\\
 & 720 & 0.539& 0.523& \textbf{0.478}& \textbf{0.493}& \underline{0.503}&  \underline{0.504}& 0.522& 0.517& 0.524& 0.513& 0.529&0.518\\
\hline
\multirow{4}{*}{ETTh2} & 96 & \textbf{0.290}& \textbf{0.346}& \underline{0.294}& 0.349& 0.301&  0.354& 0.296& 0.352& \underline{0.294}& \underline{0.348}& 0.295&0.351\\
 & 192 & \underline{0.352}& \textbf{0.388}& \textbf{0.349}& \textbf{0.388}& 0.365&  0.400& 0.363& 0.398& 0.378& \underline{0.396}& 0.369&0.403\\
 & 336 & \textbf{0.373}& \textbf{0.408}& \underline{0.380}& \underline{0.415}& 0.387&  0.419& 0.383& \underline{0.415}& 0.392& 0.418& 0.390&0.418\\
 & 720 & \underline{0.425}& 0.455& \textbf{0.424}& \underline{0.453}& 0.430&  \underline{0.453}& 0.433& 0.455& \underline{0.425}& \textbf{0.450}& 0.427&\underline{0.453}\\
\hline
\multirow{4}{*}{ETTm1} & 96 & \underline{0.291}& \textbf{0.347}& \textbf{0.290}& \underline{0.349}& 0.298&  0.353& 0.297& 0.353& 0.295& 0.352& 0.296&0.352\\
 & 192 & 0.344& \underline{0.379}& \textbf{0.334}& \textbf{0.374}& 0.342&  0.380& 0.342& \underline{0.379}& 0.341& 0.382& \underline{0.339}&0.380\\
 & 336 & 0.371& \underline{0.399}& \textbf{0.369}& \textbf{0.394}& \underline{0.370}&  0.401& 0.376& \underline{0.399}& \underline{0.370}& 0.402& 0.374&0.401\\
 & 720 & \underline{0.414}& \textbf{0.424}& \textbf{0.411}& \textbf{0.424}& 0.442&  0.440& 0.426& 0.434& 0.444& 0.443& 0.426&\underline{0.433}\\
\hline
\multirow{4}{*}{ETTm2} & 96 & 0.170& 0.263& \textbf{0.168}& 0.261& 0.172&  \underline{0.260}& \underline{0.169}& \textbf{0.259}& 0.171& \textbf{0.259}& 0.172&0.261\\
 & 192 & 0.237& 0.308& \underline{0.225}& 0.301& \textbf{0.223}&  0.300& \underline{0.225}& \underline{0.299}& 0.227& \textbf{0.298}& 0.227&\textbf{0.298}\\
 & 336 & 0.297& 0.344& 0.285& 0.340& 0.284&  \underline{0.335}& 0.282& \underline{0.335}& \underline{0.281}& \textbf{0.333}& \textbf{0.280}&\textbf{0.333}\\
 & 720 & 0.370& 0.391& \underline{0.365}& \underline{0.388}& 0.366&  0.392& 0.369& 0.391& 0.386& 0.411& \textbf{0.361}& \textbf{0.387}\\
\hline
\multicolumn{2}{c}{1st count} & 4&  6& 10&  7& 1&   0& 0& 1& 2& 3& 0&4\\
\multicolumn{2}{c}{2nd count} & 4&  3& 5&  5& 2&   4& 2& 5& 2& 3& 5&2\\
\hline
\end{tabular}
\end{adjustbox}
\caption{Full long-term forecasting results with different text prototypes extraction methods. The number of text prototypes is 100. }
\label{tab:pm-timellm-full}
\end{table*}

% ------------------------------------
% ------------ question 5 ------------
% ------------------------------------
\subsection{Number of Text Prototypes}
The full results of different numbers of text prototypes can be seen from Table \ref{tab:pn-timellm-full}. 
% The number range from 10 to 1000. And the best results only increase between 0.2\% and 2.0\%.

\begin{table*}[!htbp]
\centering
\begin{adjustbox}{max width=\textwidth}
\begin{tabular}{cccccccccccccccccccc}
\hline
\multicolumn{2}{c}{Number} & \multicolumn{2}{c}{10} & \multicolumn{2}{c}{20}& \multicolumn{2}{c}{50} & \multicolumn{2}{c}{100} & \multicolumn{2}{c}{150} & \multicolumn{2}{c}{250}& \multicolumn{2}{c}{500} & \multicolumn{2}{c}{600}& \multicolumn{2}{c}{1000}\\
 Dataset&  Window  &MSE&MAE & MSE&MAE& MSE& MAE& MSE&MAE  & MSE&MAE & MSE&MAE& MSE&MAE  & MSE&MAE& MSE&MAE\\
\hline
\multirow{2}{*}{ETTh1}& 96   &0.383& 0.408& 0.379&\underline{0.405}& \underline{0.376}&  \textbf{0.404}& 0.377&  \underline{0.405}& 0.379& 0.407& 0.378&0.406& 0.377&  \underline{0.405}& \textbf{0.375}&\underline{0.405}& 0.378&0.407\\
 & 192   &0.419& 0.433& 0.444&0.452& 0.417&  0.433& 0.421&  0.437& 0.418& 0.433& 0.422&0.436& 0.419&  0.434& \underline{0.416}& \underline{0.432}& \textbf{0.414}& \textbf{0.431}\\
 & 336   &\underline{0.431}& \textbf{0.445}& 0.442& \underline{0.454}& 0.482&  0.477& 0.469&  0.472& 0.486& 0.479& 0.450&0.459& \textbf{0.430}& \textbf{0.445}& 0.468&0.469& 0.467&0.468\\
 & 720   &0.560& 0.535& 0.557&0.534& \textbf{0.489}&  \textbf{0.493}& 0.539&  0.523& \underline{0.492}& \underline{0.497}& 0.500&0.502& 0.521&  0.513& 0.517&0.511& 0.554&0.529\\
\hline
\multirow{2}{*}{ETTm1}& 96   &0.295& 0.351& 0.293&0.348& 0.295&  0.351& \underline{0.291}&  \textbf{0.347}& 0.296& 0.351& \underline{0.291}&0.349& \textbf{0.290}&  \underline{0.348}& 0.292& \underline{0.348}& 0.293&0.350\\
 & 192   &0.336& \underline{0.375}& 0.343&0.381& \textbf{0.331}& \textbf{0.374}& 0.344&  0.379& 0.339& 0.378& 0.339&0.378& 0.337&  0.378& 0.336&0.376& \underline{0.334}& \textbf{0.374}\\
 & 336   &0.369& 0.400& \textbf{0.364}&0.395& \underline{0.365}&  0.394& 0.371&  0.399& \textbf{0.364}& \underline{0.393}& 0.376&0.397& \underline{0.365}&  \textbf{0.392}& 0.368& \underline{0.393}& 0.368&0.397\\
 & 720   &0.435& 0.435& 0.433&0.431& 0.428&  0.432& \textbf{0.414}& \textbf{0.424}& 0.422& 0.428& \underline{0.418}& \underline{0.426}& 0.438&  0.433& 0.419&0.427& 0.426&0.430\\
\hline
\multicolumn{2}{c}{1st count} & 0&  1& 1&0& 2&   3& 1&  2& 1& 0& 0&0& 2&  2& 1&0& 1&2\\
\multicolumn{2}{c}{2nd count} & 1&  1& 0&2& 2&   0& 1&  1& 1& 2& 2&1& 1&  2& 1&4& 1&0\\
\hline
\end{tabular}
\end{adjustbox}
\caption{Full long-term forecasting results with different numbers of text prototypes. }
\label{tab:pn-timellm-full}
\end{table*}

% ------------------------------------
% ------------ question 6 ------------
% ------------------------------------
\subsection{Hidden Dimensions of Text Prototypes}
The full results of different hidden dimensions of text prototypes can be seen from Table \ref{tab:pd-timellm-full}.

\begin{table*}[!htbp]
\centering
\begin{adjustbox}{max width=\textwidth}
\begin{tabular}{cccccccccccccccc}
\hline
\multicolumn{2}{c}{Dimension} & \multicolumn{2}{c}{16} & \multicolumn{2}{c}{32}& \multicolumn{2}{c}{64} & \multicolumn{2}{c}{128} & \multicolumn{2}{c}{256} & \multicolumn{2}{c}{512} & \multicolumn{2}{c}{768}\\
 Dataset&  Window  &MSE&MAE & MSE&MAE& MSE& MAE& MSE&MAE  & MSE&MAE & MSE&MAE  & MSE&MAE\\
\hline
\multirow{2}{*}{ETTh1}& 96   &0.382& 0.411& 0.380&0.408& 0.388&  0.415& 0.389&  0.415& \underline{0.379}& \underline{0.407}& \textbf{0.377}&  \textbf{0.405}& \textbf{0.377}& \textbf{0.405}\\
 & 192   &0.423& 0.436& 0.427&0.437& 0.428&  0.441& 0.428&  0.440& \textbf{0.413}& \textbf{0.429}& \underline{0.419}&  \underline{0.435}& 0.421&0.437\\
 & 336   &0.440& 0.453& 0.449&0.458& 0.451&  0.459& \textbf{0.434}& \textbf{0.448}& \textbf{0.434}& \textbf{0.448}& \underline{0.436}& \underline{0.450}& 0.469&0.472\\
 & 720   &0.492& 0.497& \underline{0.485}& \underline{0.494}& 0.509&  0.508& \textbf{0.464}& \textbf{0.482}& 0.495& 0.501& 0.552&  0.528& 0.539&0.523\\
\hline
\multirow{2}{*}{ETTm1}& 96   &0.295& 0.352& 0.294&0.352& 0.295&  0.352& \underline{0.292}& \underline{0.348}& 0.295& 0.350& 0.293&  0.352& \textbf{0.291}& \textbf{0.347}\\
 & 192   & \textbf{0.333}& \underline{0.375}& 0.339&0.377& \underline{0.334}& \underline{0.375}& \underline{0.334}&  0.376& 0.335& \underline{0.375}& \textbf{0.333}& \textbf{0.374}& 0.344&0.379\\
 & 336   &0.366& \textbf{0.392}& \textbf{0.361}& \underline{0.393}& 0.363&  0.394& \underline{0.362}& \underline{0.393}& \underline{0.362}& \underline{0.393}& 0.363&  0.395& 0.371&0.399\\
 & 720   &0.422& 0.430& 0.431&0.433& 0.423&  0.429& 0.428&  \textbf{0.423}& \underline{0.421}& 0.425& 0.438&  0.435& \textbf{0.414}& \underline{0.424}\\
\hline
\multicolumn{2}{c}{1st count} & 1&  1& 1&0& 0&   0& 2&  3& 2& 2& 2&  2& 3&2\\
\multicolumn{2}{c}{2nd count} & 0&  1& 1&2& 1&   1& 3&  2& 3& 3& 2&  2& 0&1\\
\hline
\end{tabular}
\end{adjustbox}
\caption{Full long-term forecasting results with different hidden dimensions of text prototypes.}
\label{tab:pd-timellm-full}
\end{table*}

% ------------------------------------
% ------------ question 7 ------------
% ------------------------------------
\subsection{Inner Semantics of Texts}
The full results of random replaced prompts and word embeddings can be seen from Table \ref{tab:ltf-timellm-random-replace-prompts-full} and Table \ref{tab:ltf-timellm-random-replace-words-full} respectively.

\begin{table*}[!htbp]
\centering
\begin{tabular}{cccccccccccc}
\hline
\multicolumn{2}{c}{Ratio}& \multicolumn{2}{c}{0\%} & \multicolumn{2}{c}{10\%} & \multicolumn{2}{c}{40\%} & \multicolumn{2}{c}{70\%} & \multicolumn{2}{c}{100\%}\\
 Dataset&  Window& MSE& MAE& MSE& MAE& MSE& MAE& MSE& MAE & MSE&MAE\\
\hline
\multirow{4}{*}{ETTh1} & 96 & \textbf{0.377}& \textbf{0.405}& \textbf{0.377}& \underline{0.406}& \underline{0.378}& 0.409& 0.389&  0.415& 0.384&0.410\\
 & 192 & \textbf{0.414}& \textbf{0.431}& 0.422& \underline{0.437}& \underline{0.421}& \underline{0.437}& 0.429&  0.443& 0.422& \underline{0.437}\\
 & 336 & \underline{0.447}& \underline{0.457}& 0.451& 0.461& \textbf{0.430}& \textbf{0.447}& 0.456&  0.465& 0.506&0.492\\
 & 720 & 0.554& 0.529& \textbf{0.514}& \textbf{0.507}& 0.558& 0.533& \underline{0.538}& \underline{0.523}& 0.551&0.529\\
\hline
\multirow{4}{*}{ETTh2} & 96 & \underline{0.294}& \underline{0.349}& 0.296& 0.350& 0.300& 0.357& \textbf{0.291}& \textbf{0.347}& 0.296&0.350\\
 & 192 & 0.367& 0.400& \textbf{0.356}& \underline{0.390}& \underline{0.357}& \textbf{0.388}& 0.371&  0.393& 0.365&0.392\\
 & 336 & \underline{0.380}& 0.412& 0.394& 0.417& 0.388& 0.417& \textbf{0.375}& \textbf{0.408}& \underline{0.380}& \underline{0.411}\\
 & 720 & \underline{0.421}& \textbf{0.451}& 0.422& \underline{0.452}& \textbf{0.420}& 0.453& 0.427&  0.453& 0.425&0.453\\
\hline
\multicolumn{2}{c}{1st count} & 2&  3& 3&  1& 2&  2& 2&   2& 0&0\\
\multicolumn{2}{c}{2nd count} & 4&  2& 0&  4& 3&  1& 1&   1& 1&2\\
\hline
\end{tabular}
\caption{Full long-term forecasting results with random replaced prompts. }
\label{tab:ltf-timellm-random-replace-prompts-full}
\end{table*}

\begin{table*}[!htbp]
\centering
\begin{tabular}{cccccccccccc}
\hline
\multicolumn{2}{c}{Ratio}& \multicolumn{2}{c}{0\%} & \multicolumn{2}{c}{10\%} & \multicolumn{2}{c}{40\%} & \multicolumn{2}{c}{70\%} & \multicolumn{2}{c}{100\%}\\
 Dataset&  Window& MSE& MAE& MSE& MAE& MSE& MAE& MSE& MAE & MSE&MAE\\
\hline
\multirow{4}{*}{ETTh1} & 96 & 0.399& 0.420& \textbf{0.377}& \textbf{0.408}& 0.395& \underline{0.415}& 0.387&  0.416& \underline{0.379}&\textbf{0.408}\\
 & 192 & 0.426& 0.440& \underline{0.419}& 0.439& 0.420& \underline{0.433}& 0.427&  0.441& \textbf{0.413}& \textbf{0.430}\\
 & 336 & \textbf{0.423}& \textbf{0.437}& 0.490& 0.481& 0.471& 0.472& 0.520&  0.496& \underline{0.432}& \underline{0.444}\\
 & 720 & \textbf{0.511}& \underline{0.501}& \textbf{0.511}& \textbf{0.499}& \underline{0.532}& 0.511& 0.550& 0.521& 0.539&0.524\\
\hline
\multirow{4}{*}{ETTh2} & 96 & 0.302& \underline{0.355}& 0.300& 0.357& \underline{0.298}& \textbf{0.352}& \textbf{0.297}& \underline{0.355}& 0.326&0.373\\
 & 192 & \textbf{0.367}& \underline{0.404}& 0.386& 0.410& \underline{0.374}& 0.408& 0.377&  \textbf{0.403}& 0.376&0.409\\
 & 336 & 0.400& \underline{0.425}& \underline{0.381}& \textbf{0.417}& \textbf{0.373}& \textbf{0.417}& 0.433& 0.444& 0.412& 0.433\\
 & 720 & \textbf{0.421}& \textbf{0.452}& \underline{0.432}& \underline{0.454}& 0.465& 0.477& 0.441&  0.461& 0.455&0.471\\
\hline
\multicolumn{2}{c}{1st count} & 4&  2& 2&  3& 1&  2& 1&   1& 1&2\\
\multicolumn{2}{c}{2nd count} & 0&  4& 3&  1& 3&  2& 0&   1& 2&1\\
\hline
\end{tabular}
\caption{Full long-term forecasting results with random replaced prompts. (Language model: Llama2)}
\label{tab:ltf-timellm-random-replace-prompts-full-llama2}
\end{table*}

\begin{table*}[!ht]
\centering
\begin{tabular}{cccccccccccc}
\hline
\multicolumn{2}{c}{Ratio}& \multicolumn{2}{c}{0\%} & \multicolumn{2}{c}{10\%} & \multicolumn{2}{c}{40\%} & \multicolumn{2}{c}{70\%} & \multicolumn{2}{c}{100\%}\\
 Dataset&  Window& MSE& MAE& MSE& MAE& MSE& MAE& MSE& MAE & MSE&MAE\\
\hline
\multirow{4}{*}{ETTh1} & 96 & \underline{0.377}& \textbf{0.405}& \textbf{0.375}& \textbf{0.405}& 0.392& \underline{0.419}& 0.396&  0.422& 0.390&0.421\\
 & 192 & \textbf{0.414}& \textbf{0.431}& 0.434& 0.446& \underline{0.426}& \underline{0.440}& 0.434&  0.449& 0.435&0.449\\
 & 336 & \textbf{0.447}& \textbf{0.457}& \underline{0.448}& \textbf{0.457}& \underline{0.448}& \textbf{0.457}& 0.462&  \underline{0.467}& 0.458&0.470\\
 & 720 & 0.554& 0.529& \textbf{0.499}& \textbf{0.499}& 0.540& 0.518& 0.582&  0.547& \underline{0.515}& \underline{0.515}\\
\hline
\multirow{4}{*}{ETTh2} & 96 & \underline{0.294}& \underline{0.349}& 0.318& 0.362& 0.305& 0.353& 0.297& \underline{0.349}& \textbf{0.292}& \textbf{0.347}\\
 & 192 & 0.367& 0.400& \underline{0.359}& 0.394& \textbf{0.354}& \textbf{0.386}& 0.369&  0.397& 0.365& \underline{0.391}\\
 & 336 & \textbf{0.380}& \textbf{0.412}& 0.386& 0.417& 0.390& 0.418& \underline{0.385}& \underline{0.415}& 0.390&0.417\\
 & 720 & \textbf{0.421}& \textbf{0.451}& 0.439& 0.463& \underline{0.426}& \underline{0.452}& 0.431&  0.458& 0.433&0.456\\
\hline
\multicolumn{2}{c}{1st count} & 4&  5& 2&  3& 1&  2& 0&   0& 1&1\\
\multicolumn{2}{c}{2nd count} & 2&  1& 2&  0& 3&  3& 1&   3& 1&2\\
\hline
\end{tabular}
\caption{Full long-term forecasting results with random replaced words. }
\label{tab:ltf-timellm-random-replace-words-full}
\end{table*}

\begin{table*}[!ht]
\centering
\begin{tabular}{cccccccccccc}
\hline
\multicolumn{2}{c}{Ratio}& \multicolumn{2}{c}{0\%} & \multicolumn{2}{c}{10\%} & \multicolumn{2}{c}{40\%} & \multicolumn{2}{c}{70\%} & \multicolumn{2}{c}{100\%}\\
 Dataset&  Window& MSE& MAE& MSE& MAE& MSE& MAE& MSE& MAE & MSE&MAE\\
\hline
\multirow{4}{*}{ETTh1} & 96 & \underline{0.388}& \underline{0.414}& 0.403& 0.425& 0.395& 0.423& 0.390&  0.416& \textbf{0.383}&\textbf{0.412}\\
 & 192 & 0.435& 0.447& 0.416& \underline{0.433}& 0.424& 0.439& \underline{0.415}&  \textbf{0.430}& \textbf{0.414}&\underline{0.433}\\
 & 336 & \textbf{0.423}& \textbf{0.437}& \underline{0.426}& \underline{0.444}& 0.504& 0.487& 0.462&  0.465& 0.508&0.492\\
 & 720 & \textbf{0.477}& \textbf{0.490}& 0.520& 0.503& 0.522& 0.502& \underline{0.495}&  \underline{0.497}& 0.547& 0.518\\
\hline
\multirow{4}{*}{ETTh2} & 96 & \underline{0.294}& 0.355& 0.298& \underline{0.350}& 0.321& 0.368& \textbf{0.290}& \textbf{0.347}& 0.305& 0.363\\
 & 192 & 0.404& 0.422& \textbf{0.381}& \textbf{0.405}& 0.405& 0.416& 0.407&  0.426& \underline{0.392}& \underline{0.410}\\
 & 336 & \underline{0.400}& \underline{0.425}& 0.499& 0.477& 0.499& 0.486& \textbf{0.385}& \textbf{0.415}& 0.438&0.455\\
 & 720 & \textbf{0.409}& \textbf{0.441}& 0.477& 0.484& 0.531& 0.513& \underline{0.428}&  \underline{0.451}& 0.438&0.456\\
\hline
\multicolumn{2}{c}{1st count} & 3&  3& 1&  1& 0&  0& 2&   3& 2&1\\
\multicolumn{2}{c}{2nd count} & 3&  2& 1&  3& 0&  0& 3&   2& 1&2\\
\hline
\end{tabular}
\caption{Full long-term forecasting results with random replaced words. (Language model: Llama2)}
\label{tab:ltf-timellm-random-replace-words-full-llama2}
\end{table*}

\section{Supplementary Visualization Results}

\subsection{Text prototypes}
\begin{figure}[!htbp]
    \centering
    \includegraphics[width=0.8\textwidth]{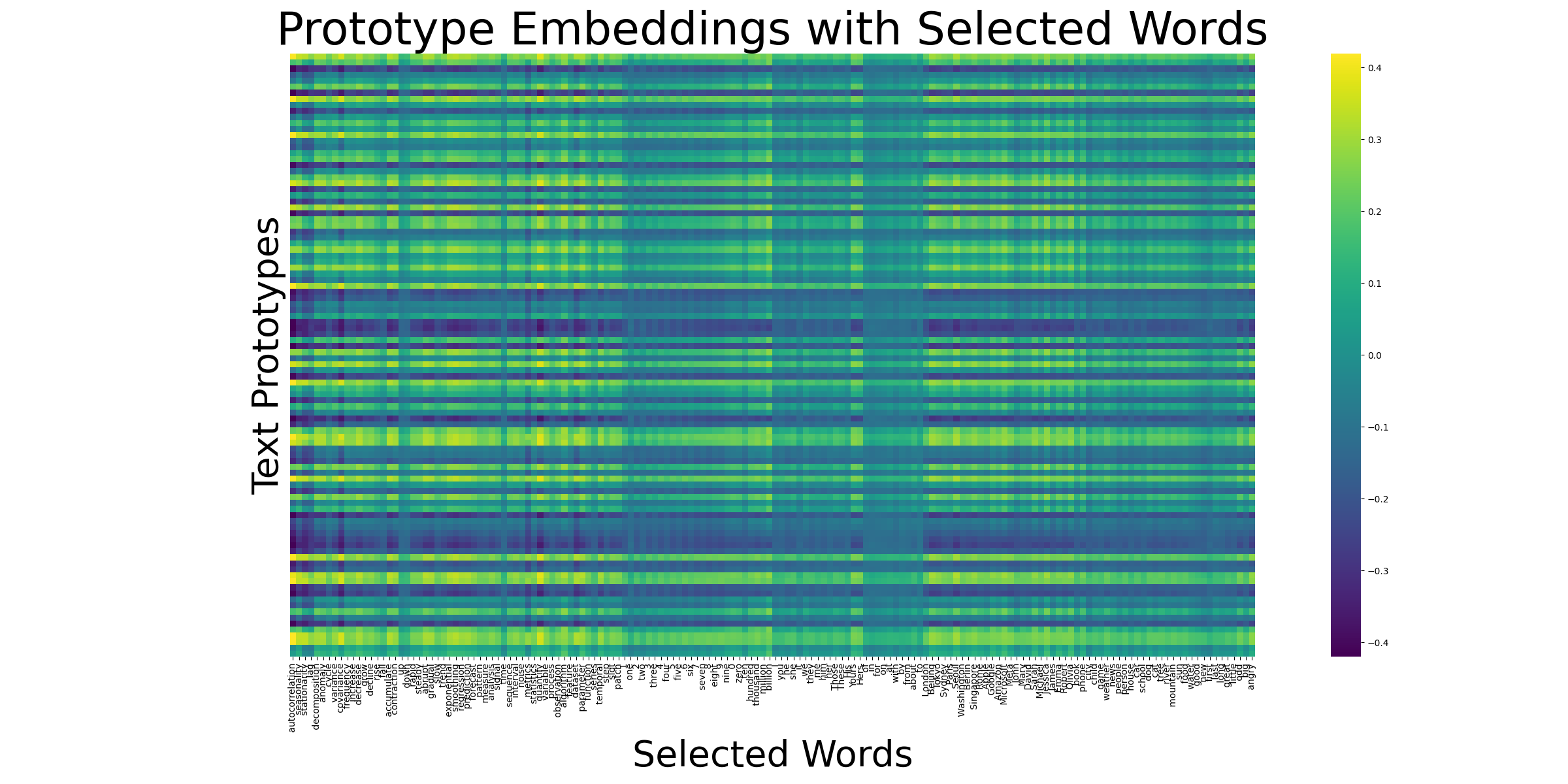}
    \caption{Similarity between text prototypes and selected words. (Baseline: TimeLLM; Language Model: BERT)}
    \label{fig:v1-proto-emb-bert}
\end{figure}

Fig.\ref{fig:v1-proto-emb-bert} is the similarity between text prototypes and selected words but using BERT as language model. We can find the relation between text prototypes and time series-related words is not significant.
Besides, we provide more tokens corresponding to text prototypes (in Table \ref{tab:v1-sup}).

\begin{table*}[!htbp]
    \centering
    \begin{adjustbox}{max width=0.9\textwidth}
    \begin{tabular}{>{\centering\arraybackslash}m{0.3\linewidth}>{\centering\arraybackslash}m{0.6\linewidth}}
    \hline
    Baseline-LM-EM & Tokens \\
    \hline
TimeLLM-GPT2-Linear& [[‘Ġadvertising'], [‘ĠexternalToEVA'], [‘Ġleaders'], [‘Ġeducation'], [‘ĠexternalToEVA'], [‘Ġrestricting'], [‘-'], [‘Ġeducation'], [‘Ġeducation'], [‘Ġleaders'], [‘Ġadvertising'], [‘ÿ'], [‘Ġartists'], [‘ĠexternalToEVA'], [‘ÿ'], [‘Ġadvertising'], [‘Ġeducation'], [‘ÿ'], [‘Ġand'], [‘Ġand'], [‘ĠEducation'], [‘Ġand'], [‘Ġeducation'], [‘ÿ'], [‘Ġand']]\\
    \hline
    TimeLLM-BERT-Linear&  [[‘[unused613]'], [‘[unused471]'], [‘[SEP]'], [‘\#\#elial'], [‘[unused239]'], [‘[unused239]'], [‘[SEP]'], [‘1738'], [‘\#\#elial'], [‘[SEP]'], [‘\#\#elial'], [‘[unused239]'], [‘\#\#elial'], [‘1738'], [‘\#\#elial'], [‘\#\#elial'], [‘\#\#\$'], [‘[unused239]'], [‘[SEP]'], [‘\#\#elial'], [‘[SEP]'], [‘[SEP]'], [‘\#\#elial'], [‘\#\#elial'], [‘[unused239]']]\\
    \hline
    CALF-GPT2-PCA&  [[‘Ġthe'], [‘Ġand'], [‘Ġhipp'], [‘ĠJack'], [‘Ġflickering'], [‘ĠiPads'], [‘Ġstories'], [‘Ġunits'], [‘ĠCharacters'], [‘Ġrefused'], [‘Ġdestroyed'], [‘ĠSever'], [‘Ġmafia'], [‘Ġmagnificent'], [‘Ġ277'], [‘Ġdeployed'], [‘Ġabducted'], [‘Ġathletic'], [‘Ġburg'], [‘develop'], [‘index'], [‘Ġ53'], [‘ĠRD'], [‘ĠLCS'], [‘ĠHF']]\\
    \hline
    \end{tabular}
    \end{adjustbox}
    \caption{Tokens corresponding to more text prototypes. LM: Language model; EM: Text prototypes extraction method.}
    \label{tab:v1-sup}
\end{table*}

\subsection{Attentions of cross-modality alignment module}
\begin{figure*}[!htbp]
    \centering
    \subfigure[TimeLLM (LM: BERT)]{
        \includegraphics[width=0.45\textwidth]{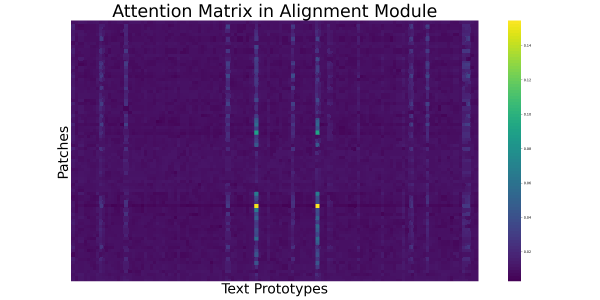}
        \label{v2-timellm-bert-att-align}
    }\hfill
    \subfigure[CALF (LM: GPT2)]{
        \includegraphics[width=0.45\textwidth]{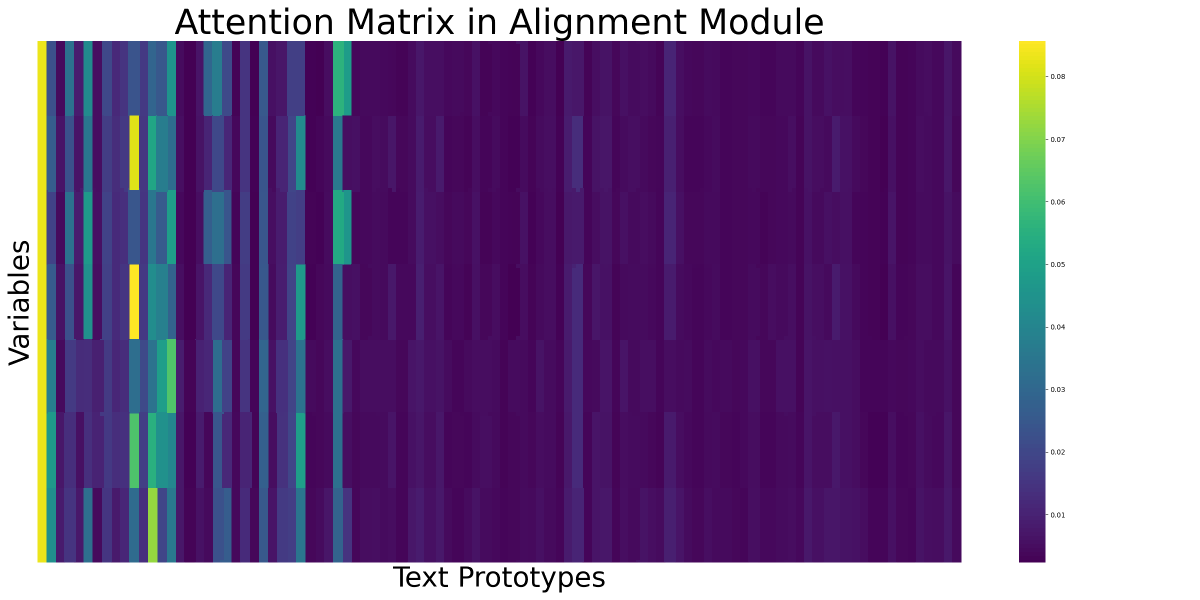}
        \label{v2-calf-gpt2-att-align}
    }
    \caption{Attentions of croess-modality module of TimeLLM using BERT and CALF using GPT2}
\end{figure*}

We supplement the results of TimeLLM based on BERT (see Fig.\ref{v2-timellm-bert-att-align}), with the results of CALF (see Fig.\ref{v2-calf-gpt2-att-align}). It can also be concluded that only a small subset of text prototypes are more important.

\subsection{Time series embeddings after alignment}
\begin{figure*}[!htbp]
    \centering
    \subfigure[TimeLLM (LM: BERT)]{
        \includegraphics[width=0.45\textwidth]{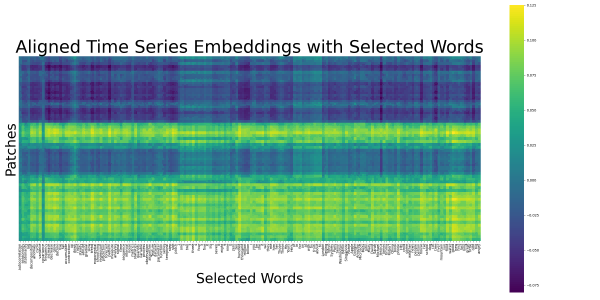}
        \label{v3-timellm-bert-align-words}
    }\hfill
    \subfigure[CALF (LM: GPT2)]{
        \includegraphics[width=0.45\textwidth]{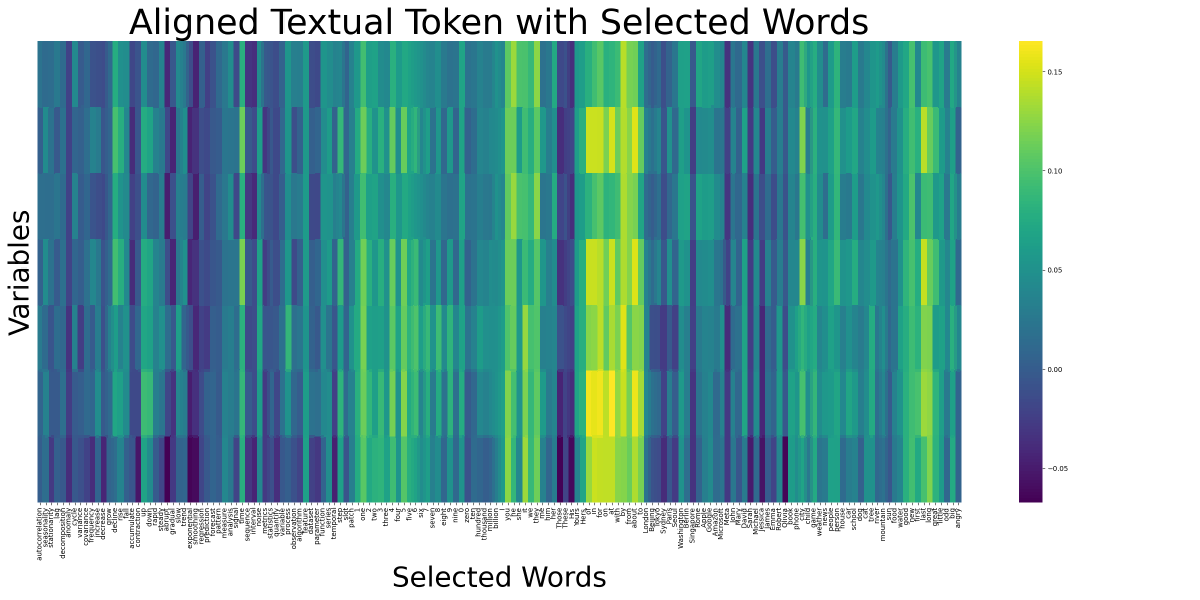}
        \label{v3-calf-gpt2-align-words}
    }
    \caption{Similarity between aligned time series embeddings and selected words. (Baseline: TimeLLM using BERT and CALF using GPT2)}
\end{figure*}

The results of TimeLLM based on BERT is showed in Fig.\ref{v3-timellm-bert-align-words}. These aligned time series embeddings are not more similar to words related to time series and some of them are not very similar to all 160 words.
The results of CALF is showed in Fig.\ref{v3-calf-gpt2-align-words}. For CALF, the similarity between variables and prepositions (such as “on”, “at", “by") is the most significant.

\subsection{Time series patches and corresponding textual representations}
\begin{figure*}[!htbp]
    \centering
    \subfigure[Tokens set “ĠMania"]{
        \includegraphics[width=0.45\textwidth]{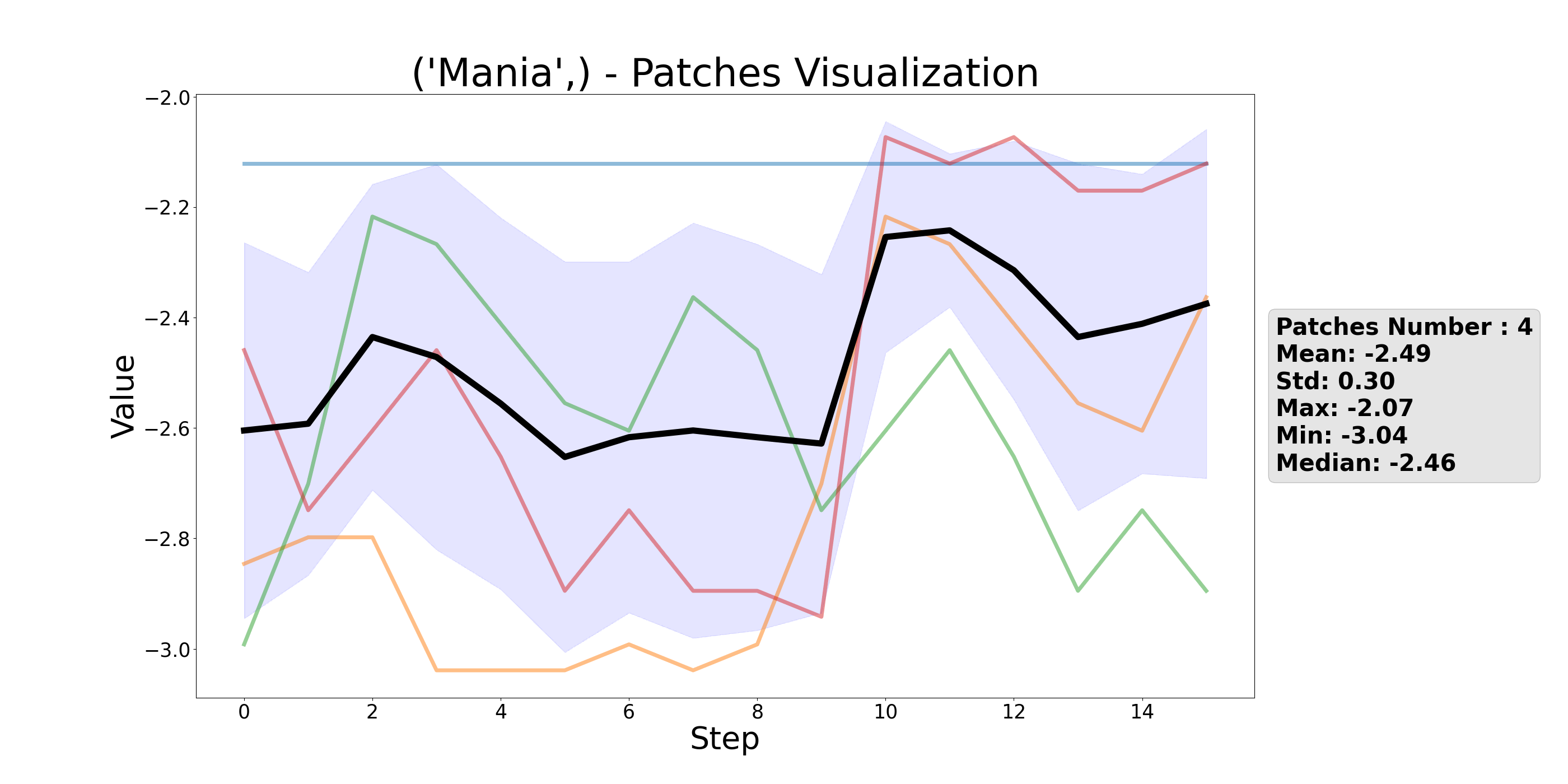}
        \label{v4-tokens-set-0}
    }\hfill
    \subfigure[Tokens set “ĠDelivery"]{
        \includegraphics[width=0.45\textwidth]{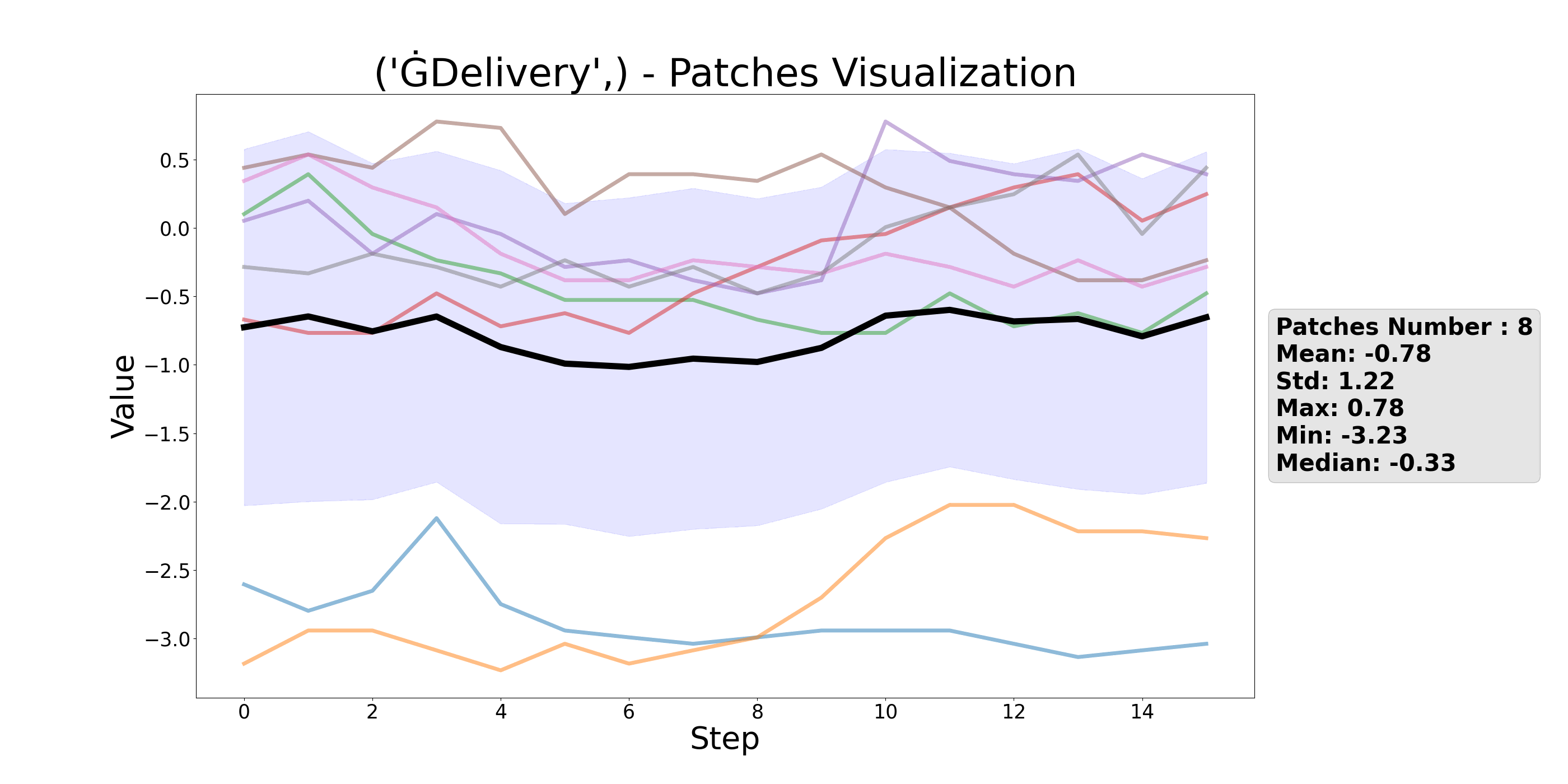}
        \label{v4-tokens-set-4}
    }\hfill
    \subfigure[Tokens set “ĠModule"]{
        \includegraphics[width=0.45\textwidth]{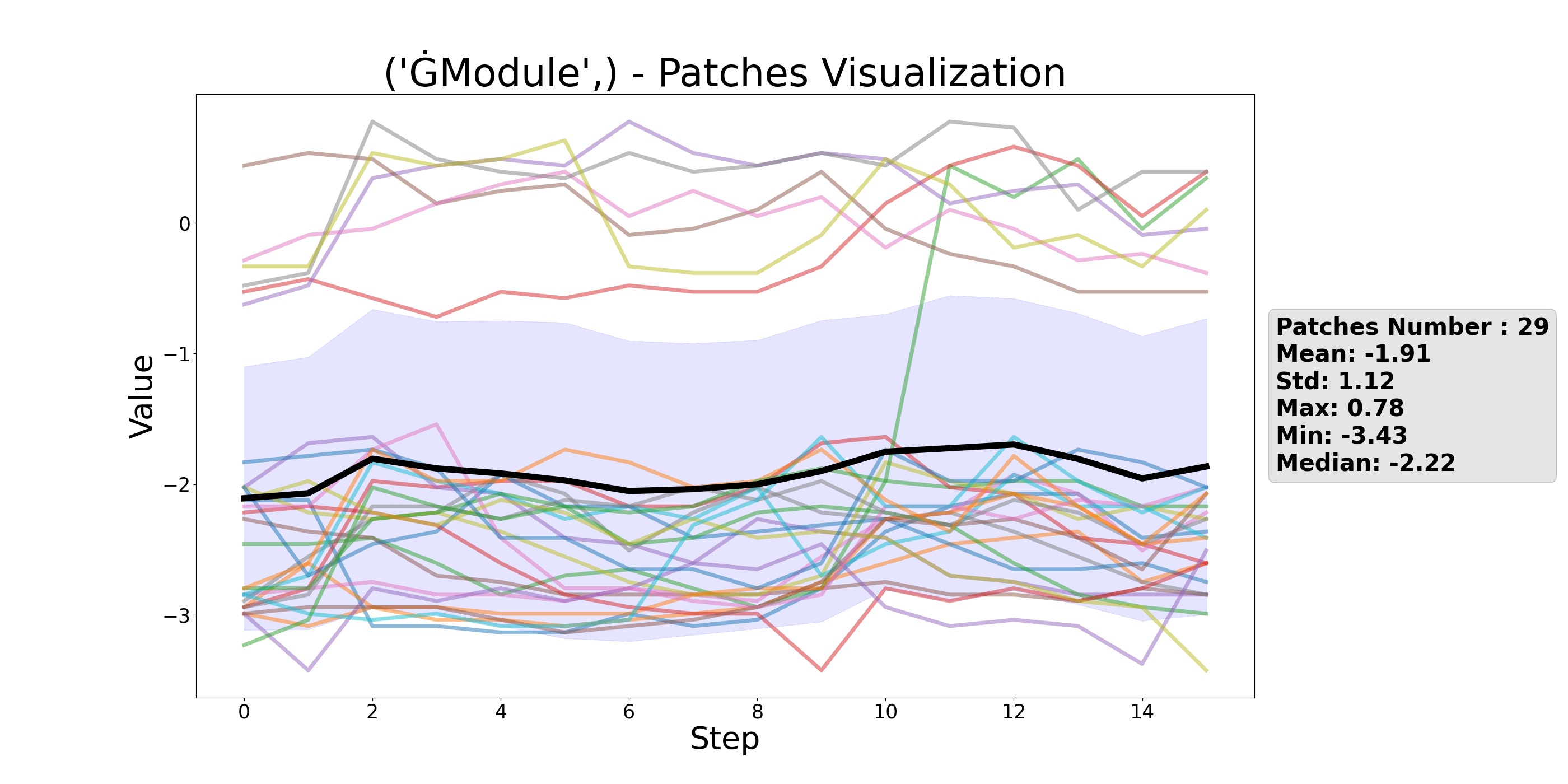}
        \label{v4-tokens-set-2}
    }\hfill
    \subfigure[Tokens set “StreamerBot"]{
        \includegraphics[width=0.45\textwidth]{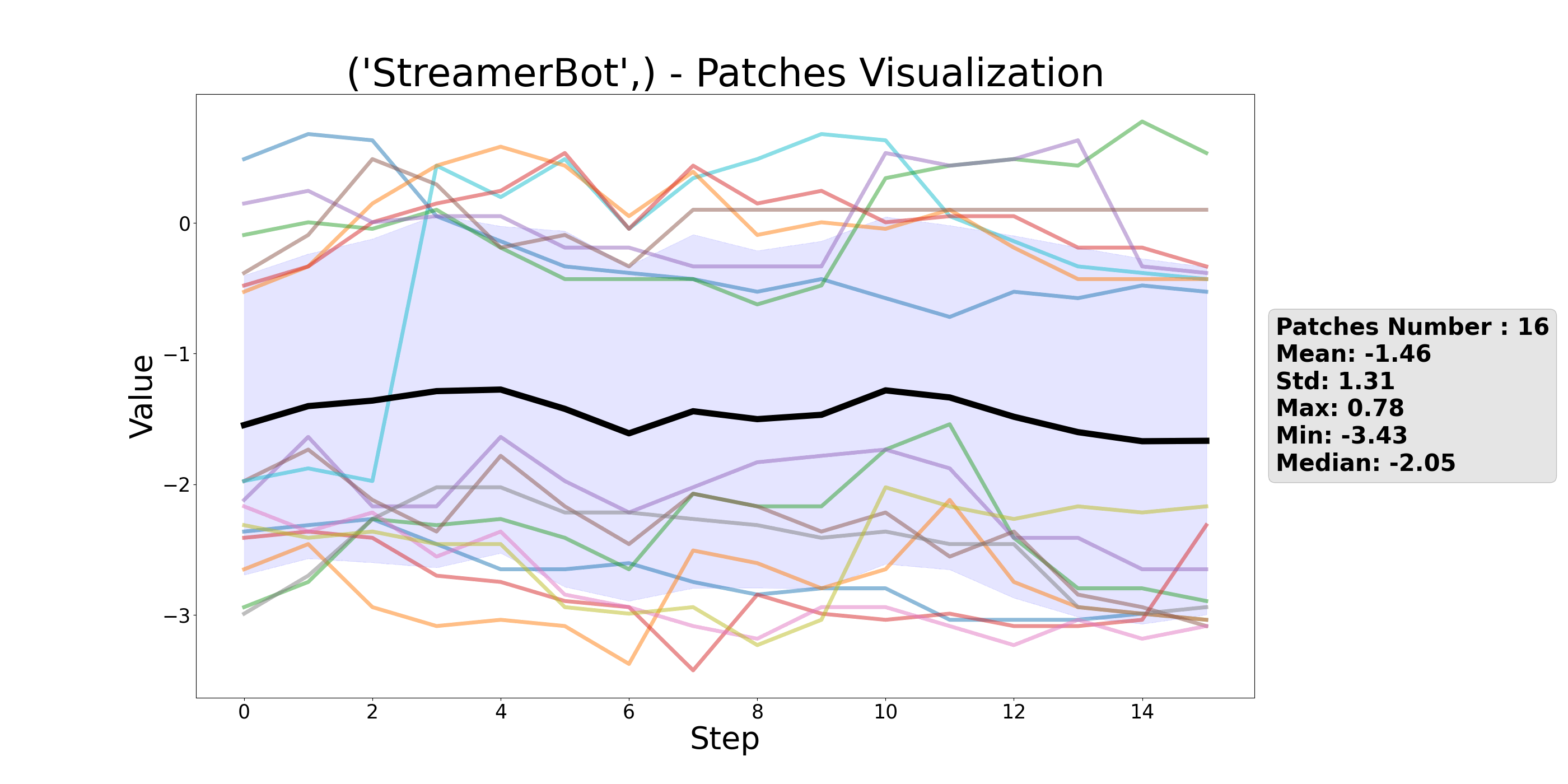}
        \label{v4-tokens-set-3}
    }
    \caption{Patches belong to more tokens sets. (Baseline: TimeLLM using GPT2)}
    \label{fig:v4-gpt2}
\end{figure*}

\begin{figure*}[!htbp]
    \centering
    \subfigure[Tokens set “like"]{
        \includegraphics[width=0.45\textwidth]{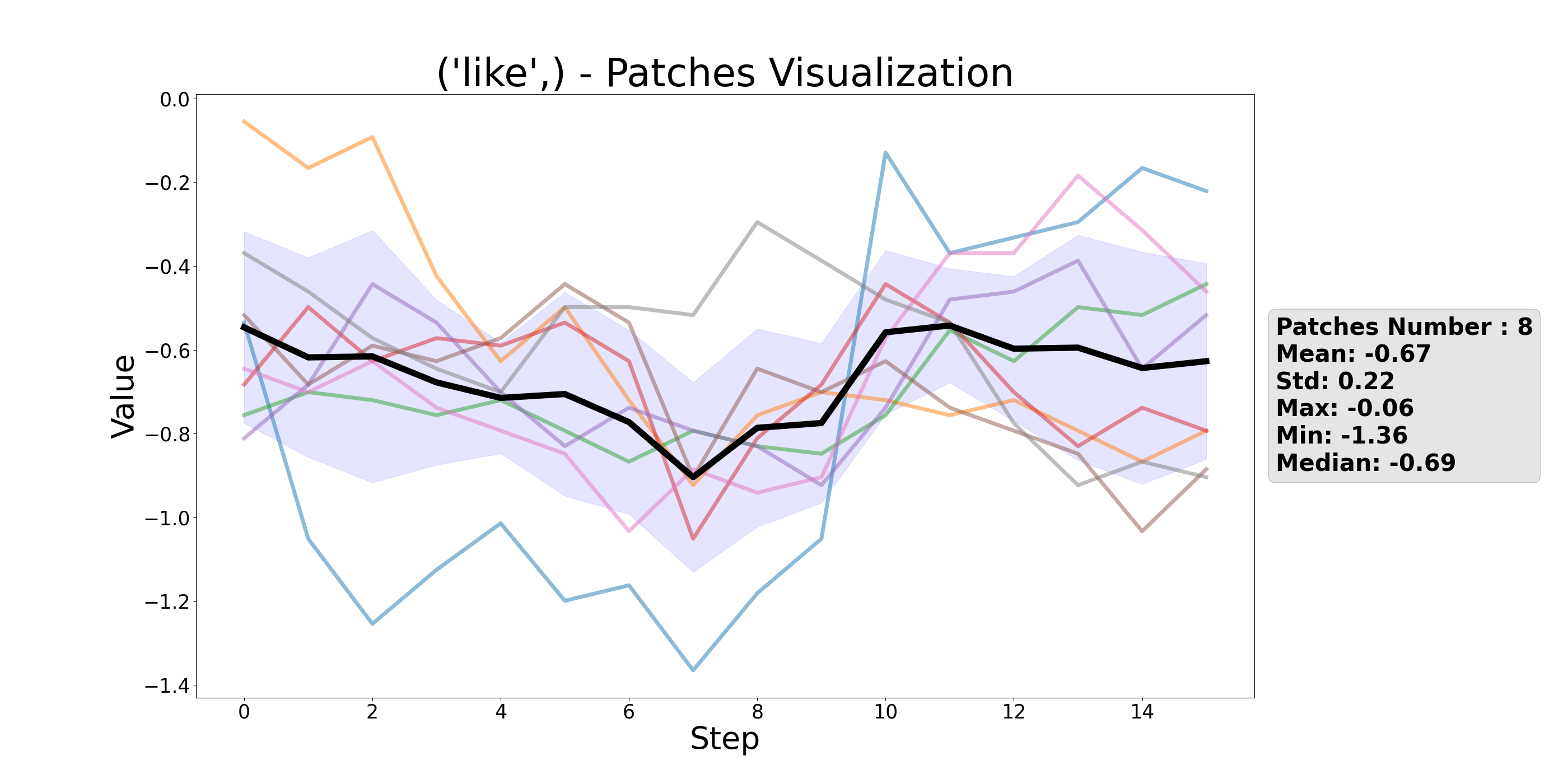}
        \label{v4-tokens-set-0-bert}
    }\hfill
    \subfigure[Tokens set “odin"]{
        \includegraphics[width=0.45\textwidth]{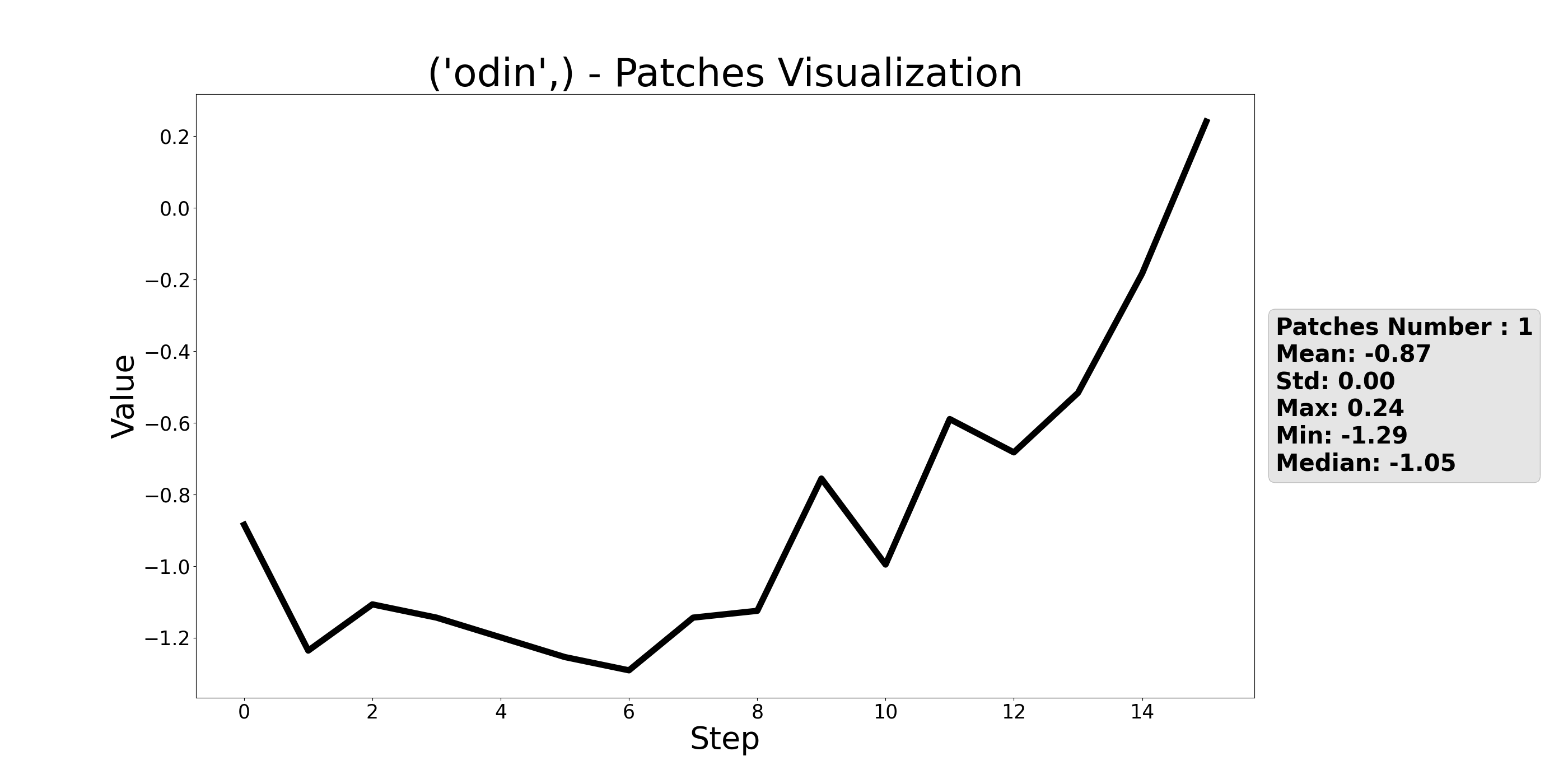}
        \label{v4-tokens-set-1-bert}
    }\hfill
    \subfigure[Tokens set “three"]{
        \includegraphics[width=0.5\textwidth]{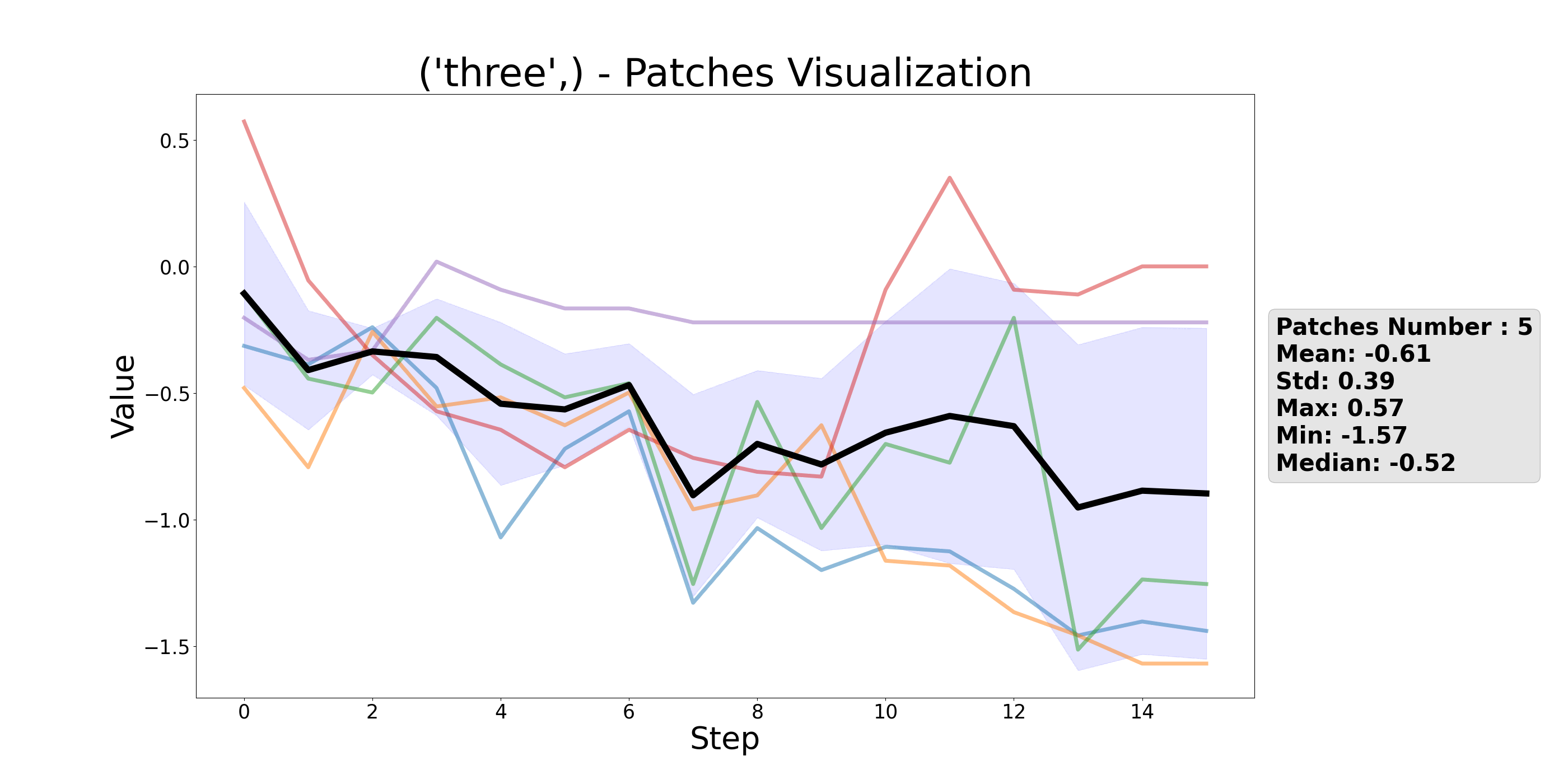}
        \label{v4-tokens-set-2-bert}
    }\hfill
    \subfigure[Tokens set “named"]{
        \includegraphics[width=0.45\textwidth]{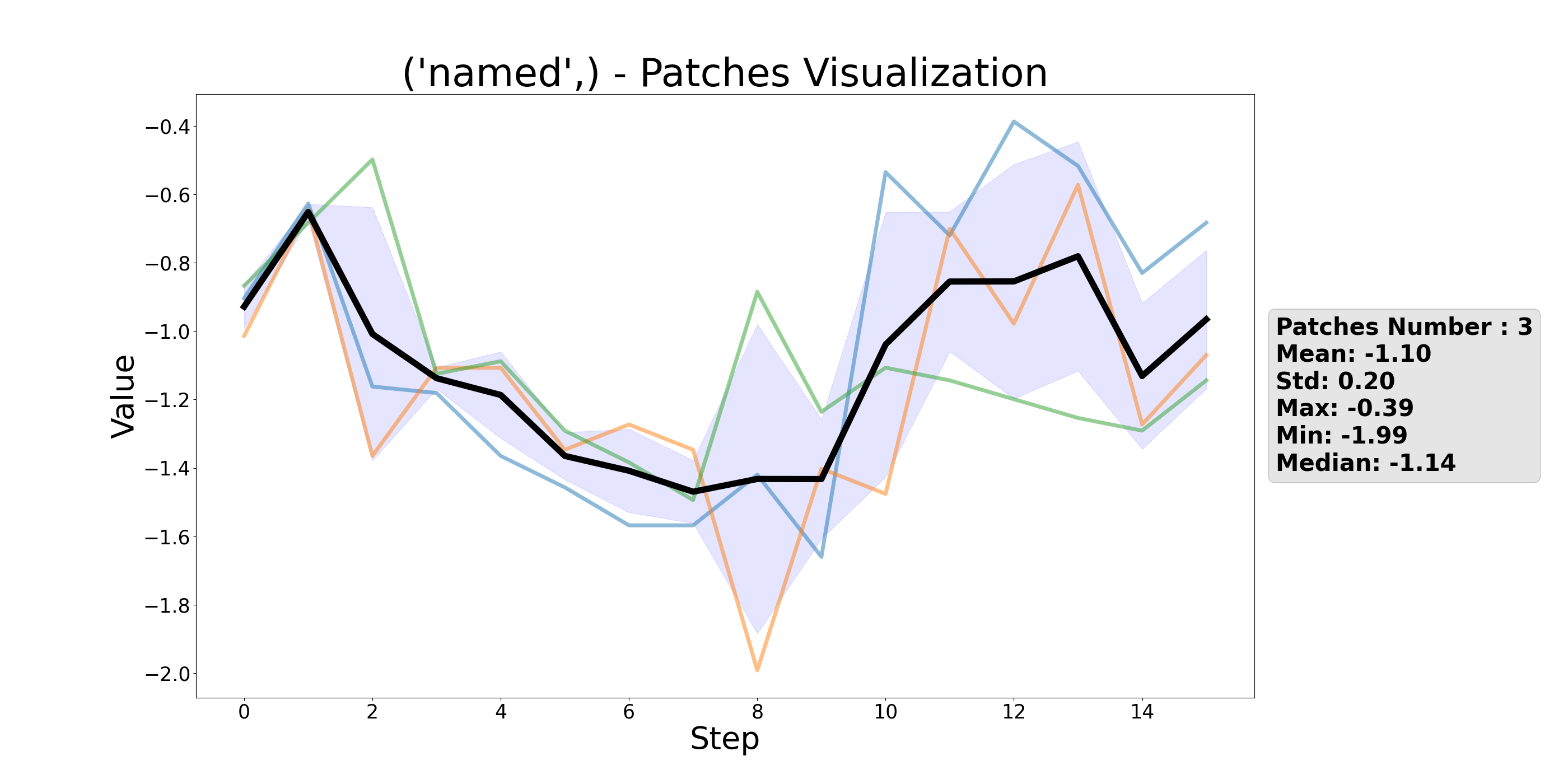}
        \label{v4-tokens-set-3-bert}
    }
    \caption{Patches belong to more tokens sets. (Baseline: TimeLLM using BERT)}
    \label{fig:v4-bert}
\end{figure*}

Fig.\ref{fig:v4-gpt2} and Fig.\ref{fig:v4-bert} display more token sets and patches of them. In Fig.\ref{fig:v4-gpt2}, the token set “ĠMania" consists of four patches, with two falling outside the standard deviation range; The “ĠDelivery" set includes eight patches, most of which are closely grouped, except for two notable outliers; The “ĠModule" and “StreamerBot" sets each split into two cohesive subgroups, showing obvious distinctions between them. In Fig.\ref{fig:v4-bert}, the token set “like" typically displays a decrease followed by an increase, with minimal deviations; The “odin" set has only one patch, suggesting a strong match; The “three" set's standard deviation range widens over time, starting consistently but diverging later; The “named" set maintains high consistency across its three patches.

\subsection{Attentions of LLMs}
\begin{figure}[!htbp]
    \centering
    \includegraphics[width=0.6\textwidth]{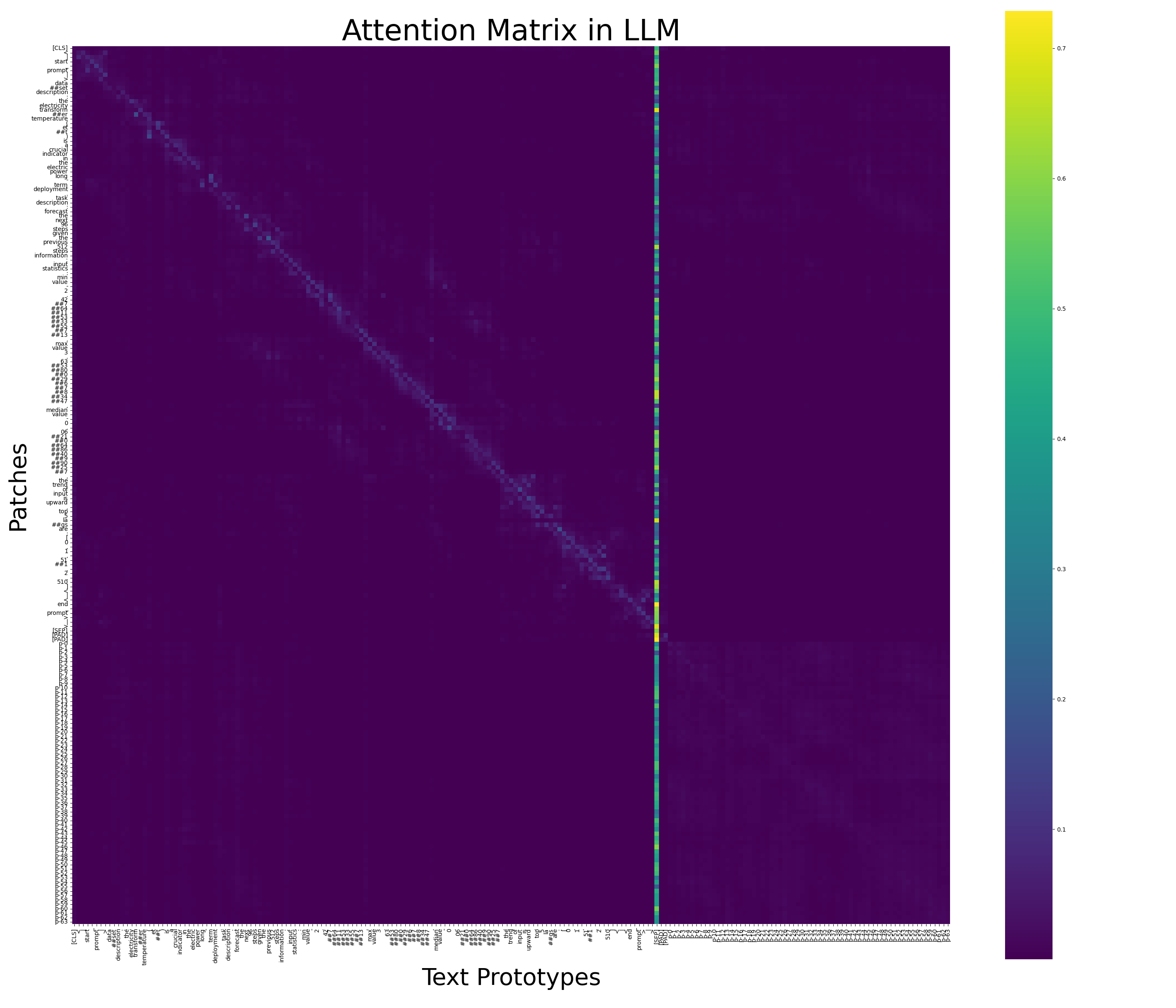}
    \caption{Attention of Language model in TimeLLM using BERT.}
    \label{fig:v5-att-llm-timellm-bert}
\end{figure}

Table \ref{tab:v5-tokens} shows 4 tokens from the attention of GPT-2 and top 5 tokens with the highest weights to them.
The attention of TimeLLM based on BERT is showed in Fig.\ref{fig:v5-att-llm-timellm-bert}. Similar to the result of GPT-2, most prompts and patches have the highest weights with the same token “[SEP]”. 

\begin{table*}[!ht]
    \centering
    \begin{adjustbox}{max width=0.9\textwidth}
    \begin{tabular}{>{\centering\arraybackslash}m{0.3\linewidth}>{\centering\arraybackslash}m{0.6\linewidth}}
    \hline
    Tokens in attention& Top 5 tokens with the highest weights to them\\
    \hline
“Ġupward”& “$<$”, “Ġtrend”, “Ġinput”, “Ġof”, “Ġupward”\\
    \hline
    “Ġtop”&  “$<$”, “,”, “Ġupward”, “Ġtop”, “Ġtrend”\\
    \hline
    “p-29”&  “$<$”, “p-27”, “p-28”, “p-29”, “p-26”\\
    \hline
    “p-54”&  “$<$”, “p-54”, “p-53”, “p-56”, “p-55”\\
    \hline
    \end{tabular}
    \end{adjustbox}
    \caption{Some tokens from attention of GPT-2 and top 5 tokens with the highest weights to them. “p-x” mean the x-th patch.}
    \label{tab:v5-tokens}
\end{table*}

\end{document}